\documentclass[final,1p,times,numbers,sort&compress]{elsarticle}

\usepackage{graphicx}
\usepackage{mathbbold}
\usepackage{subfigure}
\usepackage{amsmath}
\usepackage{amssymb}
\usepackage{chemsym}
\usepackage{multirow}
\usepackage{algorithm}
\usepackage{algorithmic}

\newtheorem{theorem}{Theorem}[section]

\newproof{pf}{Proof}

\journal{}

\begin{document}

\begin{frontmatter}



\title{Inexact Alternating Direction Method Based on Newton descent algorithm with Application to Poisson Image Deblurring}


\author[adr1,adr2]{Dai-Qiang Chen\corref{cor1}}

\cortext[cor1]{Corresponding author, telephone number: $86-13628484021$}
\ead{chener050@sina.com}

%

\address[adr1]{Department of Mathematics, School of Biomedical Engineering, Third Military Medical University, Chongqing 400038, Chongqing, People's Republic of China}
\address[adr2]{Department of Mathematics and System, School of Sciences, National University of Defense Technology, Changsha 410073, Hunan, People's Republic of China}

\begin{abstract}
The recovery of images from the observations that are degraded by a linear operator and further corrupted by Poisson noise is an important task in modern imaging applications such as astronomical and biomedical ones. Gradient-based regularizers involve the popular total variation semi-norm have become standard techniques for Poisson image restoration due to its edge-preserving ability. Various efficient algorithms have been developed for solving the corresponding minimization problem with non-smooth regularization terms. In this paper,
motivated by the idea of the alternating direction minimization algorithm and the Newton's method with upper convergent rate, we further propose inexact alternating direction methods utilizing the proximal Hessian matrix information of the objective function, in a way reminiscent of Newton descent methods. Besides, we also investigate the global convergence of the proposed algorithms under certain conditions. Finally, we illustrate that the proposed algorithms outperform the current state-of-the-art algorithms through numerical experiments on Poisson image deblurring.
\end{abstract}


\end{frontmatter}



\section{Introduction}\label{sec1}

Image deblurring is a classical ill-conditioned problem in many fields of applied sciences, including astronomy imaging and biomedical imaging. During the recording of a digital image, blurring artifacts always arise due to some unavoidable causes, e.g., the optical imaging system in a camera lens may be out of focus, in astronomy imaging the incoming light in the telescope may be slightly bent by turbulence in the atmosphere, and the same problem appears due to the diffraction of light in the fluorescence microscopy.

Mathematically, image blurring process in such applications can often be described as follows. For simplification we denote the $m\times n$ image as a one-dimensional vector in $\mathbb{R}^{N}(N=mn)$  by concatenating their columns. Let $u\in \mathbb{R}^{N}$ be the original image. The degradation model is described by
\begin{equation}\label{equ1.1}
g = K u
\end{equation}
where $g\in \mathbb{R}^{N}$ is the observed image and $K\in \mathbb{R}^{N\times N}$ is a linear blurring operator. Since the linear operator $K$ cannot be inverted, and $g$ is also possibly contaminated by random noises, the recovery of $u$ from the noisy version $f$ of the blurred observation $g$ is a ill-posed problem. Variational image restoration methods based on the regularization technique are the most popular approach for solving this problem. Typically, the variational model corresponds to solving the following minimization problem

\begin{equation}\label{equ1.2}
\min_{u} D_{f}(u)+J(u)
\end{equation}
where $D_{f}(u)$ is a data fidelity term which is derived from the the noise distribution, and $J(u)$ is a regularization term for imposing the prior on the unknown image $u$.

Generally, the data fidelity term controls the closeness between the original image $u$ and the observed image $f$. It takes different forms depending on the type of noise being added. For example, it is well known that the $l_{2}$-norm fidelity

\begin{equation}\label{equ1.3}
D_{f}(u)=\|K u-f\|_{2}^{2}
\end{equation}
is used for the additive white Gaussian noise. Such fidelity term is mostly considered in literature for its good characterization of noise of a optical imaging system. However, non-Gaussian noises are also presented in the real imaging, e.g., Poisson noise is generally observed in photon-limited images such as electronic microscopy \cite{IEEEMag:microscopy}, positron emission tomography \cite{IEEEConf:PET} and single photon emission computerized tomography \cite{IP:Poisson}. Due to its important applications in medical imaging, linear inverse problems in presence of Poisson noise have received
much interest in literature \cite{JVCI:Hybrid,VCIR:PISB,TIP:PIDAL}. The likelihood probability of the Poisson noisy data $f$ is given by

\begin{equation}\label{equ1.4}
p(f|Ku)=\prod_{i=1}^{N}\frac{\left((Ku)_{i}\right)^{f_{i}}}{f_{i}!}e^{-(Ku)_{i}}.
\end{equation}
Based on the statistics of Poisson noise and maximum a posterior (MAP) likelihood estimation approach, a generalized Kullback-Leibler (KL)-divergence \cite{AnnalStat:KL} arises as the fidelity term for Poisson deconvolution variational model, i.e.,

\begin{equation}\label{equ1.5}
D_{f}(u)=\langle \textbf{1}, Ku\rangle - \langle f, \log(Ku)\rangle.
\end{equation}

Besides the fidelity term, a regularization term is also needed to restrain the noise amplification and avoid other artifacts in the recovered image. A simple but efficient idea is to use sparse representation in some transform domain of the unknown image $u$. The choice of transform domain is crucial to obtain a suitable solution, and one popular choice is the total variation (TV) \cite{NumerMath:TV} due to the strong edge-preserving ability. In this case, we obtain the classical TV-KL model for Poisson image deblurring:

\begin{equation}\label{equ1.6}
\min_{u\geq 0} \langle \textbf{1}, Ku\rangle - \langle f, \log(Ku)\rangle + \lambda \|\nabla u\|_{1}
\end{equation}
where $\lambda>0$ is a regularization parameter, and $\|\nabla u\|_{1}=\sum_{i=1}^{N}\|(\nabla u)_{i}\|_{2}$ with $(\nabla u)_{i}=((\nabla u)_{i}^{1}, (\nabla u)_{i}^{2})$ is the total variation regularization. Since the pixel values of images represent the number of discrete photons incident over a given time interval in this application, we demand that $u\geq 0$ in model (\ref{equ1.6}). Another selection for the regularizer term is the wavelet tight framelets \cite{CPAM:Curvelet,ACHA:Framelets,JVCI:Hybrid}, which have also been proved to be efficient but may need more computational cost associated with the wavelet transform and inverse transform. In the last several years, the relationship between the total variation and wavelet framelet has also been revealed \cite{SIAMNumer:Equiva,JAMS:Equiva}.

In this paper, we focus our attention on the TV-KL model (\ref{equ1.6}). Due to the complex form of the fidelity term (\ref{equ1.5}), the ill-posed inverse problem in presence of Poisson noise has attracted
less interest in literature than their Gaussian counterpart. Recently, Sawatzky et al. \cite{Conference:EMTV} proposed an EM-TV algorithm for Poisson image deblurring which has been shown to be more efficient than earlier methods, such as TV penalized Richardson-Lucy algorithm \cite{MRT:TVRL}. S. Bonettini et al. \cite{IP:SGP,IP:SGPPoi} also developed gradient projection methods for TV-based image restoration. Later on, the augmented Lagrangian framework \cite{VCIR:PISB,TIP:PIDAL,IP:STV}, which has been successfully applied to various image processing tasks, has been used for solving the TV-KL model. In particular, in \cite{TIP:PIDAL} a very effective alternating direction method of multipliers (ADMM) called PIDAL was proposed for image deblurring in presence of Poisson noise, where a TV denoising problem is solved by Chambolle's algorithm in each iteration. It has been proved to be more efficient than the ADMM algorithm proposed in \cite{VCIR:PISB}. The relation between the two ADMMs with and without nested iteration has been analyzed in \cite{JVCI:Hybrid}.

Although the augmented Lagrangian methods have been shown to be very useful, inner iterations or inverse operators involving the linear operator $K$ and Laplacian operator are required in each iteration. Besides, at least three auxiliary variables, which may reduce the convergence speed of the iterative algorithm, need to be introduced in the augmented Lagrangian method due to the fidelity term is non-quadratic. In order to further improve the efficiency of the augmented Lagrangian method, alternating direction minimization methods based on the linearized technique have been widely investigated \cite{SIAMJSC:Linearized,JSC:LADM,IP:Linearized,SIAMImage:GILADM} very recently. The key idea of these methods is to use the proximal linearized term instead of the whole or part of the augmented Lagrangian function. As a result, sub-minimization problems which have closed solutions are obtained in the iteration process. In literature \cite{SIAMJSC:Linearized}, an efficient optimization algorithm using the linearized alternating direction method was proposed, and further applied to solve the TV minimization problem for multiplicative Gamma noise removal. Numerical examples demonstrate that it is more efficient than the augmented Lagrangian algorithms in this application. The primal-dual hybrid gradient (PDHG) method proposed by Zhu et al. \cite{UCLAReport:PDHG} is another efficient iterative algorithm. The core idea is to alternately update the primal and dual variables by the gradient descent scheme and the gradient ascend scheme. The recent study on variants of the original PDHG algorithm, and on the connection with the linearized version of ADMM reveals the equivalence relation between the two algorithms framework. For more details refer to \cite{JSC:LADM,JMIV:PDHG} and the references cited therein.

However, in the previous linearized alternating direction methods, the second-order derivative (or the Hessian matrix) of the objective function of the sub-minimization problem is just approximated by an identity matrix multiplied by some constant. This approximation is obviously not exact in most cases, and therefore may reduce the convergence speed of the iterative algorithms. In fact, from the numerical comparison shown in section \ref{sec4} we find that the convergence rate of the linearized alternating direction method proposed in \cite{SIAMJSC:Linearized} is obviously influenced by the inexact linearized approximation while applying it to solve the TV-KL model for Poisson image deblurring. It is observed that the computational efficiency of the linearized alternating direction is even lower than the previous augmented Lagrangian algorithms such as the PIDAL algorithm. Refer to the experiments below for details.

The main contribution of this work is to propose a novel inexact alternating direction method utilizing the second-order information of the objective function. Specifically, in one sub-minimization problem of the proposed algorithm, the solution is obtained by a one-step iteration, in a way reminiscent of Newton descent methods. In other words, the second-order derivative of the corresponding objective function in the sub-minimization problem is just approximated by a proximal Hessian matrix which can be computed easily, rather than a constant multiplied by the identity matrix. The improved iterative algorithm is proved to be more efficient than the current state-of-the-art methods with application to Poisson image deblurring, including the PIDAL algorithm and the linearized alternating direction methods.

The rest of this paper is organized as follows. In section \ref{sec2} we briefly review the recently proposed proximal linearized alternating direction (PLAD) method \cite{SIAMJSC:Linearized}. In section \ref{sec3}, in order to overcome the drawback of the previous linearized alternating direction method, we develop an inexact alternating direction method based on the Newton descent algorithm. The updating strategy of the proximal Hessian matrix in the Newton descent algorithm is also discussed, and then the convergence of the proposed algorithms is further investigated under certain conditions. In section \ref{sec4} the numerical examples on Poisson image deblurring problem are reported to compare the proposed algorithms with the recent state-of-the-art algorithms.

\section{Existing algorithms}\label{sec2}
\setcounter{equation}{0}

In this section, we briefly review the PLAD method proposed in \cite{SIAMJSC:Linearized}, and through further investigation we find that the PLAD method can be regarded as a linearized version of another widely used iterative algorithm----primal-dual hybrid gradient algorithm (PDHG), which was firstly proposed by Zhu et.al \cite{UCLAReport:PDHG}.

Fist of all, we consider the following TV regularized minimization problem

\begin{equation}\label{equ2.1}
\min\limits_{u\in U} D_{f}(u)+ \lambda \|\nabla u\|_{1}
\end{equation}
where $U=[u_{\min}, u_{\max}]^{N}$. Note that (\ref{equ2.1}) can also be reformulated as a constrained optimization problem as follows

\begin{equation}\label{equ2.2}
\min\limits_{u\in U, d} \left\{ D_{f}(u) + \lambda \|d\|_{1}~~ | ~~\nabla u = d\right\}.
\end{equation}
The augmented Lagrangian function for (\ref{equ2.2}) is given by

\begin{equation}\label{equ2.3}
\mathcal{L}_{\alpha}(u,d,p) = D_{f}(u)+ \lambda \|d\|_{1} + \langle p, d-\nabla u\rangle + \frac{\alpha}{2}\|d-\nabla u\|_{2}^{2}.
\end{equation}
Therefore, the well-known ADMM for solving (\ref{equ2.2}) can be formulated as

\begin{equation}\label{equ2.4}
\left\{\begin{array}{lll}u^{k+1} = \textrm{arg}\min \limits_{u\in U}\left\{D_{f}(u)+\langle p^{k}, d^{k}-\nabla u\rangle + \frac{\alpha}{2}\|d^{k}-\nabla u\|_{2}^{2} \right\}, &~ &~ ~\\
d^{k+1} = \textrm{arg}\min \limits_{d}\left\{\lambda\|d\|_{1}+ \langle p^{k}, d-\nabla u^{k+1}\rangle + \frac{\alpha}{2}\|d-\nabla u^{k+1}\|_{2}^{2} \right\},
&~ &~ ~\\ p^{k+1}=p^{k} + \alpha(d^{k+1} - \nabla u^{k+1}).& ~
&~
\end{array} \right.
\end{equation}
The solution $u^{k+1}$ of the first subproblem in (\ref{equ2.4}) satisfies the first-order optimality condition, i.e., it is the solution of the following nonlinear system of equations:
\[
\nabla D_{f}(u)-\alpha\Delta u + \textrm{div}(p^{k} + \alpha d^{k})=0
\]
which has no closed solution. Note that we have $\textrm{div}=-\nabla^{T}$.

Therefore, the linearization of the convex function $D_{f}(u)+\frac{\alpha}{2}\|d^{k}-\nabla u\|_{2}^{2}$ is adopted and the first subproblem of (\ref{equ2.4}) is simplified as

\begin{equation}\label{equ2.5}
u^{k+1} = \textrm{arg}\min_{u\in U} \left\{\langle \nabla D_{f}(u^{k}) + \alpha \textrm{div}(d^{k}-\nabla u^{k}), u-u^{k}\rangle + \langle p^{k}, d^{k}-\nabla u\rangle + \frac{1}{2\delta}\|u - u^{k}\|_{2}^{2} \right\}.
\end{equation}
Substituting the first subproblem in (\ref{equ2.4}) with (\ref{equ2.5}), we obtain the PLAD algorithm proposed in \cite{SIAMJSC:Linearized}, which is given by

\begin{equation}\label{equ2.9}
\left\{\begin{array}{lll}u^{k+1} = \textrm{arg}\min\limits_{u\in U} \left\{\langle \nabla D_{f}(u^{k}) + \alpha \textrm{div}(d^{k}-\nabla u^{k}), u-u^{k}\rangle + \langle p^{k}, d^{k}-\nabla u\rangle + \frac{1}{2\delta}\|u - u^{k}\|_{2}^{2} \right\}, &~ &~ ~\\
d^{k+1} = \textrm{arg}\min \limits_{d}\left\{\lambda\|d\|_{1}+ \langle p^{k}, d-\nabla u^{k+1}\rangle + \frac{\alpha}{2}\|d-\nabla u^{k+1}\|_{2}^{2} \right\},
&~ &~ ~\\ p^{k+1}=p^{k} + \alpha(d^{k+1} - \nabla u^{k+1}).& ~
&~
\end{array} \right.
\end{equation}

Note that the second-order information of the objective function is just approximated by $\frac{1}{\delta}\textbf{I}$, which is obviously inexact.

Choose $D_{f}(u)=\langle \textbf{1}, Ku\rangle - \langle f, \log(Ku)\rangle$ in (\ref{equ2.1}). Then we obtain the PLAD algorithm for solving the TV-KL model as follows:

\begin{equation}\label{equ2.6}
\left\{\begin{array}{lll}u^{k+1} = \mathcal{P}_{U}\left(u^{k}-\delta \left(K^{T}\left(1-\frac{f}{K u^{k}}\right)+\alpha \textrm{div}(d^{k}-\nabla u^{k})+\textrm{div} p^{k}\right)\right), &~ &~ ~\\
d^{k+1} = shrink\left(\nabla u^{k+1} - \frac{p^{k}}{\alpha},\frac{\lambda}{\alpha}\right),
&~ &~ ~\\ p^{k+1}=p^{k} + \alpha(d^{k+1} - \nabla u^{k+1})& ~
&~
\end{array} \right.
\end{equation}
where $\mathcal{P}_{U}$ denotes the projection onto the set $U$. In order to avoid the special case of $K u^{k}=0$ in the $u$-iteration step, we choose $u_{\min}=1$ in the following experiments, which is also adopted for the proposed algorithms. The shrinkage operator is componentwise, i.e., it is defined by
\[
shrink(s, c)_{i} = \max\left(\|s_{i}\|_{2}-c,0\right)\frac{s_{i}}{\|s_{i}\|_{2}}
\]
where $s_{i}\in \mathbb{R}^{2}$.

The PLAD algorithm can also be regarded as a linearized version of the state-of-the-art PDHG proposed in \cite{UCLAReport:PDHG,SIAM:GFPDHG,JMIV:PDHG}. Consider the minimization problem
\[
\min\limits_{u\in U} D_{f}(u)+J(Bu)
\]
where both $D_{f}$ and $J$ are convex and lower-semicontinuous (l.s.c.) functions, and $B$ is a linear operator.
The original PDHGMp in \cite{SIAM:GFPDHG,JMIV:PDHG} can be expressed as follows

\begin{equation}\label{equ2.7}
\left\{\begin{array}{lll}u^{k+1}=\textrm{arg}\min \limits_{u\in U} D_{f}(u)+ \frac{1}{2\tau}\|u-(u^{k}-\tau B^{T}\bar{b}^{k})\|_{2}^{2}, &~ &~ ~\\
b^{k+1}=\textrm{arg}\min \limits_{b} J^{*}(b)+\frac{1}{2\sigma}\|b-(b^{k}+\sigma B u^{k+1})\|_{2}^{2},
&~ &~ ~\\ \bar{b}^{k+1}=b^{k+1}+\theta(b^{k+1}-b^{k}).
\end{array} \right.
\end{equation}
where $J^{*}$ is the convex conjugate of $J$. Let $B=\nabla$, $J(B u)=\lambda\|\nabla u\|_{1}$, $\tau=\delta$ and $\sigma=\alpha$. Then utilizing the celebrated Moreau's identity \cite{MMS:PFBS}
\[
u = (1+\sigma F)^{-1}(u) + \sigma (1+\sigma^{-1} F^{*})^{-1}(\sigma^{-1}u),
\]
equations (\ref{equ2.7}) can be reformulated as
\begin{equation}\label{equ2.8}
\left\{\begin{array}{lll}u^{k+1}=\textrm{arg}\min \limits_{u\in U} D_{f}(u)+ \frac{1}{2\delta}\|u-(u^{k}-\delta \nabla^{T}\bar{b}^{k})\|_{2}^{2}, &~ &~ ~\\
d^{k+1}=\textrm{arg}\min \limits_{d} \lambda\|d\|_{1}+\frac{\alpha}{2}\|d-(\nabla u^{k+1}+\alpha^{-1}b^{k})\|_{2}^{2}, &~ &~ ~\\
b^{k+1}=b^{k}+\alpha(\nabla u^{k+1}-d^{k+1}), &~ &~ ~\\
\bar{b}^{k+1}=b^{k+1}+\theta(b^{k+1}-b^{k}).
\end{array} \right.
\end{equation}
Choose $\theta=1$, $b^{k}=-p^{k}$, $\bar{b}^{k}=-\bar{p}^{k}$, and replace $D_{f}(u)$ by its linearized version $D_{f}(u^{k})+\langle \nabla D_{f}(u^{k}), u-u^{k}\rangle$. Then we can easily deduce the PLAD algorithm shown in (\ref{equ2.9}).

\section{Proposed inexact alternating direction method based on the Newton descent algorithm}\label{sec3}
\setcounter{equation}{0}

\subsection{Algorithm description}\label{subsec3.1}

Consider the $u$-subproblem in (\ref{equ2.4}). Let
\[
G(u)=D_{f}(u)+\langle p^{k}, d^{k}-\nabla u\rangle + \frac{\alpha}{2}\|d^{k}-\nabla u\|_{2}^{2}.
\]
It is easy to observe from the first formulas of (\ref{equ2.9}) and (\ref{equ2.6}) that the solution $u^{k+1}$ in the PLAD algorithm is obtained by

\begin{equation}\label{equ3.1rev2}
u^{k+1} = \mathcal{P}_{U}(u^{k}-\delta\nabla G(u^{k}))
\end{equation}
which implies that an approximate solution of the $u$-subproblem in (\ref{equ2.4}) is obtained by the projection gradient descent algorithm, which only utilizes the first-order information of the objective function, typically have a sub-linear convergence rate. It is well-known that Newton or quasi-Newton methods, which further utilize the Hessian matrix of the objective function, have been presented with a super-linear convergence rate. This fact motivates us to design a more efficient algorithm based on the Newton methods to obtain an approximate solution of the $u$-subproblem. Here we adopt the expression of the Newton descent algorithm, i.e., the solution $u^{k+1}$ is obtained by an one-step projection Newton descent algorithm as follows.

\begin{equation}\label{equ3.2rev2}
u^{k+1} = \mathcal{P}_{U}(u^{k}-\omega^{k}(\nabla^{2} G(u^{k}))^{-1}\nabla G(u^{k}))
\end{equation}
where $\omega^{k}\in [0, 1]$ is the relaxed parameter. The iterative formula (\ref{equ3.2rev2}) can be reformulated as
\begin{equation}\label{equ3.2}
u^{k+1} = \mathcal{P}_{U}\left(u^{k} - \omega^{k}(\nabla^{2} D_{f}(u^{k})+\alpha \nabla^{T}\nabla)^{-1}\left(\nabla D_{f}(u^{k}) + \alpha \textrm{div}(d^{k}-\nabla u^{k}) +\textrm{div} p^{k}\right)\right)
\end{equation}
Here we assume that the Hessian matrix $\nabla^{2} D_{f}(u^{k})+\alpha \nabla^{T}\nabla$ is inverse.

In what follows, we consider the special case of TV-KL model for Poisson image deblurring. In this case, we have $D_{f}(u)=\langle \textbf{1}, Ku\rangle - \langle f, \log(Ku)\rangle$, and (\ref{equ3.2}) can be reformulated as

\begin{equation}\label{equ3.3}
u^{k+1} = \mathcal{P}_{U}\left(u^{k} - \omega^{k}\left(K^{T}\left(\frac{f}{(Ku^{k})^{2}}\right)K+\alpha \nabla^{T}\nabla\right)^{-1}\\
\left(K^{T}\left(1-\frac{f}{K u^{k}}\right) + \alpha \textrm{div}(d^{k}-\nabla u^{k}) +\textrm{div} p^{k}\right)\right).
\end{equation}
However, the computation of the inverse of the operator $L^{k}=K^{T}\left(\frac{f}{(Ku^{k})^{2}}\right)K+\alpha \nabla^{T}\nabla$ is difficult in the update formula (\ref{equ3.3}). One simple strategy is to use a proximal Hessian matrix $\tilde{L} = \delta^{k} K^{T}K + \alpha \nabla^{T}\nabla$, which is a block-circulant matrix with periodic boundary conditions and hence can be easily computed by fast Fourier
transforms (FFTs), instead of the original operator $L^{k}$. In this situation, we obtain the following inexact alternating direction method based on the Newton descent algorithm with adaptive parameters (IADMNDA): 

\begin{equation}\label{equ3.4}
\left\{\begin{array}{lll}u^{k+1} = \mathcal{P}_{U}\left(u^{k}-\omega^{k}(\delta^{k} K^{T}K + \alpha \nabla^{T}\nabla)^{-1} \left(K^{T}\left(1-\frac{f}{K u^{k}}\right)+\alpha \textrm{div}(d^{k}-\nabla u^{k})+\textrm{div} p^{k}\right)\right), &~ &~ ~\\
d^{k+1} = shrink\left(\nabla u^{k+1} - \frac{p^{k}}{\alpha},\frac{\lambda}{\alpha}\right),
&~ &~ ~\\ p^{k+1}=p^{k} + \alpha(d^{k+1} - \nabla u^{k+1}).& ~
&~
\end{array} \right.
\end{equation}
In the proposed IADMNDA algorithm (\ref{equ3.4}), a parameter $\delta^{k}$ is used to approximate the term $\frac{f}{(Ku^{k})^{2}}$ in the operator $L^{k}$, and therefore $L^{k}$ is replaced by a simple block-circulant matrix $\tilde{L}$.

In what follows, we further discuss the selection of the parameters $\delta^{k}$ and $\omega^{k}$ in the IADMNDA algorithm. For the relaxed parameter $\omega^{k}$, we choose it to satisfy that

\begin{equation}\label{equ3.3rev2}
u^{k}-\omega^{k}r^{k}\in U
\end{equation}
for guaranteeing the convergence of the proposed algorithm.
Here
\[
r^{k}=(\delta^{k} K^{T}K + \alpha \nabla^{T}\nabla)^{-1} \left(K^{T}\left(1-\frac{f}{K u^{k}}\right)+\alpha \textrm{div}(d^{k}-\nabla u^{k})+\textrm{div} p^{k}\right).
\]
In the Poisson image deblurring problem, we always choose $u_{\max}=+\infty$, and hence
the condition (\ref{equ3.3rev2}) comes into existence while choosing $\omega^{k}$ monotone non-increasing and small enough. In this setting, the projection operator $\mathcal{P}_{U}$ can be removed from the first formula of (\ref{equ3.4}).


For the parameter $\delta^{k}$, one strategy is to update its value in the iterative step according to the widely used Barzilai-Borwein (BB) spectral approach \cite{NumerAnal:BB}. Let $v=Ku$, $v^{k}=Ku^{k}$, and $H(v)=\langle \textbf{1}, v\rangle - \langle f, \log v\rangle$. The parameter $\delta^{k}$ is chosen such that $\delta^{k}I$ mimics the Hessian matrix $\nabla^{2}H(v)$ over the most recent step. Specifically, we require that

\begin{equation}\label{equ3.5}
\delta^{k} = \textrm{arg}\min_{\delta}\|\nabla H(v^{k})-\nabla H(v^{k-1})-\delta(v^{k}-v^{k-1})\|_{2},
\end{equation}
and immediately get

\begin{equation}\label{equ3.6}
\delta^{k} = \frac{\left\langle  \left(1-\frac{f}{K u^{k}}\right)-\left(1-\frac{f}{K u^{k-1}}\right), Ku^{k}-Ku^{k-1}\right\rangle}{\|Ku^{k}-Ku^{k-1}\|_{2}^{2}}.
\end{equation}
The whole process of the proposed algorithm is summarized as Algorithm 1. It is observed that the update of $\delta^{k}$ introduces the extra convolution operation including in $K u^{k}$. Therefore, one simple strategy is to use a unchanged value for $\delta^{k}$ during the iteration, i.e., $\delta^{k}\equiv \delta$, where $\delta$ is a constant. In this setting, we abbreviate the proposed algorithm as IADMND.

\begin{algorithm}[htb]
\caption{ Proximal Hessian matrix based inexact alternating direction method with adaptive parameter (IADMNDA) for Poisson image deblurring}
\begin{algorithmic}[1]
\REQUIRE  observation $f$; regularization parameter $\lambda$; parameters $\delta_{0}$ and $\alpha$; inner iteration number $m$.\\
\textbf{Initialization}: $k=0$; $u^{0}=f$; $d^{0}=\nabla f$; $p^{0}=\textbf{0}$; $\delta^{0}=\delta_{0}$; $\omega^{-1}=\textbf{1}$.\\
\textbf{Iteration}: \\
~~~(\romannumeral1) update $u$: \\
~~~$r^{k}=(\delta^{k} K^{T}K + \alpha \nabla^{T}\nabla)^{-1} \left(K^{T}\left(1-\frac{f}{K u^{k}}\right)+\alpha \textrm{div}(d^{k}-\nabla u^{k})+\textrm{div} p^{k}\right)$; \\
~~~update $\omega^{k}$ according to (\ref{equ3.3rev2}); \\
~~~$u^{k+1} = u^{k}-\omega^{k}r^{k}$; \\
~~~(\romannumeral2) update $d$: \\
~~~$d^{k+1} = shrink\left(\nabla u^{k+1} - \frac{p^{k}}{\alpha},\frac{\lambda}{\alpha}\right)$; \\
~~~(\romannumeral3) update $p$: \\
~~~$p^{k+1}=p^{k} + \alpha(d^{k+1} - \nabla u^{k+1})$; \\
~~~update $\delta^{k+1}$ according to (\ref{equ3.6}); \\
~~~(\romannumeral4) $k = k+1$; \\
until some stopping criterion is satisfied. \\
\textbf{Output} the recovered image $u = u^{k+1}$.
\end{algorithmic}
\end{algorithm}

\subsection{Convergence analysis}\label{subsec3.2}

In this subsection, we further investigate the global convergence of the proposed IADMND(A) algorithms for Poisson image deblurring under certain conditions. The bound constrained TV regularized minimization
problem (\ref{equ2.2}) can be reformulated as

\begin{equation}\label{equ3.50c}
\min\limits_{u, d} \left\{ D_{f}(u) + \iota_{U}(u) + \lambda \|d\|_{1}~~ | ~~\nabla u = d\right\}
\end{equation}
where $\iota_{U}$ denotes the indicator function of set $U$, i.e., $\iota_{U}(x)=0$ if $x\in U$ and $\iota_{U}(x)=+\infty$ otherwise.

Assume that $(u^{*}, d^{*})$ is one solution of the above bound constrained optimization problem corresponding to TV-KL model with $D_{f}(u)=\langle \textbf{1}, Ku\rangle - \langle f, \log(Ku)\rangle$, and $p^{*}$ is the corresponding Lagrangian multiplier. Then the point $(u^{*}, d^{*}, p^{*})$ is a Karush-Kuhn-Tucker (KKT) \cite{Book:NP} point of problem (\ref{equ3.50c}), i.e., it satisfies the following conditions:

\begin{eqnarray}\label{equ3.8}
\nabla D_{f}(u^{*}) - \nabla^{T}p^{*} + \partial \iota_{U}(u^{*})\ni 0, \nonumber \\
\lambda \partial \|d^{*}\|_{1} + p^{*}\ni 0,\\
d^{*} = \nabla u^{*} \nonumber
\end{eqnarray}
where $\partial \iota_{U}(u^{*})$ denotes the set of the subdifferential of $\iota_{U}$ at $u^{*}$, and $\partial \|d^{*}\|_{1}$ denotes the set of the subdifferential of $\|\cdot\|_{1}$ at $d^{*}$. From literature \cite{Book:BertsekasConvex} we know that $\partial \iota_{U}(u^{*})$ is also equal to the normal cone $\mathcal{N}_{U}(u^{*})$ at $u^{*}$. Besides, assume that the convex function $D_{f}(u)$ satisfies:
\begin{equation}\label{equ3.21}
\underline{\gamma_{D}}\|K(u_{1}-u_{2})\|_{2}^{2}\leq (u_{1}-u_{2})^{T}\nabla^{2}D_{f}(u)( u_{1}-u_{2})\leq \overline{\gamma_{D}}\|K(u_{1}-u_{2})\|_{2}^{2}
\end{equation}
for any $u, u_{1}, u_{2}\in U$, where $\underline{\gamma_{D}}$ and $\overline{\gamma_{D}}$ are two positive constants (the estimation of $\underline{\gamma_{D}}$ and $\overline{\gamma_{D}}$ is discussed at the end of this section). 


\begin{theorem}\label{the1}
(The convergence of the proposed IADMNDA algorithm) Let $\{u^{k}, d^{k}, p^{k}\}$ be the sequence generated by the IADMNDA algorithm with $\delta_{\min}=\min\limits_{k}\delta^{k}\geq \overline{\gamma_{D}}$, $\delta_{\max}=\max\limits_{k}\delta^{k}<+\infty$, $\max\limits_{k}(\delta^{k+1}-\delta^{k})\leq \underline{\gamma_{D}}$ and $\|K^{T}K\|_{2}>0$. Then $\{u^{k}\}$ converges to a solution of the minimization problem (\ref{equ2.1}).
\end{theorem}

\begin{pf}
Denote $L_{k}=(\omega^{k})^{-1}\left(\delta^{k} K^{T}K + \alpha \nabla^{T}\nabla\right)$. According to the iterative formula with respect to $u$ we know that $u^{k+1}$ is the solution of the minimization problem

\begin{equation}\label{equ3.9}
\min\limits_{u\in U} \left\{\langle \nabla D_{f}(u^{k}) + \alpha \textrm{div}(d^{k}-\nabla u^{k}) +\textrm{div} p^{k}, u-u^{k}\rangle  + \frac{1}{2}(u - u^{k})^{T}L_{k}(u - u^{k}) \right\}.
\end{equation}
Therefore, the sequence $\{u^{k}, d^{k}, p^{k}\}$ generated by the IADMNDA algorithm satisfies

\begin{equation}\label{equ3.10}
\left\{\begin{array}{lll}\nabla D_{f}(u^{k}) + L_{k}(u^{k+1}-u^{k})+\alpha \nabla^{T}(\nabla u^{k}-d^{k}-\alpha^{-1}p^{k}) + \partial \iota_{U}(u^{k+1})\ni 0, &~ &~ ~\\
\lambda \partial \|d^{k+1}\|_{1} + \alpha(d^{k+1}-\nabla u^{k+1}+\alpha^{-1}p^{k})\ni 0,
&~ &~ ~\\ p^{k+1}=p^{k} + \alpha(d^{k+1}-\nabla u^{k+1})& ~ &~
\end{array} \right.
\end{equation}
where $\partial \|d^{k+1}\|_{1}$ denotes the set of the subdifferential of $\|\cdot\|_{1}$ at $d^{k+1}$.
Due to $\left(u^{*}, d^{*}, p^{*}\right)$ is one solution of (\ref{equ3.50c}), it is also the KKT point that satisfies:

\begin{equation}\label{equ3.11}
\left\{\begin{array}{lll}\nabla D_{f}(u^{*}) - \nabla^{T}p^{k} + \partial \iota_{U}(u^{*})\ni 0, &~ &~ ~\\
\lambda \partial \|d^{*}\|_{1} + p^{k}\ni 0,
&~ &~ ~\\ p^{*}=p^{*} + \alpha(d^{*}-\nabla u^{*}).& ~ &~
\end{array} \right.
\end{equation}

Denote the errors by $u^{k}_{e} = u^{k}-u^{*}$, $d^{k}_{e} = d^{k}-d^{*}$, and $p^{k}_{e} = p^{k}-p^{*}$.
Subtracting (\ref{equ3.11}) from (\ref{equ3.10}), and taking the inner product with $u^{k+1}_{e}$, $d^{k+1}_{e}$ and $p^{k}_{e}$ on both sides of the three equations, we obtain that

\begin{equation}\label{equ3.12}
\left\{\begin{array}{lll}\langle \nabla D_{f}(u^{k})-\nabla D_{f}(u^{*}), u^{k+1}_{e}\rangle + \langle u^{k+1}_{e}-u^{k}_{e}, u^{k+1}_{e}\rangle_{L_{k}}+ \\ \alpha\langle \nabla u^{k}_{e}-d^{k}_{e}-\alpha^{-1}p^{k}_{e}, \nabla u^{k+1}_{e} \rangle + \langle r^{k+1}-r^{*}, u^{k+1}_{e}\rangle= 0, &~ &~ ~\\
\lambda \langle s^{k+1}-s^{*}, d^{k+1}_{e}\rangle + \alpha \langle d^{k+1}_{e}-\nabla u^{k+1}_{e}+\alpha^{-1}p^{k}_{e}, d^{k+1}_{e}\rangle= 0,
&~ &~ ~\\ -\langle p^{k}_{e}-p^{k+1}_{e}, p^{k}_{e}\rangle + \alpha \langle \nabla u^{k+1}_{e}-d^{k+1}_{e}, p^{k}_{e}\rangle =0.& ~ &~
\end{array} \right.
\end{equation}
where $r^{k+1}\in \partial \iota_{U}(u^{k+1})$, $r^{*}\in \partial \iota_{U}(u^{*})$, $s^{k+1}\in \partial \|d^{k+1}\|_{1}$, and $s^{*}\in \partial \|d^{*}\|_{1}$.

Utilizing  $\langle x-y, x\rangle_{Q} = \frac{1}{2}\left(\|x\|^{2}_{Q} + \|x-y\|^{2}_{Q}- \|y\|^{2}_{Q}\right)$ $\left(\|x\|^{2}_{Q}=x^{T}Qx\right)$, and $p^{k+1}=p^{k} + \alpha(d^{k+1} - \nabla u^{k+1})$, we can reformulate the third equation of (\ref{equ3.12}) as
\begin{equation}\label{equ3.13}
\frac{1}{2\alpha}\left(\|p^{k+1}_{e}\|_{2}^{2}-\|p^{k}_{e}\|_{2}^{2}\right)-\frac{\alpha}{2}\|d^{k+1} - \nabla u^{k+1}\|_{2}^{2}-\langle p^{k}_{e}, d^{k+1}_{e} - \nabla u^{k+1}_{e}\rangle = 0.
\end{equation}
Therefore, summing three formulas in (\ref{equ3.12}) we can obtain that

\begin{equation}\label{equ3.14}
\begin{split}
\langle \nabla D_{f}(u^{k})-\nabla D_{f}(u^{*}), u^{k+1}_{e}\rangle +  \langle r^{k+1}-r^{*}, u^{k+1}_{e}\rangle + \lambda \langle s^{k+1}-s^{*}, d^{k+1}_{e}\rangle \\
+ \frac{1}{2}\left(\|u^{k+1}_{e}\|^{2}_{L_{k}}-\|u^{k}_{e}\|^{2}_{L_{k}}+\|u^{k+1}-u^{k}\|^{2}_{L_{k}}\right)+ \frac{1}{2\alpha}\left(\|p^{k+1}_{e}\|_{2}^{2}-\|p^{k}_{e}\|_{2}^{2}\right)\\
-\frac{\alpha}{2}\|d^{k+1} - \nabla u^{k+1}\|_{2}^{2} + \alpha\langle \nabla u^{k}_{e}-d^{k}_{e}, \nabla u^{k+1}_{e} \rangle +\alpha \langle d^{k+1}_{e}-\nabla u^{k+1}_{e}, d^{k+1}_{e}\rangle =0.
\end{split}
\end{equation}
Due to $\nabla D_{f}(u^{k})-\nabla D_{f}(u^{*})=\nabla^{2}D_{f}(\zeta^{k})u^{k}_{e}$ ($\zeta^{k}\in [u^{*}, u^{k}]$, with $[u^{*}, u^{k}]$ denoting the line segment between $u^{*}$ and $u^{k}$), we have
\[
\begin{split}
\langle \nabla D_{f}(u^{k})-\nabla D_{f}(u^{*}), u^{k+1}_{e}\rangle = (u^{k}_{e})^{T}\nabla^{2}D_{f}(\zeta^{k})u^{k+1}_{e} \\ =\frac{1}{2}\left(
\|u^{k}_{e}\|^{2}_{\nabla^{2}D_{f}(\zeta^{k})}+\|u^{k+1}_{e}\|^{2}_{\nabla^{2}D_{f}(\zeta^{k})}- \|u^{k+1}-u^{k}\|^{2}_{\nabla^{2}D_{f}(\zeta^{k})}\right).
\end{split}
\]
Besides, we also have $d^{k+1} - \nabla u^{k+1}=d^{k+1}_{e}-d^{k}_{e}+d^{k}_{e}-\nabla u^{k+1}_{e}$, and
\[
\langle \nabla u^{k}_{e}-d^{k}_{e}, \nabla u^{k+1}_{e} \rangle = \langle \nabla u^{k}_{e}-\nabla u^{k+1}_{e}, \nabla u^{k+1}_{e}\rangle + \langle \nabla u^{k+1}_{e}-d^{k}_{e}, \nabla u^{k+1}_{e}\rangle.
\]
Based on the above three relations, expression (\ref{equ3.14}) can be reformulated as

\begin{equation}\label{equ3.15}
\begin{split}
2\langle r^{k+1}-r^{*}, u^{k+1}_{e}\rangle + 2\lambda \langle s^{k+1}-s^{*}, d^{k+1}_{e}\rangle + \frac{1}{\alpha}\|p^{k+1}_{e}\|^{2}_{2} + \alpha\|d^{k+1}_{e}\|^{2}_{2}  \\
+ \left(\|u^{k+1}_{e}\|^{2}_{L_{k}}+ \|u^{k+1}_{e}\|^{2}_{\nabla^{2}D_{f}(\zeta^{k})} + \|u^{k}_{e}\|^{2}_{\nabla^{2}D_{f}(\zeta^{k})}-\alpha \|\nabla u^{k+1}_{e}\|^{2}_{2}\right) + \alpha\|d^{k} - \nabla u^{k+1}\|_{2}^{2} \\
+ \left(\|u^{k+1}-u^{k}\|^{2}_{L_{k}}-\|u^{k+1}-u^{k}\|^{2}_{\nabla^{2}D_{f}(\zeta^{k})}-\alpha \|\nabla (u^{k+1}-u^{k})\|^{2}_{2}\right) \\
\leq \frac{1}{\alpha}\|p^{k}_{e}\|^{2}_{2} + \alpha\|d^{k}_{e}\|^{2}_{2} + \|u^{k}_{e}\|^{2}_{L_{k}}-\alpha \|\nabla u^{k}_{e}\|^{2}_{2}.
\end{split}
\end{equation}

Due to $u^{k+1}, u^{*}\in U$, by the convexity of $\iota_{U}$ and the definition of the subdifferential we conclude that $\langle r^{k+1}-r^{*}, u^{k+1}_{e}\rangle\geq 0$. Similarly, by the convexity of the function $\|\cdot\|_{1}$ we also have that $\langle s^{k+1}-s^{*}, d^{k+1}_{e}\rangle\geq 0$.
Due to $\delta_{\min}\geq \overline{\gamma_{D}}$, we get

\begin{equation}\label{equ3.17}
\|u^{k+1}-u^{k}\|^{2}_{L_{k}}-\|u^{k+1}-u^{k}\|^{2}_{\nabla^{2}D_{f}(\zeta^{k})}-\alpha \|\nabla (u^{k+1}-u^{k})\|^{2}_{2}\geq \left(\delta_{\min}-\overline{\gamma_{D}}\right)\|K(u^{k+1}-u^{k})\|^{2}_{2}\geq 0.
\end{equation}
Therefore, removing the first two non-negative terms in (\ref{equ3.15}), and utilizing the inequalities (\ref{equ3.17}) we obtain that

\begin{equation}\label{equ3.18}
\begin{split}
\frac{1}{\alpha}\|p^{k+1}_{e}\|^{2}_{2} + \alpha\|d^{k+1}_{e}\|^{2}_{2}
+ \left(\|u^{k+1}_{e}\|^{2}_{L_{k}}+ \|u^{k+1}_{e}\|^{2}_{\nabla^{2}D_{f}(\zeta^{k})} + \|u^{k}_{e}\|^{2}_{\nabla^{2}D_{f}(\zeta^{k})}-\alpha \|\nabla u^{k+1}_{e}\|^{2}_{2}\right) \\
+ \alpha \|\nabla u^{k+1} - d^{k}\|^{2}_{2}  + \left(\delta_{\min}-\overline{\gamma_{D}}\right)\|K(u^{k+1}-u^{k})\|^{2}_{2}
\leq \frac{1}{\alpha}\|p^{k}_{e}\|^{2}_{2} + \alpha\|d^{k}_{e}\|^{2}_{2} + \left(\|u^{k}_{e}\|^{2}_{L_{k}}-\alpha \|\nabla u^{k}_{e}\|^{2}_{2}\right).
\end{split}
\end{equation}


According to the definition of $\omega^{k}$ in Algorithm 1, we know that $\omega^{k}$ is monotone non-increasing, and hence there exists $\omega^{*}$ such that $\lim\limits_{k\rightarrow +\infty}\omega^{k}=\omega^{*}$. Denote $\tilde{L}_{k}=(\omega^{*})^{-1}\left(\delta^{k} K^{T}K + \alpha \nabla^{T}\nabla\right)$. By the boundedness of $\delta^{k}$ and $\lim\limits_{k\rightarrow +\infty}\omega^{k}=\omega^{*}$ we have that
\begin{equation}\label{equ3.6rev2}
\lim\limits_{k\rightarrow +\infty} \tilde{L}_{k}-L_{k}=\textbf{0}.
\end{equation}
Since $\delta^{k+1}-\delta^{k}\leq \underline{\gamma_{D}}$, we know that
$\delta^{k}K^{T}K + \nabla^{2}D_{f}(\zeta^{k})\succeq \delta^{k+1}K^{T}K$ according to the condition (\ref{equ3.21}). Therefore, by the definition of $\tilde{L}_{k}$ we further have

\begin{equation}\label{equ3.16}
\|u^{k+1}_{e}\|^{2}_{\tilde{L}_{k}}+ \|u^{k+1}_{e}\|^{2}_{\nabla^{2}D_{f}(\zeta^{k})}-\alpha \|\nabla u^{k+1}_{e}\|^{2}_{2}\geq \|u^{k+1}_{e}\|^{2}_{\tilde{L}_{k+1}}-\alpha \|\nabla u^{k+1}_{e}\|^{2}_{2}\geq 0.
\end{equation}
Based on (\ref{equ3.18}) and (\ref{equ3.16}) we immediately get

\begin{equation}\label{equ3.5rev2}
\begin{split}
\frac{1}{\alpha}\|p^{k+1}_{e}\|^{2}_{2} + \alpha\|d^{k+1}_{e}\|^{2}_{2}
+ \left(\|u^{k+1}_{e}\|^{2}_{\tilde{L}_{k+1}}-\alpha \|\nabla u^{k+1}_{e}\|^{2}_{2}\right) + \left(\|u^{k}_{e}\|^{2}_{\nabla^{2}D_{f}(\zeta^{k})} + \|u^{k+1}_{e}\|^{2}_{L_{k}-\tilde{L}_{k}}-\|u^{k}_{e}\|^{2}_{L_{k}-\tilde{L}_{k}}\right)\\
+ \alpha \|\nabla u^{k+1} - d^{k}\|^{2}_{2}  + \left(\delta_{\min}-\overline{\gamma_{D}}\right)\|K(u^{k+1}-u^{k})\|^{2}_{2}
\leq \frac{1}{\alpha}\|p^{k}_{e}\|^{2}_{2} + \alpha\|d^{k}_{e}\|^{2}_{2} + \left(\|u^{k}_{e}\|^{2}_{\tilde{L}_{k}}-\alpha \|\nabla u^{k}_{e}\|^{2}_{2}\right).
\end{split}
\end{equation}
According to (\ref{equ3.6rev2}) we can easily conclude that, there exists some $k_{0}$ such that
\[
\sum_{k=k_{0}}^{+\infty}\left(\|u^{k}_{e}\|^{2}_{\frac{1}{2}\nabla^{2}D_{f}(\zeta^{k})} + \|u^{k+1}_{e}\|^{2}_{L_{k}-\tilde{L}_{k}}-\|u^{k}_{e}\|^{2}_{L_{k}-\tilde{L}_{k}}\right)>0.
\]
In the next, summing (\ref{equ3.18}) from some $k_{0}$ to $+\infty$ we obtain that

\begin{equation}\label{equ3.19}
\begin{split}
\sum_{k=k_{0}}^{+\infty}\|u^{k}_{e}\|^{2}_{\frac{1}{2}\nabla^{2}D_{f}(\zeta^{k})} +\alpha\sum_{k=k_{0}}^{+\infty}\|\nabla u^{k+1} - d^{k}\|^{2}_{2}+ \left(\delta_{\min}-\overline{\gamma_{D}}\right)\sum_{k=k_{0}}^{+\infty}\|K(u^{k+1}-u^{k})\|^{2}_{2}\\
\leq \frac{1}{\alpha}\|p^{k_{0}}_{e}\|^{2}_{2} + \alpha\|d^{k_{0}}_{e}\|^{2}_{2} + \left(\|u^{k_{0}}_{e}\|^{2}_{L_{k_{0}}}-\alpha \|\nabla u^{k_{0}}_{e}\|^{2}_{2}\right)
\end{split}
\end{equation}
which implies that

\begin{equation}\label{equ3.20}
\lim_{k\rightarrow +\infty}\|u^{k}-u^{*}\|^{2}_{\frac{1}{2}\nabla^{2}D_{f}(\zeta^{k})}=0, ~~~~~\lim_{k\rightarrow +\infty}\|\nabla u^{k+1} - d^{k}\|_{2}=0.
\end{equation}
Due to $\|u^{k}-u^{*}\|^{2}_{\frac{1}{2}\nabla^{2}D_{f}(\zeta^{k})}\geq \frac{1}{2}\underline{\gamma_{D}} \|Ku^{k}-Ku^{*}\|^{2}_{2}$, we get
\[
\lim\limits_{k\rightarrow +\infty}\|Ku^{k}-Ku^{*}\|_{2}=0.
\]
Due to $\|K^{T}K\|_{2}>0$, we further have $\lim\limits_{k\rightarrow +\infty}\|u^{k}-u^{*}\|_{2}=0$, which implies that $\{u^{k}\}$ converges to a solution of the minimization problem (\ref{equ2.1})

\end{pf}

\begin{theorem}\label{the2}
(The convergence of the proposed IADMND algorithm) Let $\{u^{k}, d^{k}, p^{k}\}$ be the sequence generated by the IADMND algorithm with $\delta\geq \overline{\gamma_{D}}$, and $\|K^{T}K\|_{2}>0$. Then $\{u^{k}\}$ converges to a solution of the minimization problem (\ref{equ2.1}).
\end{theorem}

\begin{pf}
The proof is analogous to that presented in Theorem \ref{the1}, and the only difference lies in that $\delta^{k}$ is replaced by a constant $\delta$. Here we neglect the proof due to limited space.
\end{pf}

In the above proof, we observe that the constants $\underline{\gamma_{D}}$ and $\overline{\gamma_{D}}$ in (\ref{equ3.21}) are crucial for the convergence of the proposed algorithms, since they decide the range of the parameters $\delta^{k}$ in IADMNDA algorithm and $\delta$ in IADMND algorithm respectively. For the Poisson image deblurring problem, we have $D_{f}(u)=\langle \textbf{1}, Ku\rangle - \langle f, \log(Ku)\rangle$. Then we can obtain that

\begin{equation}\label{equ3.22}
\begin{split}
\min_{i}\left(\frac{f_{i}}{(K\zeta)^{2}_{i}}\right)\|K(u_{1}-u_{2})\|_{2}^{2}\leq (u_{1}-u_{2})^{T}\nabla^{2}D_{f}(\zeta)( u_{1}-u_{2}) \\= \left(K(u_{1}-u_{2})\right)^{T}\textrm{diag}\left(\frac{f}{(K\zeta)^{2}}\right)\left(K(u_{1}-u_{2})\right)
\leq \max_{i}\left(\frac{f_{i}}{(K\zeta)^{2}_{i}}\right)\|K(u_{1}-u_{2})\|_{2}^{2}
\end{split}
\end{equation}
where $\textrm{diag}(u)$ denotes a diagonal matrix with diagonal elements of the components of $u$. Due to $\frac{f_{\max}}{(K u)_{\min}^{2}}$ and $\frac{f_{\min}}{(K u)_{\max}^{2}}$ are upper and lower bounds of $\left(\frac{f_{i}}{(K\zeta)^{2}_{i}}\right)$, they are also some estimation of $\overline{\gamma_{D}}$ and $\underline{\gamma_{D}}$. In this extreme case, the value of $\overline{\gamma_{D}}$ is too large and the value of $\underline{\gamma_{D}}$ is too small, and thus the parameter $\delta^{k}$ can be too large, and the ``step size" $(\delta^{k} K^{T}K + \alpha \Delta)^{-1}$ can be too small. This may cause the proposed algorithms converge very slowly. Therefore, similarly to \cite{SIAMJSC:Linearized}, we use the average of the second derivative $\frac{f}{(K u)^{2}}$ instead of the worst estimation of $\overline{\gamma_{D}}$ and $\underline{\gamma_{D}}$, which implies that a smaller $\delta^{k}$ can be selected during the implementation of the proposed algorithms.

Assume that $\Omega_{j}$ is a collection of the image region with $u=u_{j}$. When $\zeta$ is sufficiently close to the unknown image $u$,
\begin{equation}\label{equ3.23}
\begin{split}
E\left(\frac{f}{(K \zeta)^{2}}\right)\approx E\left(\frac{f}{(K u)^{2}}\right)\approx \sum_{j}\sum_{i\in \Omega_{j}}\frac{f_{i}}{(K u_{j})^{2}}\approx \sum_{j} \frac{|\Omega_{j}|}{(K u_{j})}\approx E\left(\frac{1}{K u}\right)
\approx \frac{1}{E(Ku)}\left(1+\frac{\textrm{Var}(Ku)}{E^{2}(Ku)}\right)
\end{split}
\end{equation}
where the third approximation equation uses the relation of $\sum_{i\in \Omega_{j}}f_{i}\approx |\Omega_{j}|E(f)\approx |\Omega_{j}|(Ku_{j})$, and the last approximation is obtained by the second-order Taylor expansion of the function $1/Ku$. The rough estimation of $\overline{\gamma_{D}}$ and $\underline{\gamma_{D}}$ shown in (\ref{equ3.23}) depends on the mean and variance of the unknown blurring image $Ku$. However, in general, we have $\textrm{Var}(Ku)\ll E^{2}(Ku)$. Therefore, $\overline{\gamma_{D}}$ and $\underline{\gamma_{D}}$ can be simply approximated by
\[
1/E(f)~~ \textrm{or}~~ 1/E(f)(1+\textrm{Var}(f)/E^{2}(f))
\]
due to $E(f)=E(Ku)$, which implies that larger $\delta$ is demanded for images with smaller image intensity (corresponding to images with higher noise level).

Besides, in the proof of the convergence of the proposed IADMNDA algorithm, we demand that $\max\limits_{k}(\delta^{k+1}-\delta^{k})\leq \underline{\gamma_{D}}$ and $\delta^{k}\geq \overline{\gamma_{D}}$ for any $k\in \mathbb{N}$. This condition can be satisfied by modifying the update formula of $\delta^{k}$ as follows:
\begin{equation}\label{equ3.24e}
\delta^{k}=\min\left\{\delta^{k-1}+\underline{\gamma_{D}}, \max\{\tilde{\delta}^{k}, \overline{\gamma_{D}}\}, M\right\}
\end{equation}
where $\tilde{\delta}^{k}$ is computed by the formula (\ref{equ3.6}) in the $k$-th iteration, and $M$ is a large positive number. However, in our experiments we observe that the IADMNDA algorithm still converges without the monotone decreasing condition $\max\limits_{k}(\delta^{k+1}-\delta^{k})\leq \underline{\gamma_{D}}$. In fact, the IADMNDA algorithm using the new update strategy in (\ref{equ3.24e}) cannot obviously improve the convergence speed compared with the counterpart using the original update strategy in (\ref{equ3.6}).

\section{Numerical examples}\label{sec4}
\setcounter{equation}{0}

In this section, we evaluate the performance of the proposed algorithms by numerical experiments on Poissonian image deblurring problem. First, the convergence of the proposed algorithms, which has been investigated in section \ref{sec3} under certain conditions, is further verified through several experiment examples, and meanwhile the influence of the parameter $\delta^{k}$ on the rate of convergence is also investigated. Second, the proposed algorithms are compared with the widely used augmented Lagrangian methods for Poisson image restoration \cite{TIP:PIDAL} and the recently proposed PLAD algorithm \cite{SIAMJSC:Linearized}, which can also be understood from the view of the linearized PDHG algorithm.

\begin{figure}
  \centering
  \subfigure[]{
    \label{fig4.1:subfig:a} 
    \includegraphics[width=1.3in,clip]{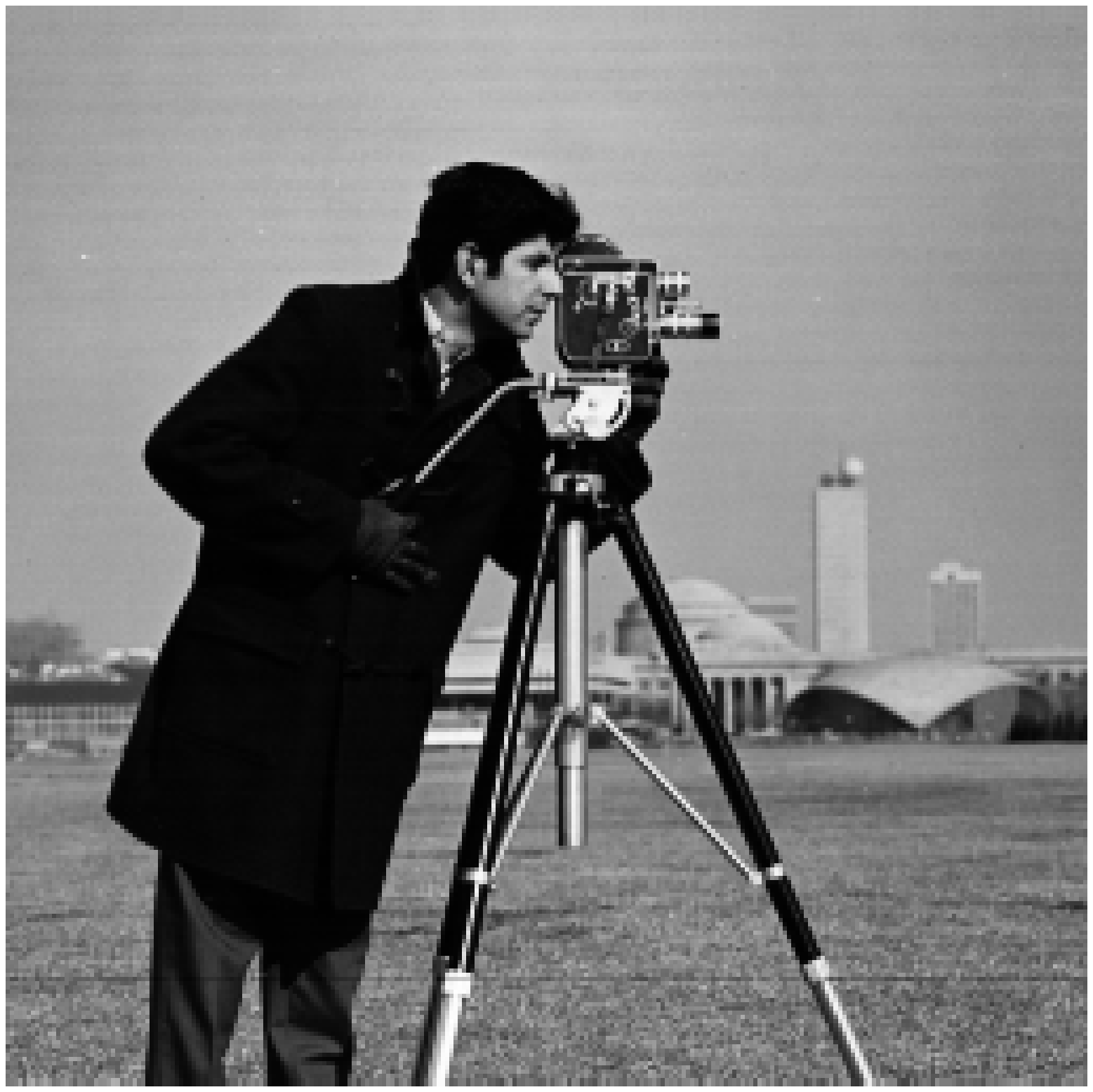}}
  \hspace{0pt}
  \subfigure[]{
    \label{fig4.1:subfig:b} 
    \includegraphics[width=1.3in,clip]{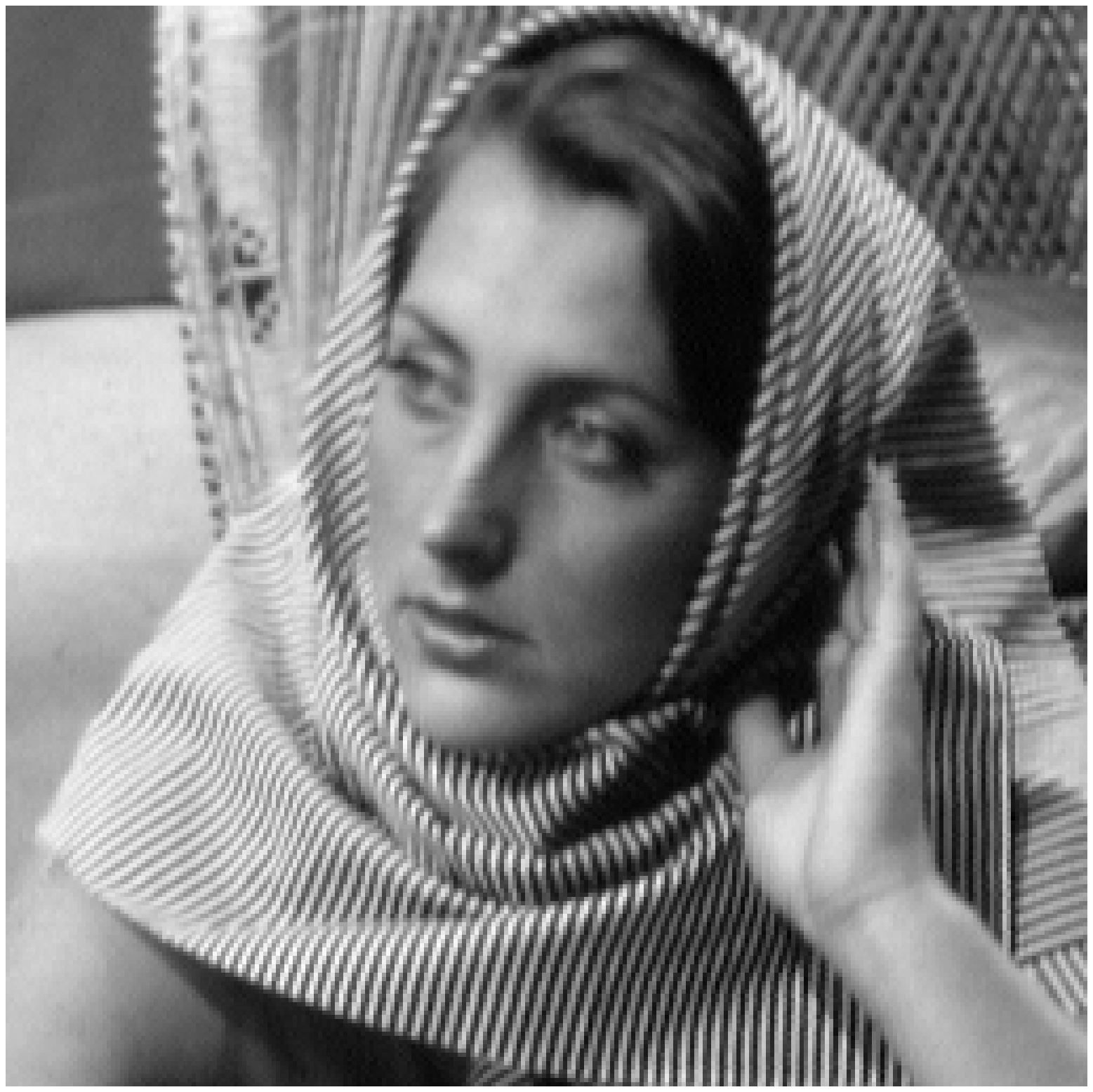}}
  \subfigure[]{
    \label{fig4.1:subfig:c} 
    \includegraphics[width=1.3in,clip]{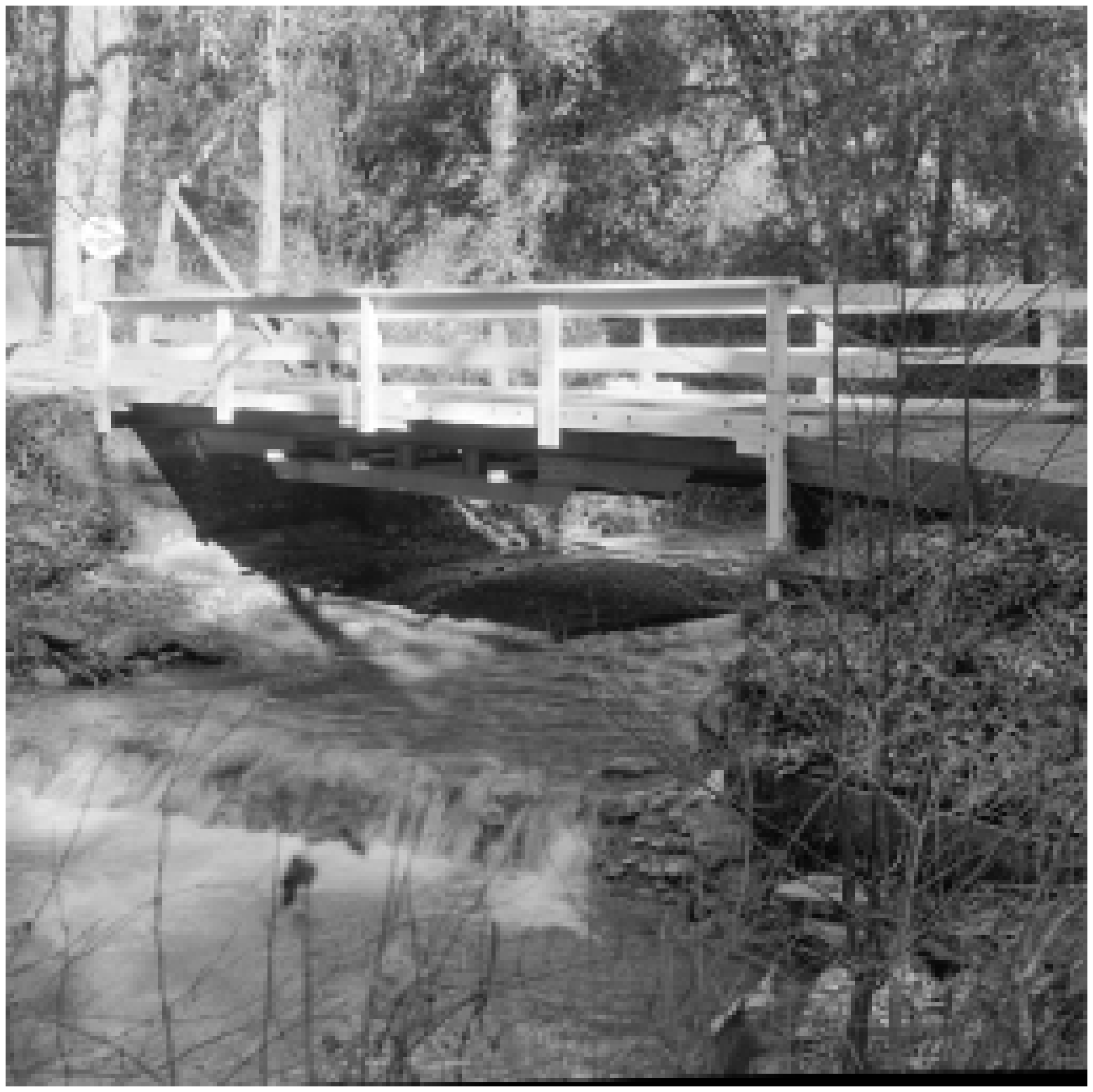}}  \\
  \hspace{0pt}
  \subfigure[]{
    \label{fig4.1:subfig:d} 
    \includegraphics[width=1.3in,clip]{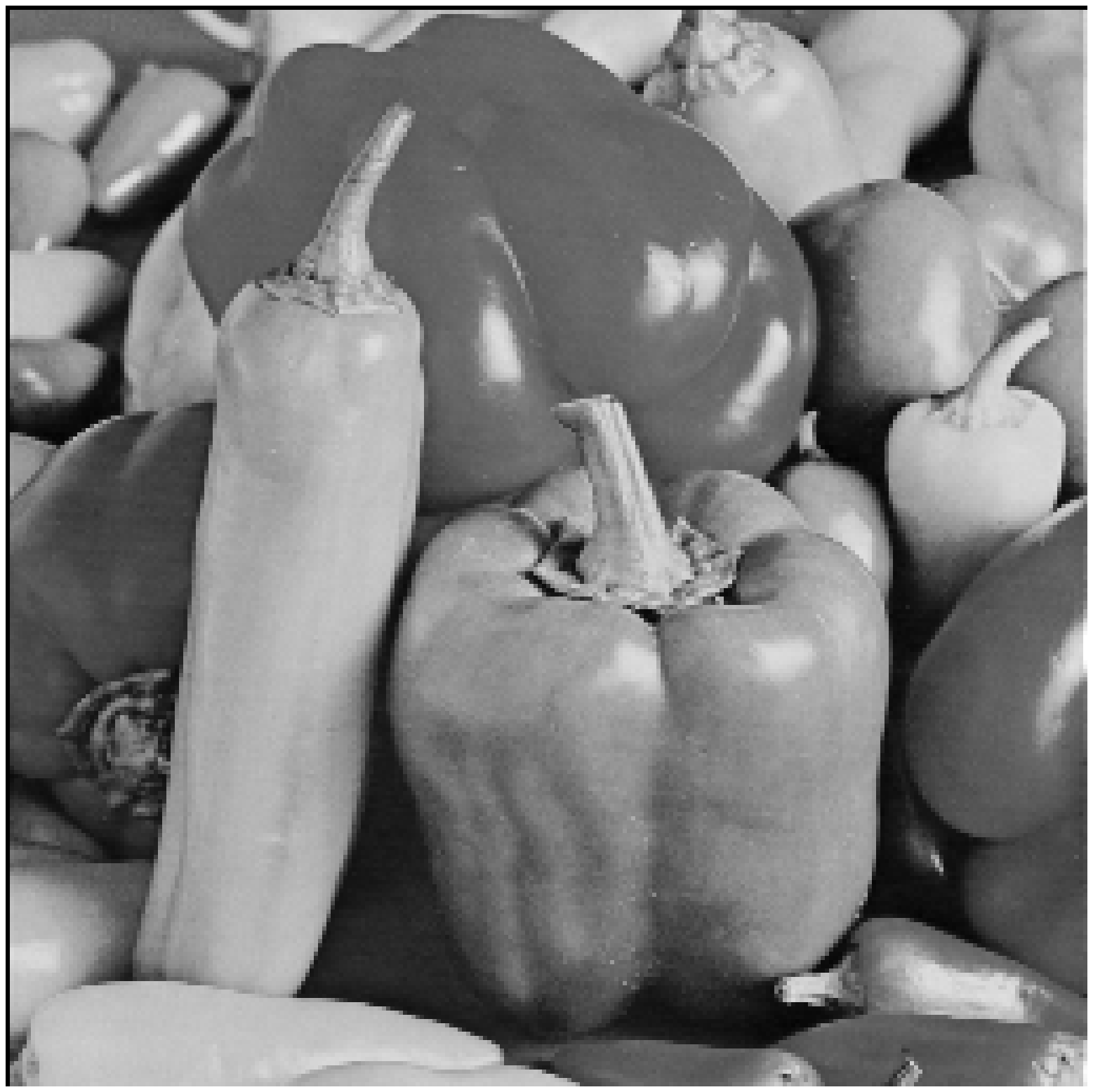}}
  \hspace{0pt}
  \subfigure[]{
    \label{fig4.1:subfig:e} 
    \includegraphics[width=1.3in,clip]{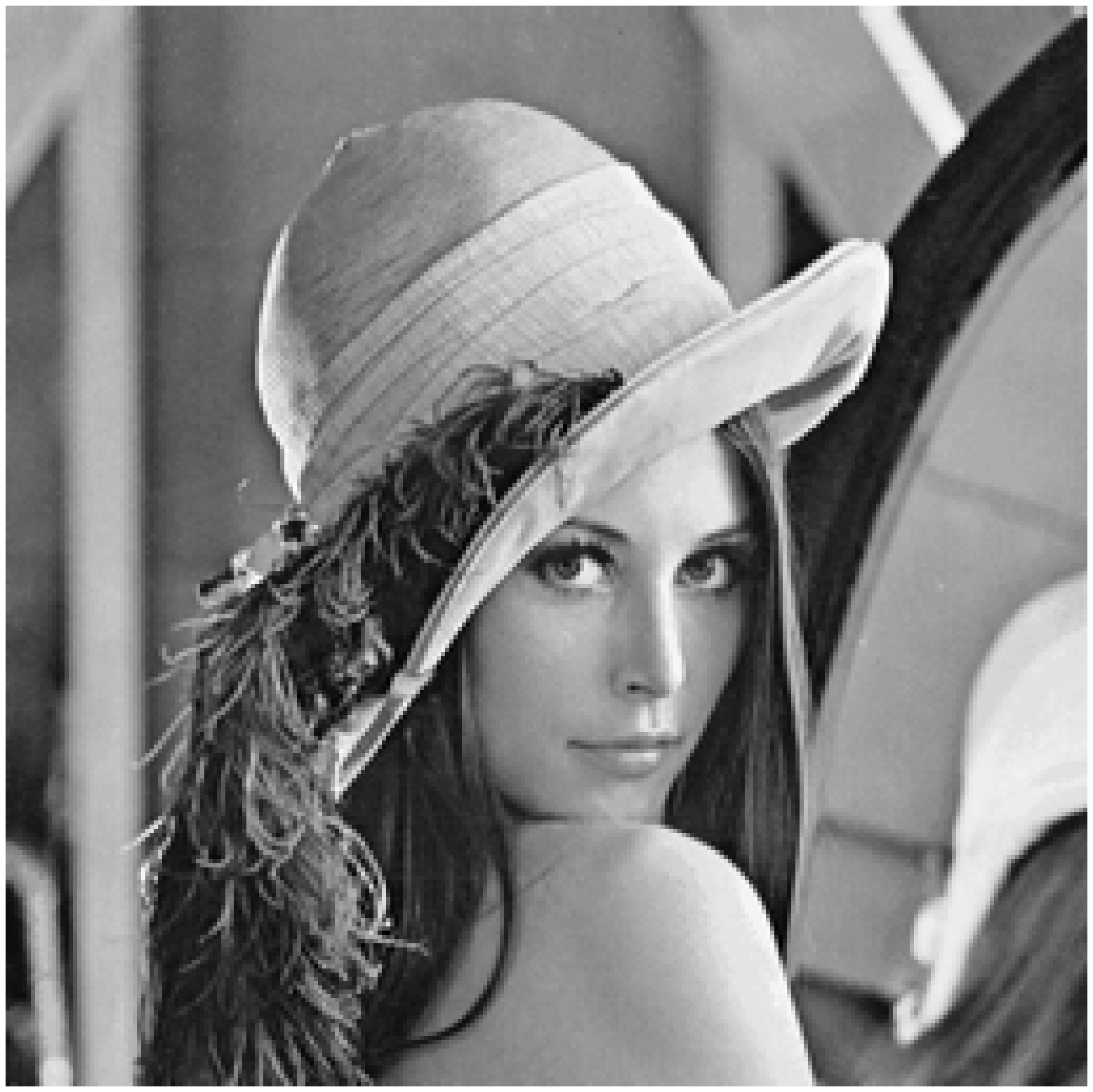}}
  \hspace{0pt}
  \subfigure[]{
    \label{fig4.1:subfig:f} 
    \includegraphics[width=1.3in,clip]{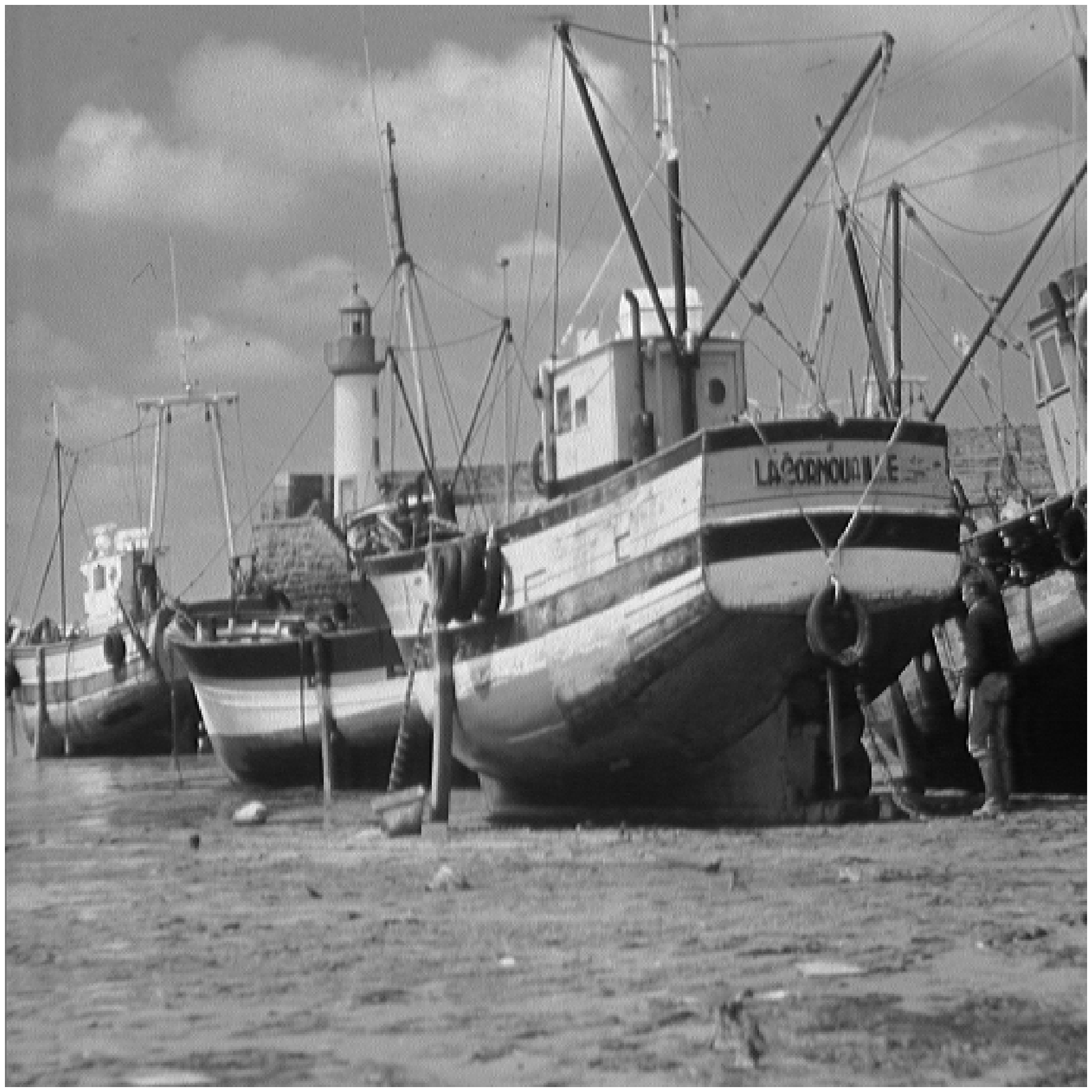}}
\caption{Original images. (a) Cameraman ($256\times 256$), (b) Barbara ($256\times 256$), (c) Bridge ($256\times 256$), (d) Pepper ($256\times 256$), (e) Lenna ($256\times 256$), (f) Boat ($512\times 512$).
}
\label{fig4.1}
\end{figure}

The codes of proposed algorithms and methods used for comparison are written entirely in Matlab, and all the numerical examples are implemented under Windows XP and MATLAB 2009 running on a laptop with an Intel Core i5 CPU (2.8 GHz) and 8 GB Memory. In the following experiments, six standard nature images (see Figure \ref{fig4.1}), which consist of complex components in different scales and with different patterns, are used for our test. Among them, the size of the Boat image is $512\times 512$, and the size of other images is $256\times 256$.

\subsection{The evaluation of the performance of the proposed algorithms}\label{subsec4.1}

In the proposed IADMNDA algorithm, there are two parameters needed to be manually adjusted. One is the regularization parameter $\lambda$, the other is the penalty parameter $\alpha$. It is well-known that $\lambda$ is decided by the noise level, and the value of $\alpha$ does influence the convergence speed of the proposed algorithm. Here we use the strategy similarly to that adopted in \cite{TIP:PIDAL} to choose $\alpha$, i.e., we set $\alpha=20\lambda/I_{\max}$, where $I_{\max}$ denotes the maximum intensity of the original image. Moreover, for the step parameter $\omega^{k}$, we choose it to be the largest value to satisfy the condition (\ref{equ3.3rev2}).    

\begin{figure}
  \centering
  \subfigure[]{
    \label{fig4.2:subfig:a} 
    \includegraphics[width=2.4in,clip]{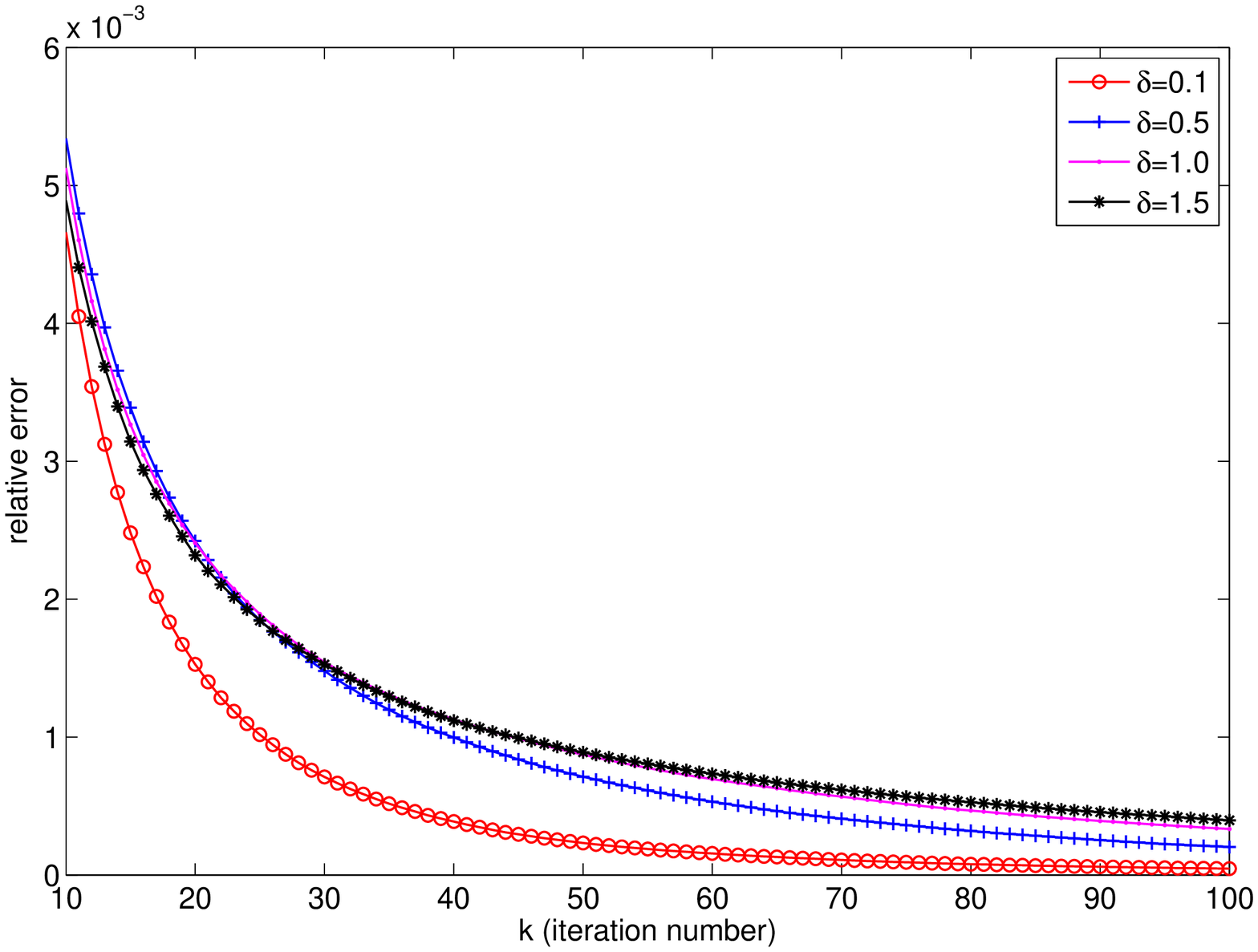}}
  \subfigure[]{
    \label{fig4.2:subfig:b} 
    \includegraphics[width=2.4in,clip]{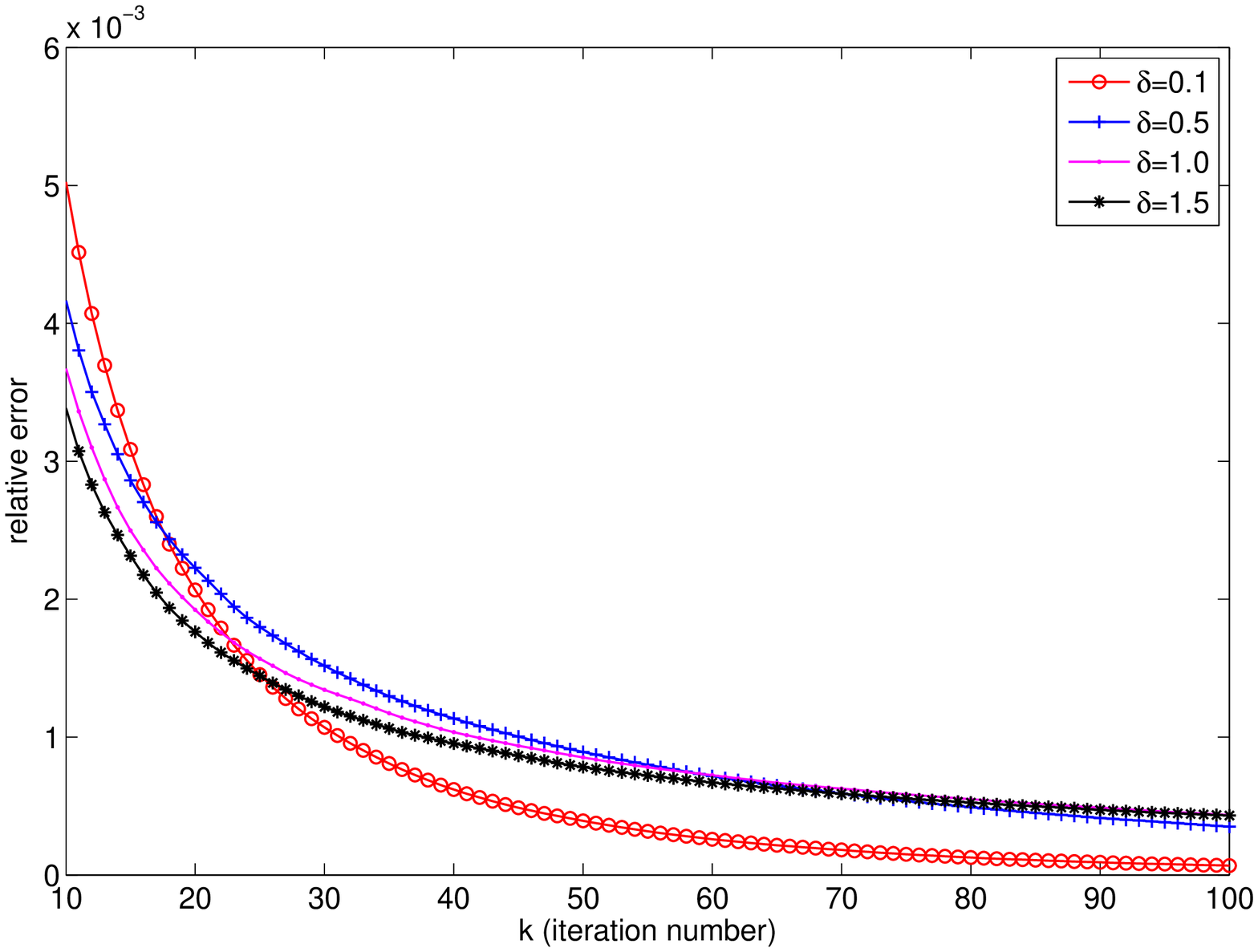}}
  \subfigure[]{
    \label{fig4.2:subfig:c} 
    \includegraphics[width=2.4in,clip]{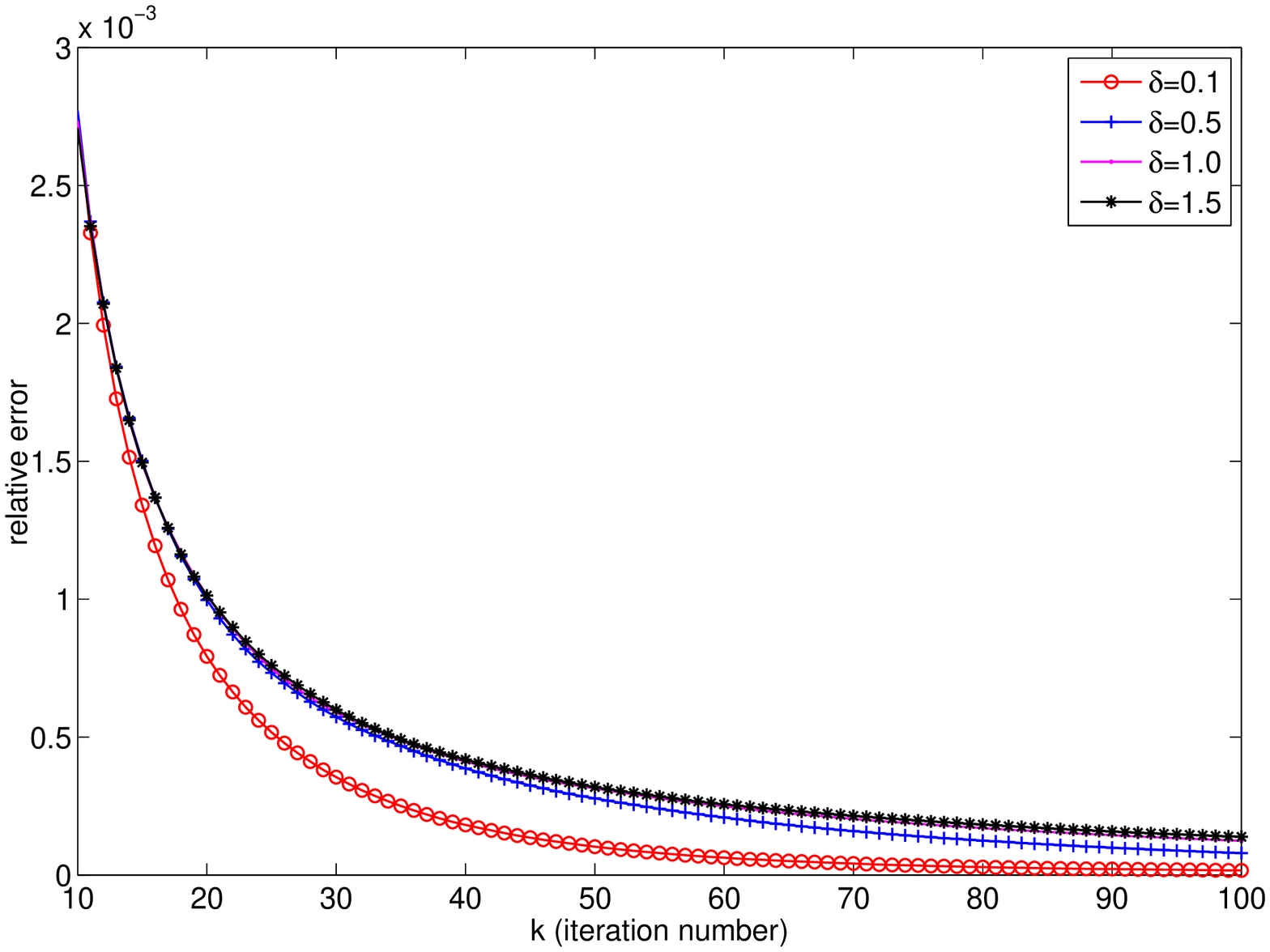}}
  \subfigure[]{
    \label{fig4.2:subfig:d} 
    \includegraphics[width=2.4in,clip]{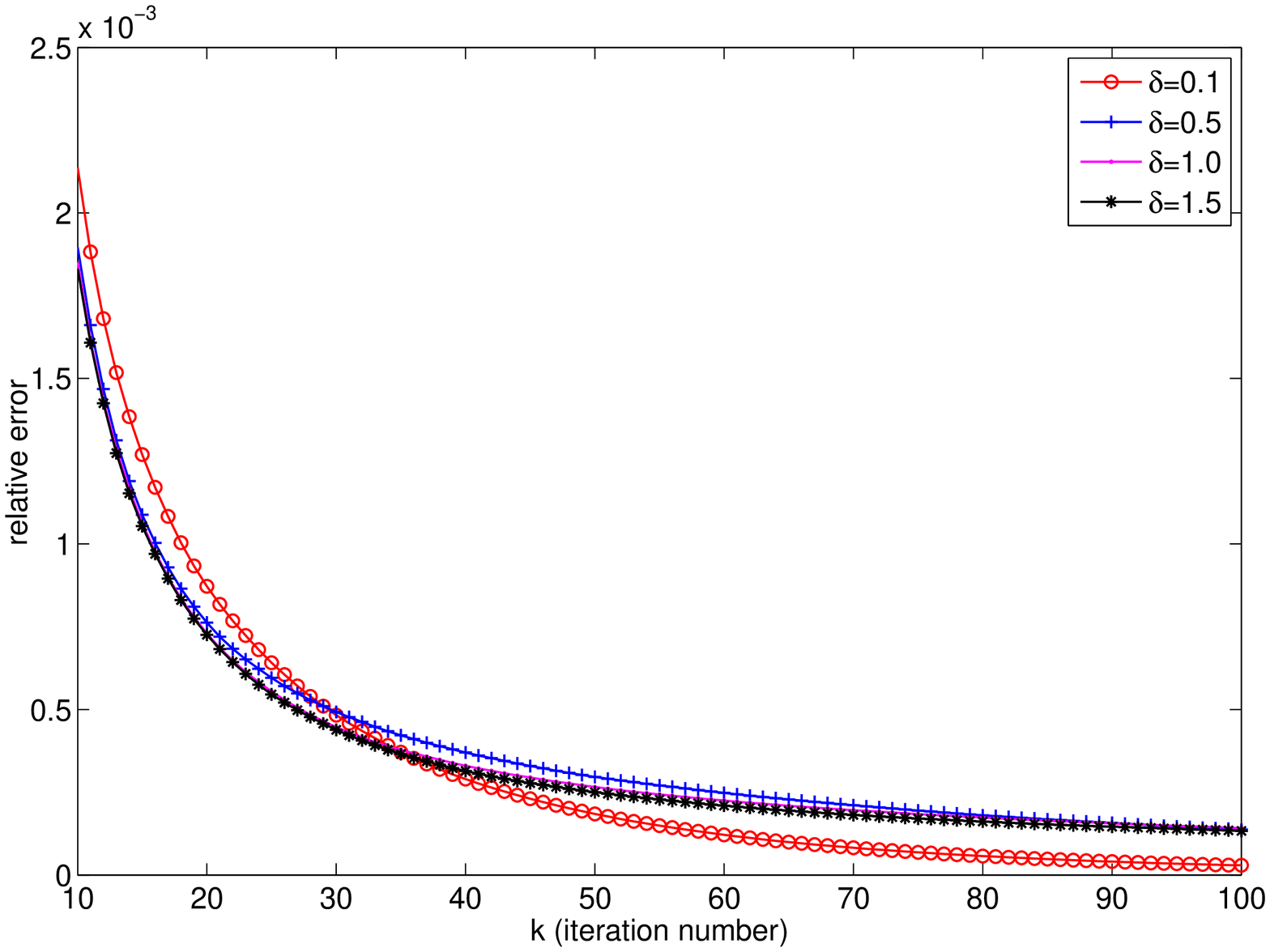}}
\caption{The evolution curves of the relative error $\frac{\|u^{k}-u^{k-1}\|_{2}}{\|u^{k-1}\|_{2}}$ for images with different blur kernels and noise levels. (a) Cameraman image with $I_{\max}=200$: $7\times 7$ uniform blur kernel; (b) Cameraman image with $I_{\max}=400$: $7\times 7$ uniform blur kernel; (c) Barbara image with $I_{\max}=200$: $9\times 9$ Gaussian blur kernel with standard deviation $1$; (d) Barbara image with $I_{\max}=400$: $9\times 9$ Gaussian blur kernel with standard deviation $1$.
}
\label{fig4.2}
\end{figure}

In what follows, we further investigate influence of the parameter $\delta$ on the rate of convergence of proposed IADMND algorithm. Two images named ``Cameraman" and ``Barbara" (see Figure \ref{fig4.1}) are used for the test. Here we consider two types of blur effects with different levels of Poisson noise:
the Cameraman image is scaled to a maximum value of $200$ and $400$ respectively, and blurred with a
$7\times 7$ uniform blur kernel; the Barbara image is scaled to the same range and convoluted by a $9\times 9$ Gaussian kernel of unit variance. Then the blurred images are contaminated by Poisson noise. Figure \ref{fig4.2} depicts the evolution curves of the relative error $\frac{\|u^{k}-u^{k-1}\|_{2}}{\|u^{k-1}\|_{2}}$ with different $\delta$ values.
It is observed that the value of the parameter $\delta$ does influence the convergence speed. Generally speaking, the proposed algorithm with a small $\delta$ converges faster. However, if $\delta$ is too small, the convergence cannot be guaranteed. In our experiments, we observe that $\delta=0.1$ is not suitable for images with $I_{\max}\leq 100$, and in this case the IADMND algorithm become unstable. This is also consistent with the analysis in section \ref{subsec3.2}. Besides, the plots in Figure \ref{fig4.2} implicitly verify that the convergence of the proposed IADMND algorithm is really guaranteed with suitable values of $\delta$.

\begin{figure}
  \centering
  \subfigure[]{
    \label{fig4.3:subfig:a} 
    \includegraphics[width=2.5in,clip]{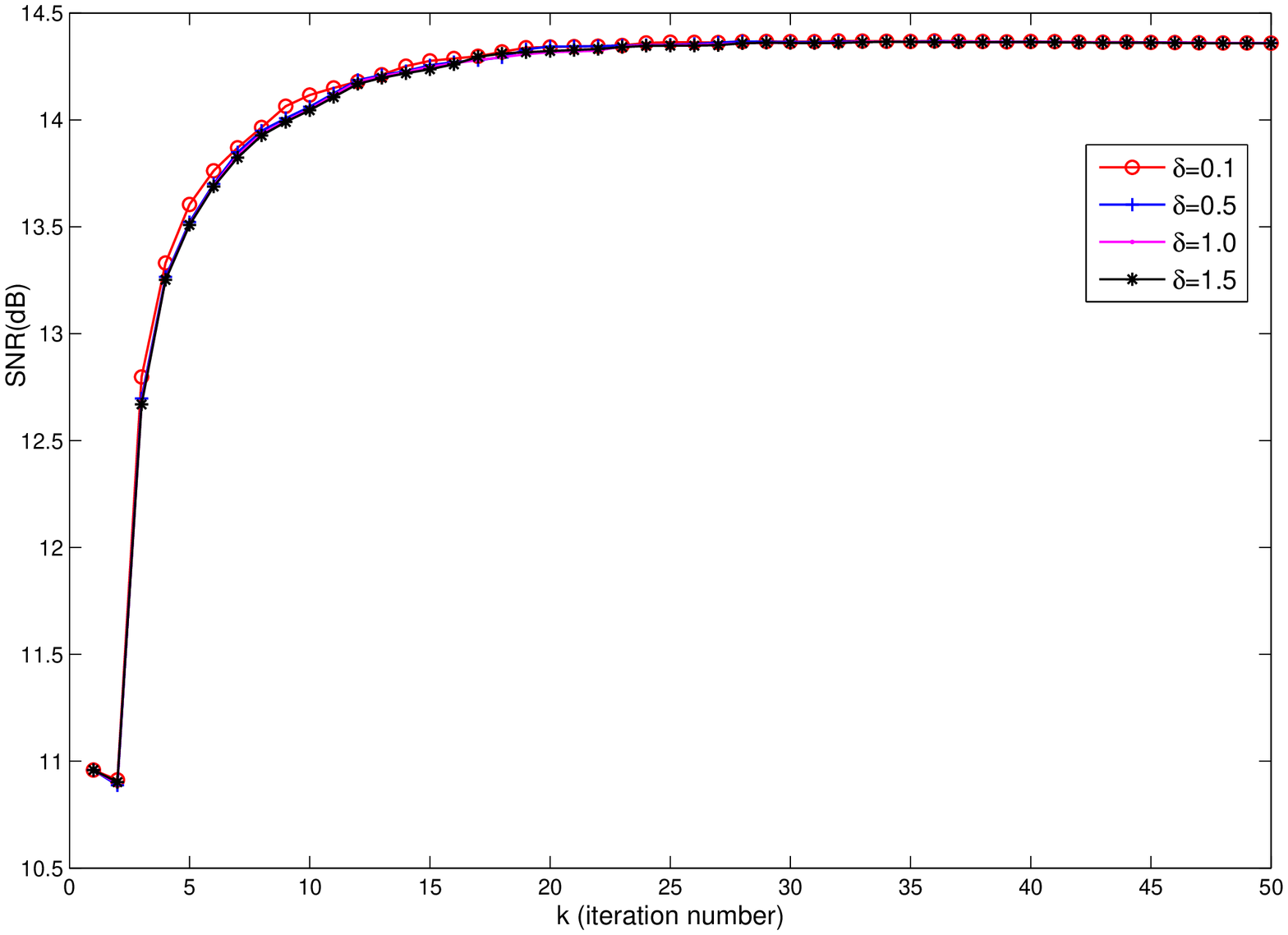}}
  \subfigure[]{
    \label{fig4.3:subfig:b} 
    \includegraphics[width=2.5in,clip]{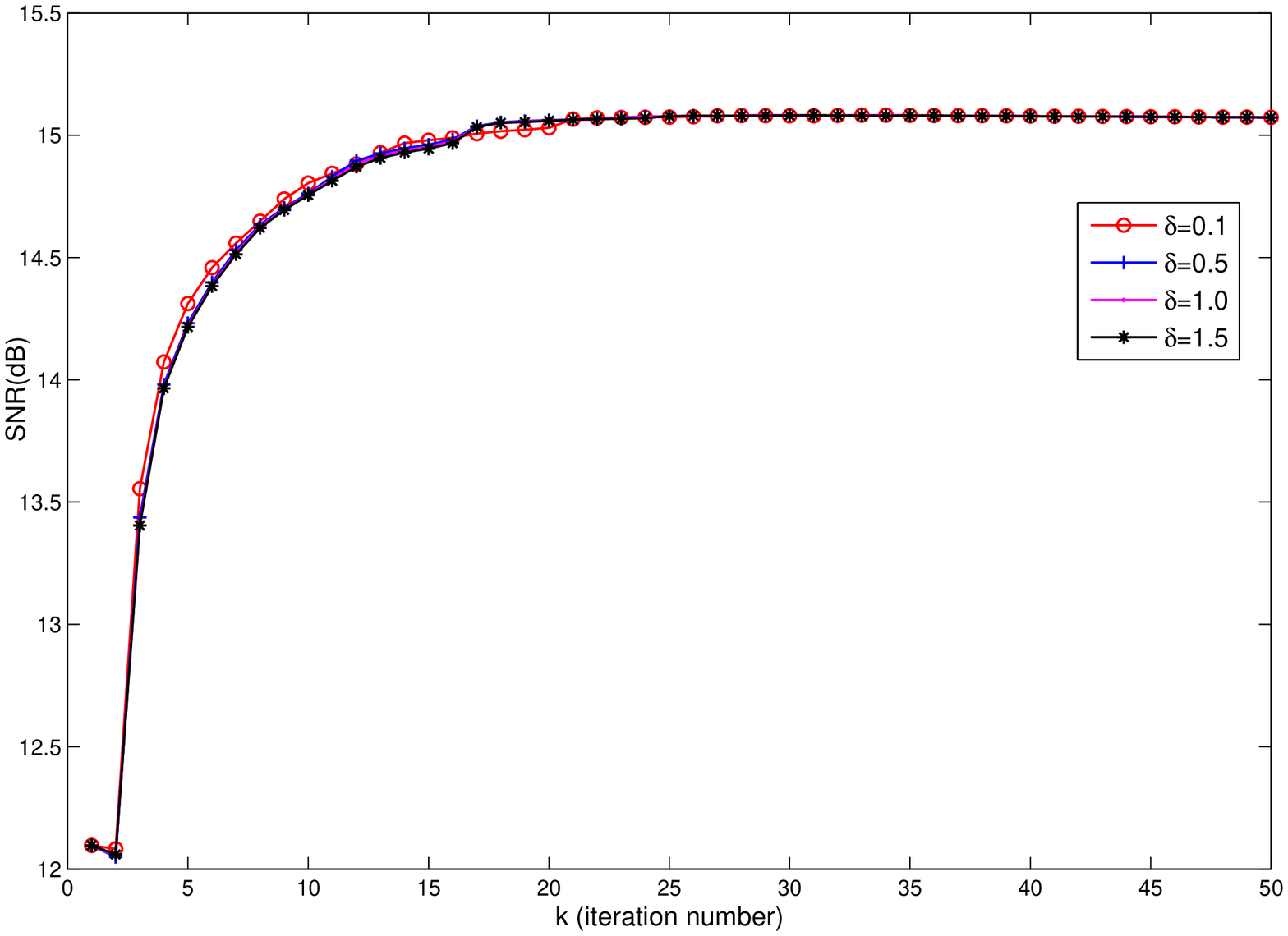}}
  \subfigure[]{
    \label{fig4.3:subfig:c} 
    \includegraphics[width=2.5in,clip]{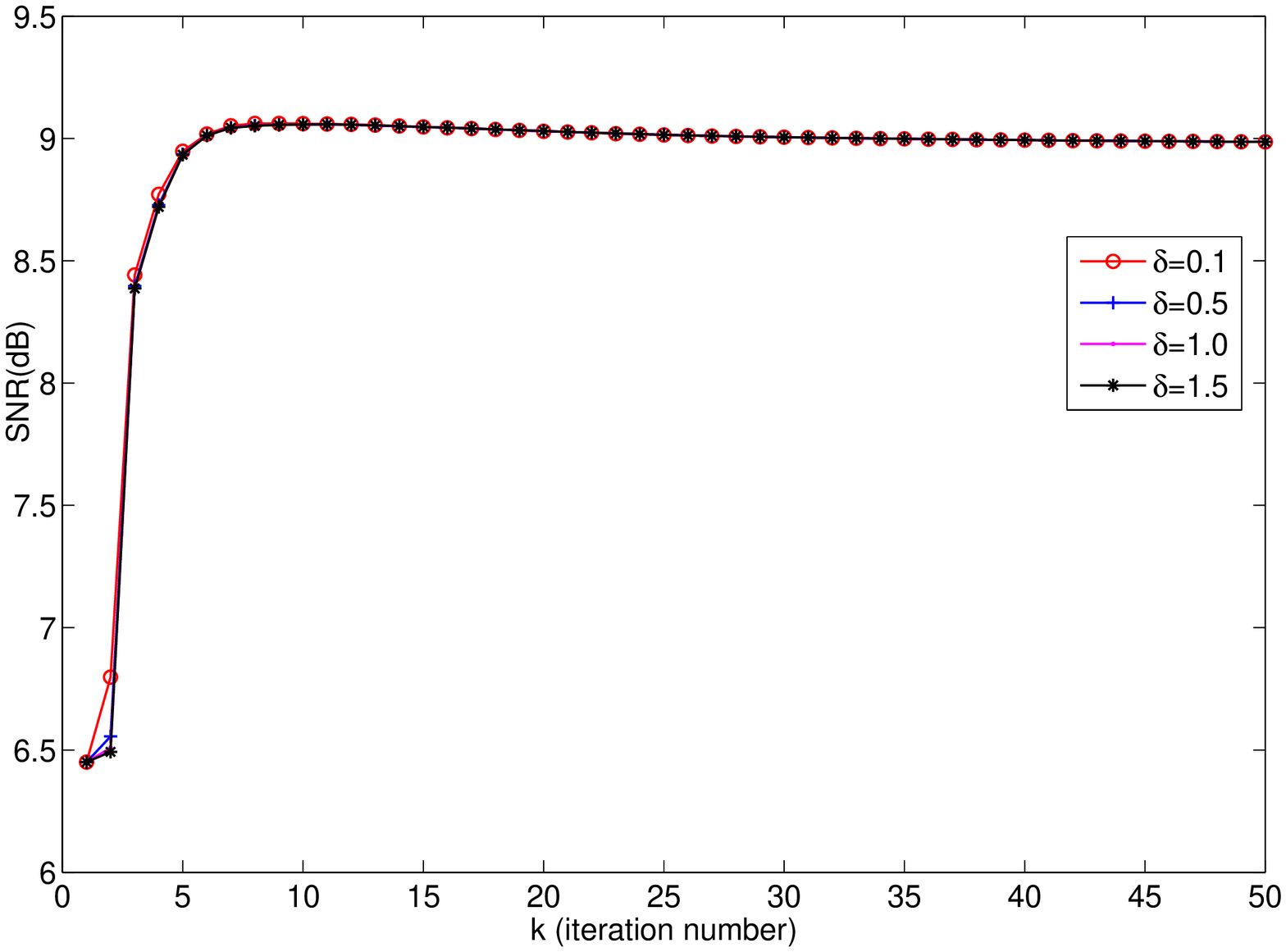}}
  \subfigure[]{
    \label{fig4.3:subfig:d} 
    \includegraphics[width=2.5in,clip]{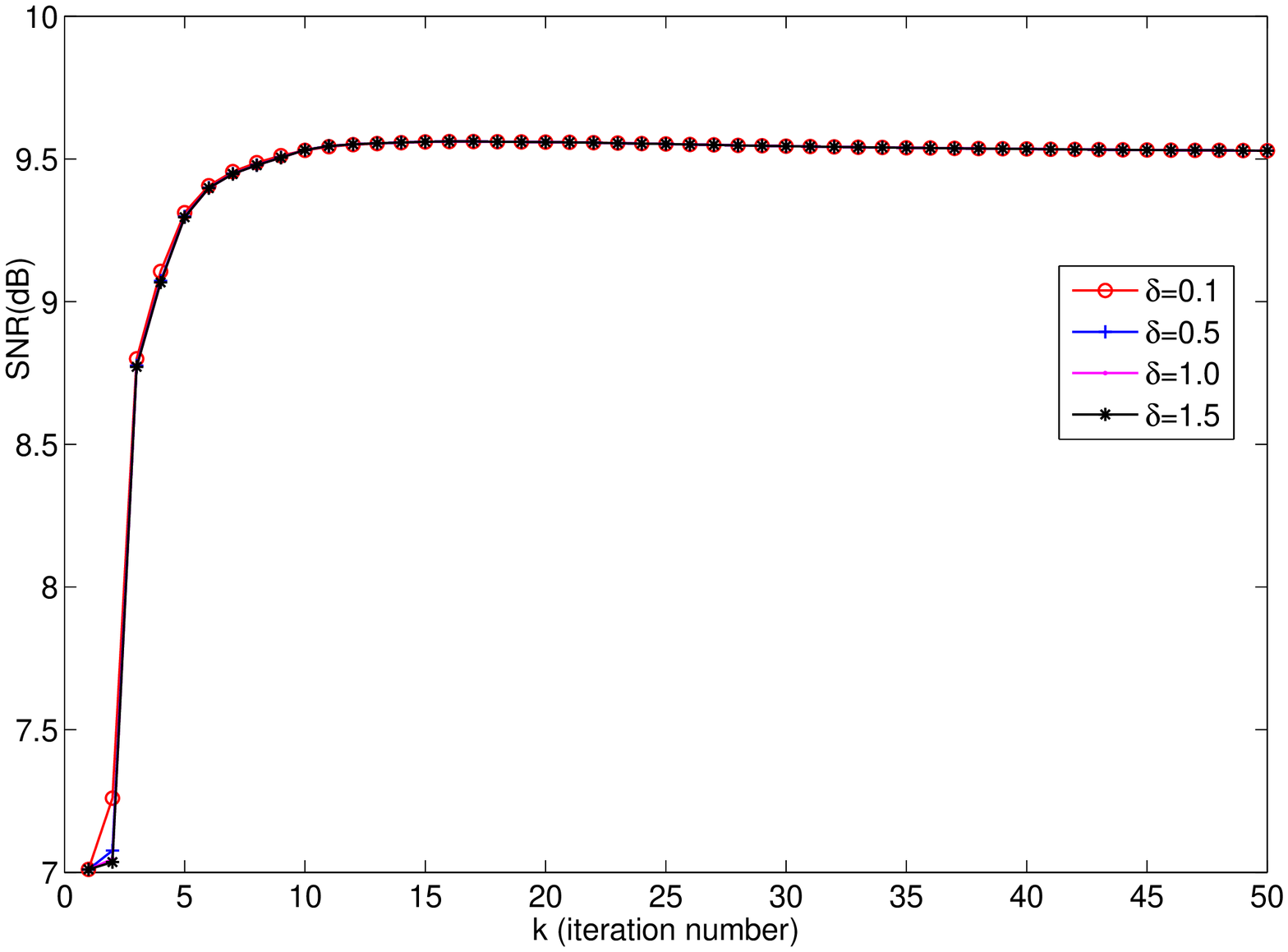}}
\caption{The evolution curves of SNR (dB) for images with different blur kernels and noise levels. (a) Cameraman image with $I_{\max}=200$: $9\times 9$ Gaussian blur kernel with standard deviation $1$; (b) Cameraman image with $I_{\max}=400$: $9\times 9$ Gaussian blur kernel with standard deviation $1$; (c) Bridge image with $I_{\max}=200$: $7\times 7$ uniform blur kernel; (d) Bridge image with $I_{\max}=400$: $7\times 7$ uniform blur kernel.
}
\label{fig4.3}
\end{figure}

In the proposed IADMNDA algorithm, we use the formula (\ref{equ3.6}) to update the parameter $\delta^{k}$ in each iteration, and hence we further discuss the selection of the initial value $\delta=\delta_{0}$. Two images called ``Cameraman" and ``Bridge" (see Figure \ref{fig4.1}) are adopted here. Figure \ref{fig4.3} shows the evolution curves of SNR (signal-to-noise ratio) values with different $\delta_{0}$. Note that there is almost no difference between the results with different values of $\delta_{0}$, except the SNR values in the first several iteration steps. Therefore, we set $\delta_{0}$ to be a fixed constant in the following experiments.


Finally, we compare the performance of the IADMNDA algorithms which use the update formulas (\ref{equ3.6}) and (\ref{equ3.24e}) for $\delta^{k}$ respectively. In the formula (\ref{equ3.24e}), the values of $\underline{\gamma_{D}}$ and $\overline{\gamma_{D}}$ are estimated by
\[
\frac{1}{E(f)}\left(1+\frac{\textrm{Var}(f)}{E^{2}(f)}\right).
\]
Table \ref{tab4.3} lists the SNR values and the iteration number of the IADMNDA algorithms with different update formulas for $\delta^{k}$. Here we use the parameters setting in Table \ref{tab4.1} for the IADMNDA algorithms,
and the stopping criterion is defined such that the relative error is below some small constant, i.e.,

\begin{equation}\label{equ4.2}
\|u^{k+1}-u^{k}\|_{2}\leq\epsilon\|u^{k}\|_{2}.
\end{equation}
Here we choose $\epsilon=2\times 10^{-4}$. 
In this table, the serial numbers ``1" and ``2" denote the results of the IADMNDA algorithms with the update formulas (\ref{equ3.6}) and (\ref{equ3.24e}) respectively, and $(\cdot, \cdot )$ denotes the SNR values, iteration numbers in sequence. It is observed that the performances of algorithms with both update formulas are almost the same. Therefore, in the following compared experiments, we use the update formulas (\ref{equ3.6}) for the IADMNDA algorithm.

\begin{table} [htbp]
\centering \caption{The SNR (dB) and iteration number with different update formulas for $\delta^{k}$.}
\scalebox{0.8}{
\begin{tabular}{c|c|c|c|c|c|c|c|c|c}
  \hline
  \multicolumn{10}{c}{\textbf{$9\times 9$ Gaussian kernel}} \\
  \hline
  Image & \multicolumn{3}{c|}{Barbara} & \multicolumn{3}{c|}{Bridge} &  \multicolumn{3}{c}{Boat} \\
  \hline
  I_{\max} & 100 & 200 &  500 & 100 & 200 &  500 & 100 & 200 &  500 \\
  \hline
  1 &  (9.92, 38) & (10.54, 33) &  (12.22, 31) &  (10.51, 45) &  (11.26, 41) & (12.08,35) & (12.75,40) &  (13.50, 37)  & (14.44, 33)\\
  \hline
  2 & (9.92, 37) & (10.54, 32) &  (12.22, 32) &  (10.51, 45) &  (11.26, 40) & (12.08,35) & (12.75,40) &  (13.49, 38)  & (14.45, 32)\\
  \hline\hline
  \multicolumn{10}{c}{\textbf{$7\times 7$ uniform blur}} \\
  \hline
  Image & \multicolumn{3}{c|}{Barbara} & \multicolumn{3}{c|}{Bridge} &  \multicolumn{3}{c}{Boat} \\
  \hline
  I_{\max} & 100 & 200 &  500 & 100 & 200 &  500 & 100 & 200 &  500 \\
  \hline
  1 &  (9.02, 45) & (9.17, 38) &  (9.55, 35) &  (8.43, 51) &  (8.99, 46) & (9.64,42) & (10.42,46) &  (10.90, 43)  & (11.74, 38)\\
  \hline
  2 & (9.02, 44) & (9.17, 38) &  (9.55, 36) &  (8.42, 53) &  (8.99, 47) & (9.64,42) & (10.42,45) &  (10.91, 41)  & (11.74, 40)\\
  \hline
\end{tabular}}
\label{tab4.3}
\end{table}

\subsection{Comparison with the current state-of-art methods}\label{subsec4.2}

In this subsection, we further compare the proposed algorithms with the current state-of-the-art algorithms, including the PIDAL algorithm proposed in \cite{TIP:PIDAL}, and the recently proposed PLAD algorithm \cite{SIAMJSC:Linearized}. Note that several parameters are required to be manually adjusted in the compared algorithms: the regularization parameter $\lambda$, the penalty parameter $\alpha$ for all these algorithms; the step parameter $\delta$ (see (\ref{equ2.6})) for the PLAD algorithm, and the parameter $\delta$ for the proposed IADMND algorithm. Through many trials we use the rules of thumb: $\alpha$ in the PIDAL algorithm is set to $60\lambda/I_{\max}$, while in other algorithms it is chosen to be $20\lambda/I_{\max}$; the initial value $\delta_{0}$ in the proposed IADMNDA algorithm is fixed as $0.1$; the other parameters setting is summarized in Table \ref{tab4.1}, found to guarantee the convergence and achieve satisfactory results. Moreover, a Rudin-Osher-Fatemi (ROF) denoising problem is included in each iteration of the PIDAL algorithm, and it is solved by using a small and fixed number of iterations (just 5) of Chambolle's algorithm. For more details refer to \cite{TIP:PIDAL}.

\begin{table} [htbp]
\centering \caption{The parameter setting for the numerical experiment}
\scalebox{0.9}{
\begin{tabular}{c|c|c|c|c|c|c}
  \hline
    &  \multicolumn{3}{|c}{\textbf{$9\times 9$ Gaussian kernel}} & \multicolumn{3}{|c}{\textbf{$7\times 7$ uniform blur}} \\
  \hline
  $I_{\max}$ & 100 & 200 &  500 & 100 & 200 &  500 \\
  \hline
  $\lambda$ & 0.04 & 0.02 &  0.008 & 0.03 & 0.01 &  0.005 \\
  \hline
  $\delta_{PLAD}$ & 0.15 & 0.15 &  0.03 & 0.15 & 0.05 &  0.02 \\
  \hline
  $\delta_{IADMND}$ & 0.3 & 0.1 &  0.1 & 0.3 & 0.1 &  0.1 \\
  \hline
\end{tabular}}
\label{tab4.1}
\end{table}

In the following numerical experiments, the stopping criterion in (\ref{equ4.2}) is used for all the algorithms here. Table \ref{tab4.2} lists the SNR values, the number of iterations and CPU time of different algorithms for images with different blur kernels and noise levels. In this table, ``Gaussian" and ``Uniform" denote a $9\times 9$ Gaussian kernel of unit variance and a $7\times 7$ uniform blur kernel respectively. The two cases can be seen as examples of mild blur and strong blur. Besides, ``$\cdot/\cdot/\cdot$" denotes the SNR values, iteration numbers and CPU time in sequence. Note that the iteration numbers of the PIDAL algorithm represent the outer iteration numbers.

From the results in Table \ref{tab4.2} we observe that the proposed algorithms are much faster than the PIDAL and PLAD algorithms, and meanwhile the SNR values of the recovered images achieved with the proposed algorithms are comparable to those achieved with the PIDAL and PLAD algorithms. Therefore, it is verified that the strategy of using the proximal Hessian matrix to approximate the second-order derivatives is more efficient than the simple approximation of an identity matrix multiplied by some constant in the PLAD algorithm. It is also noted that the iteration numbers of the IADMNDA algorithm are the least in most cases. However, the update of $\delta^{k}$ generates extra computational cost in the IADMNDA algorithm, which makes its implementation time longer than the IADMND algorithm in some cases.

\begin{figure}
  \centering
  \subfigure[]{
    \label{fig4.4:subfig:a}
    \includegraphics[width=1.5in,clip]{Cameraman.eps}}
  \subfigure[]{
    \label{fig4.4:subfig:b}
    \includegraphics[width=1.5in,clip]{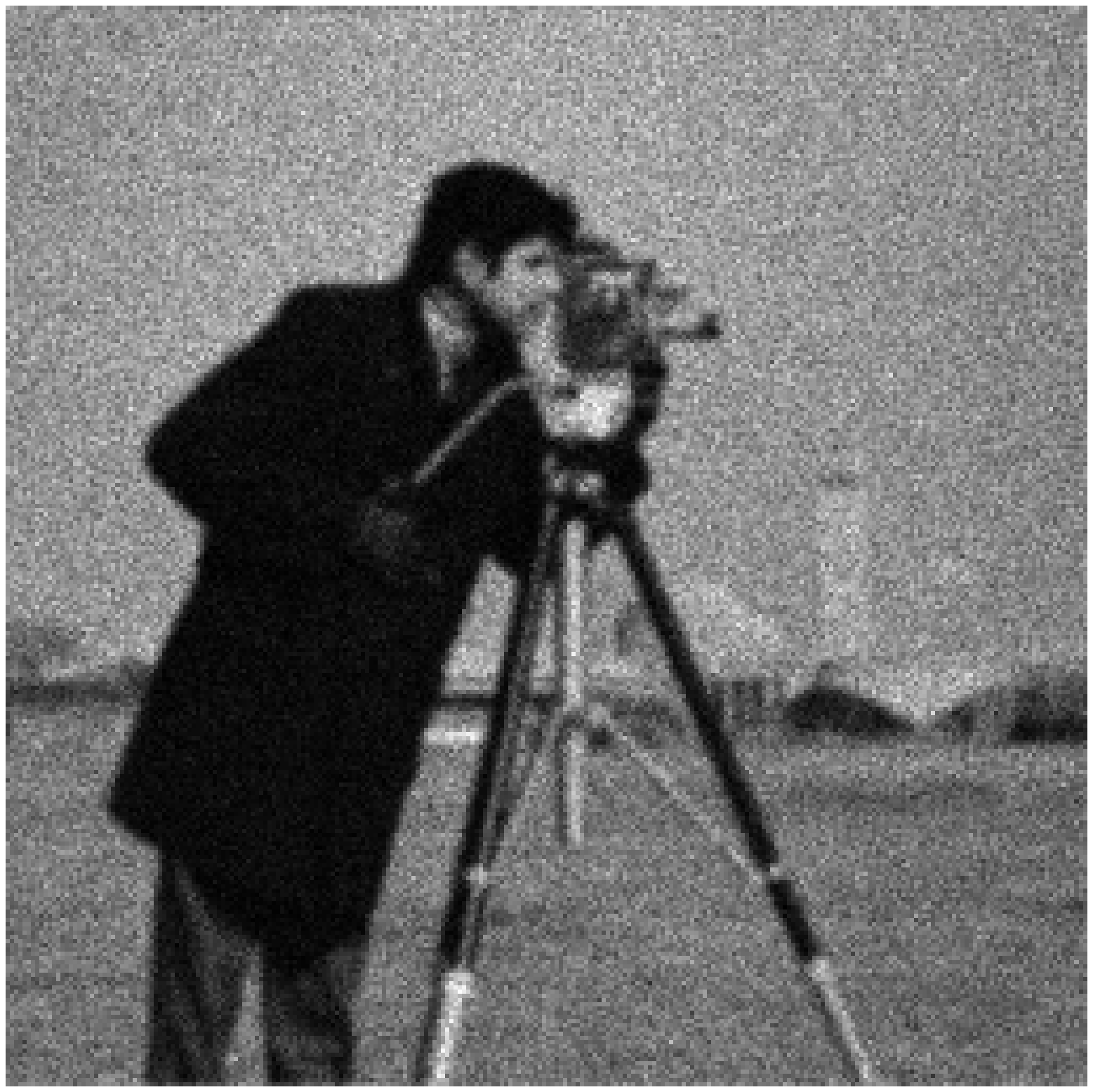}}
  \subfigure[]{
    \label{fig4.4:subfig:c}
    \includegraphics[width=1.5in,clip]{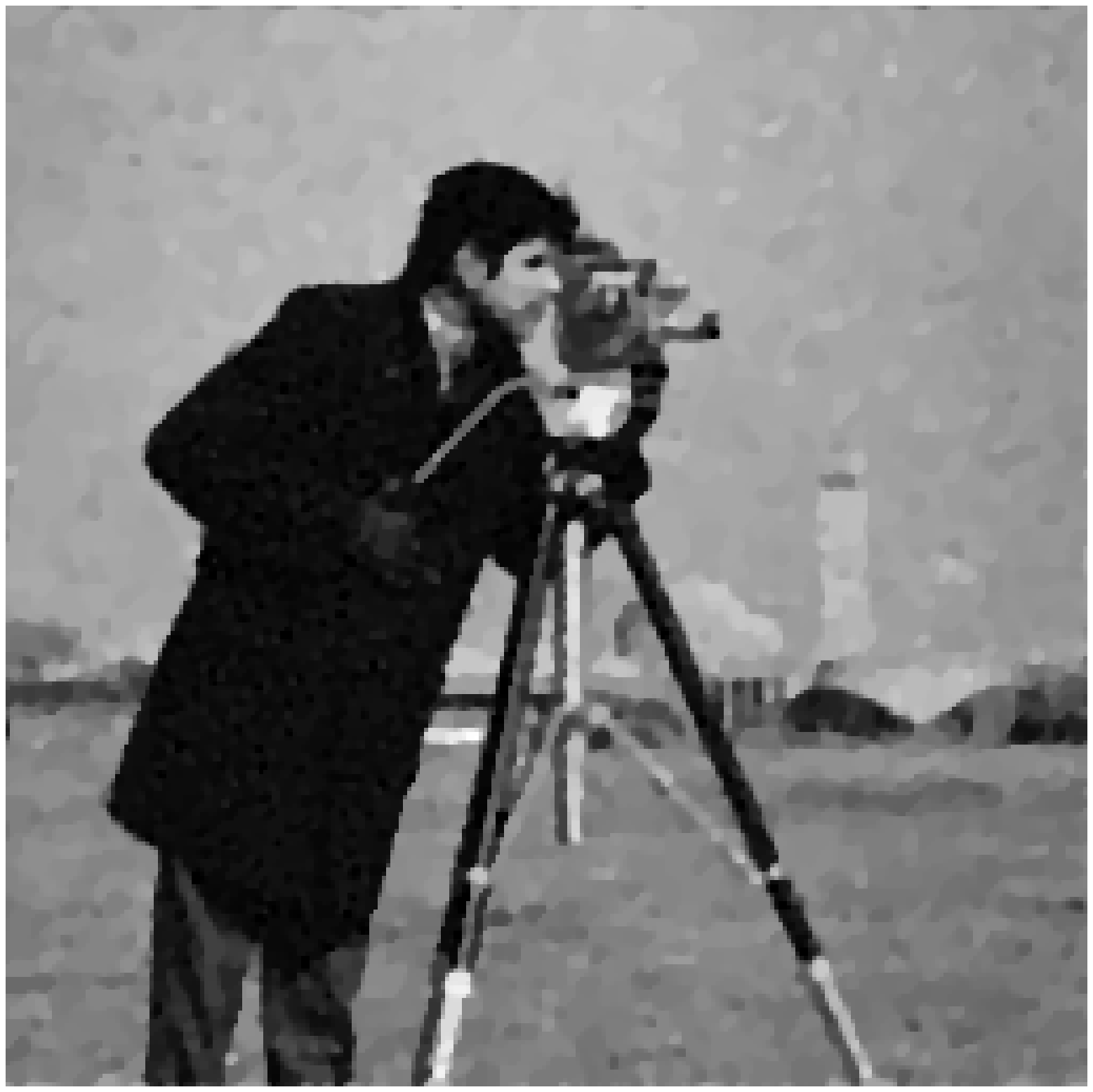}}
  \subfigure[]{
    \label{fig4.4:subfig:d}
    \includegraphics[width=1.5in,clip]{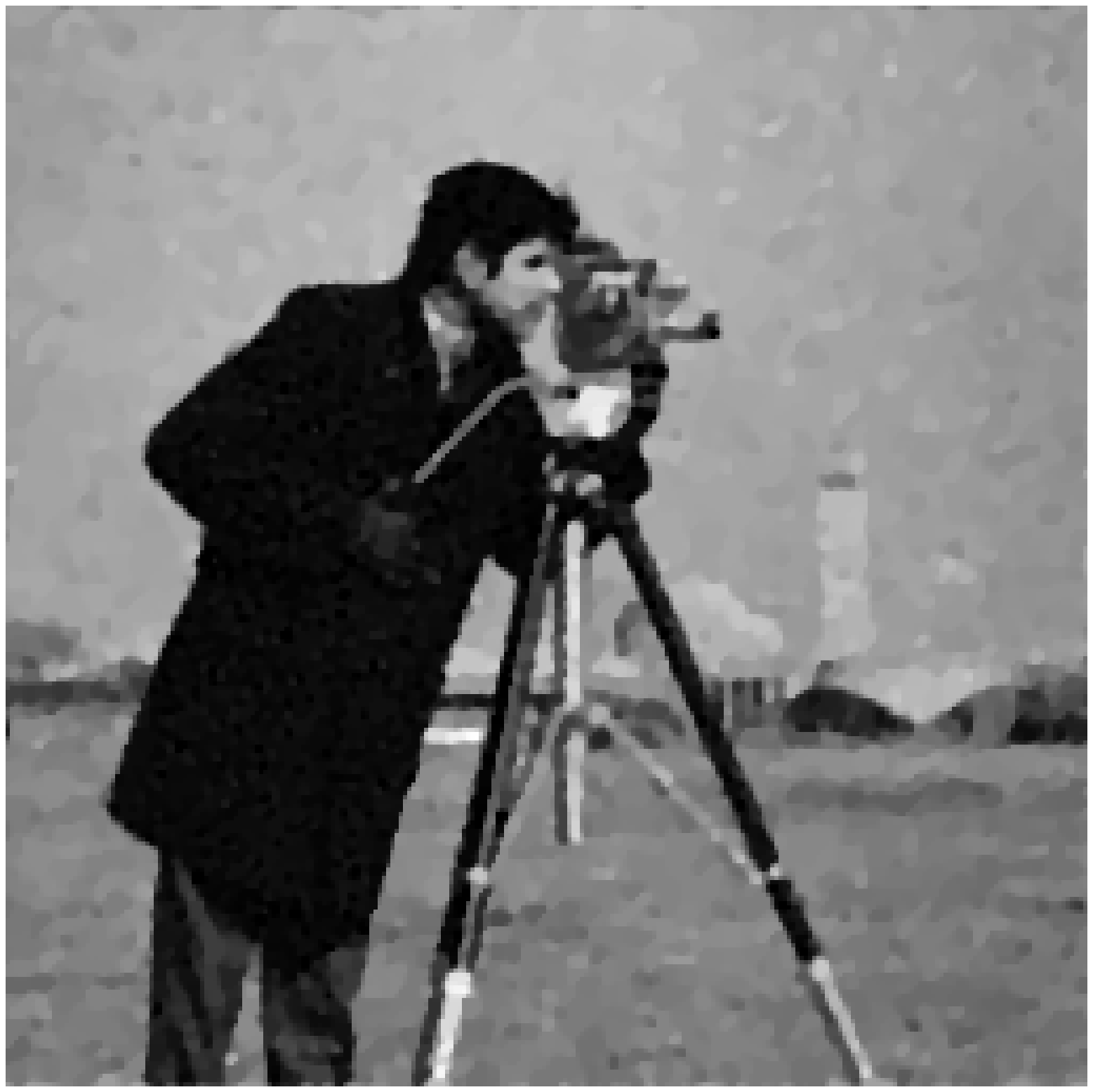}}
  \subfigure[]{
    \label{fig4.4:subfig:e}
    \includegraphics[width=1.5in,clip]{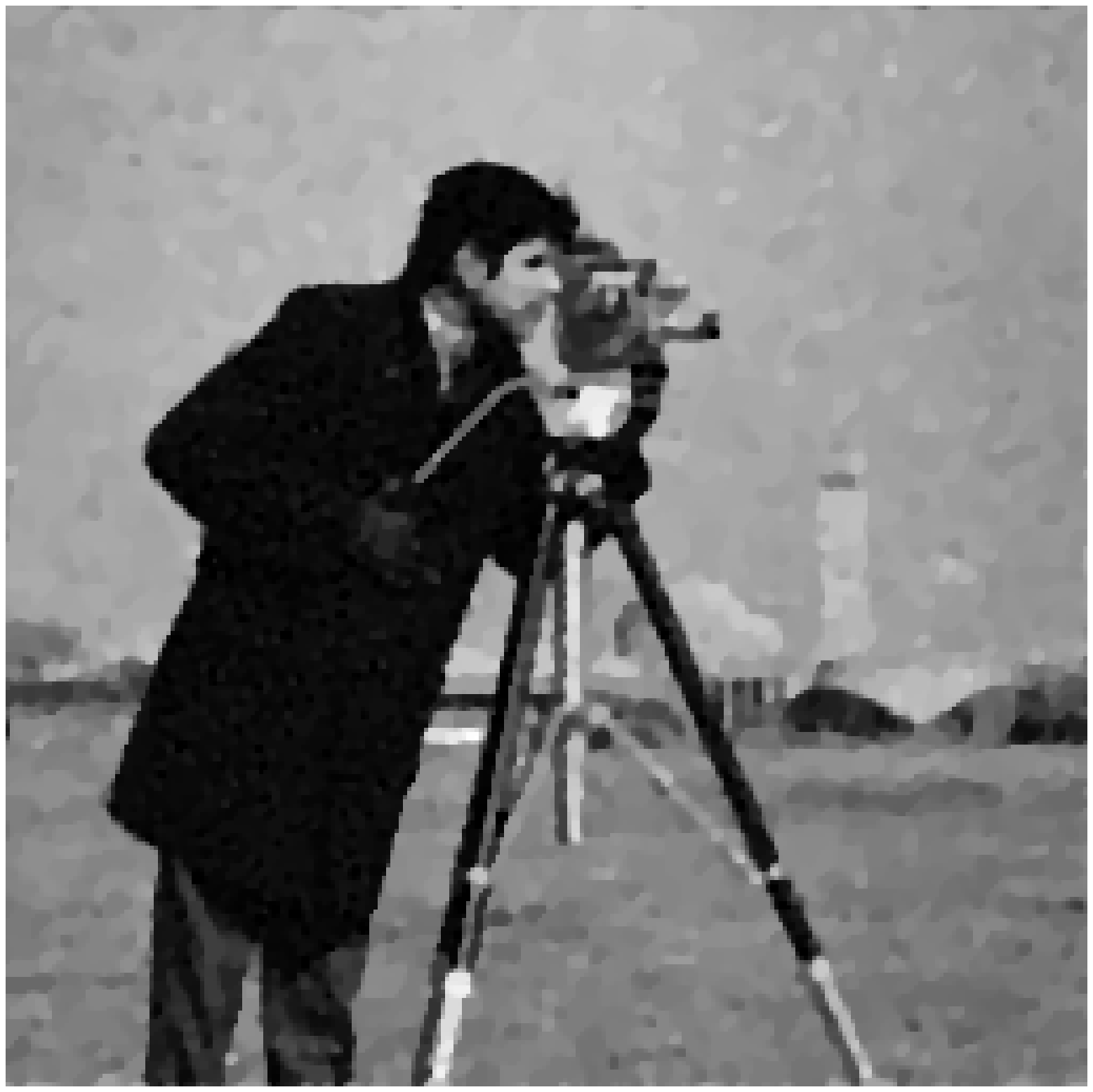}}
  \subfigure[]{
    \label{fig4.4:subfig:f}
    \includegraphics[width=1.5in,clip]{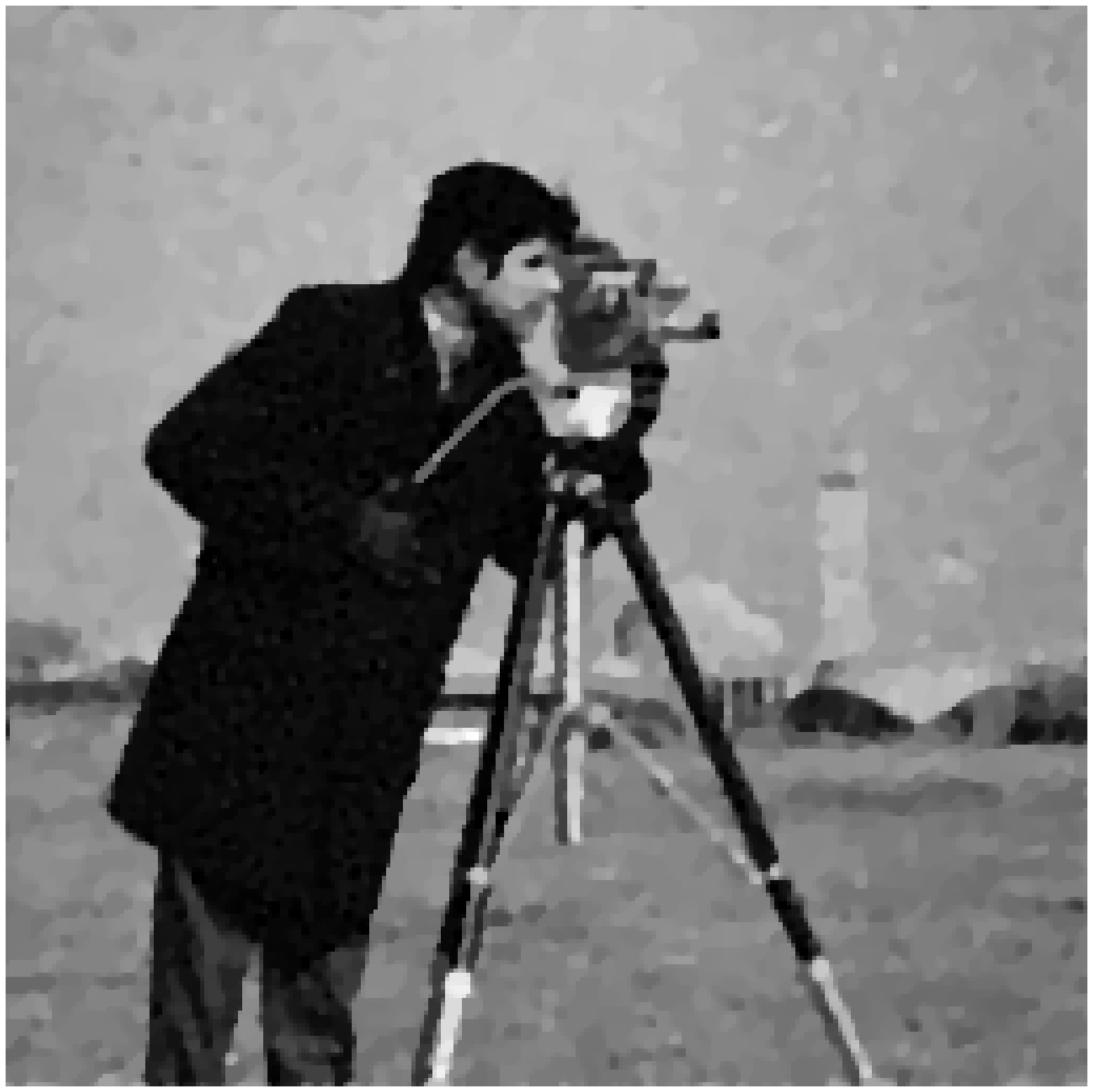}}
\caption{(a) The original Cameraman image, (b) the blurry and noisy image: Gaussian blur with $I_{\max}=100$, (c) the image restored by PIDAL, (d) the image restored by PLAD, (e) the image restored by IADMND, (f) the image restored by IADMNDA.
}
\label{fig4.4}
\end{figure}

\begin{figure}
  \centering
  \subfigure[]{
    \label{fig4.6:subfig:a}
    \includegraphics[width=1.5in,clip]{bridge.eps}}
  \subfigure[]{
    \label{fig4.6:subfig:b}
    \includegraphics[width=1.5in,clip]{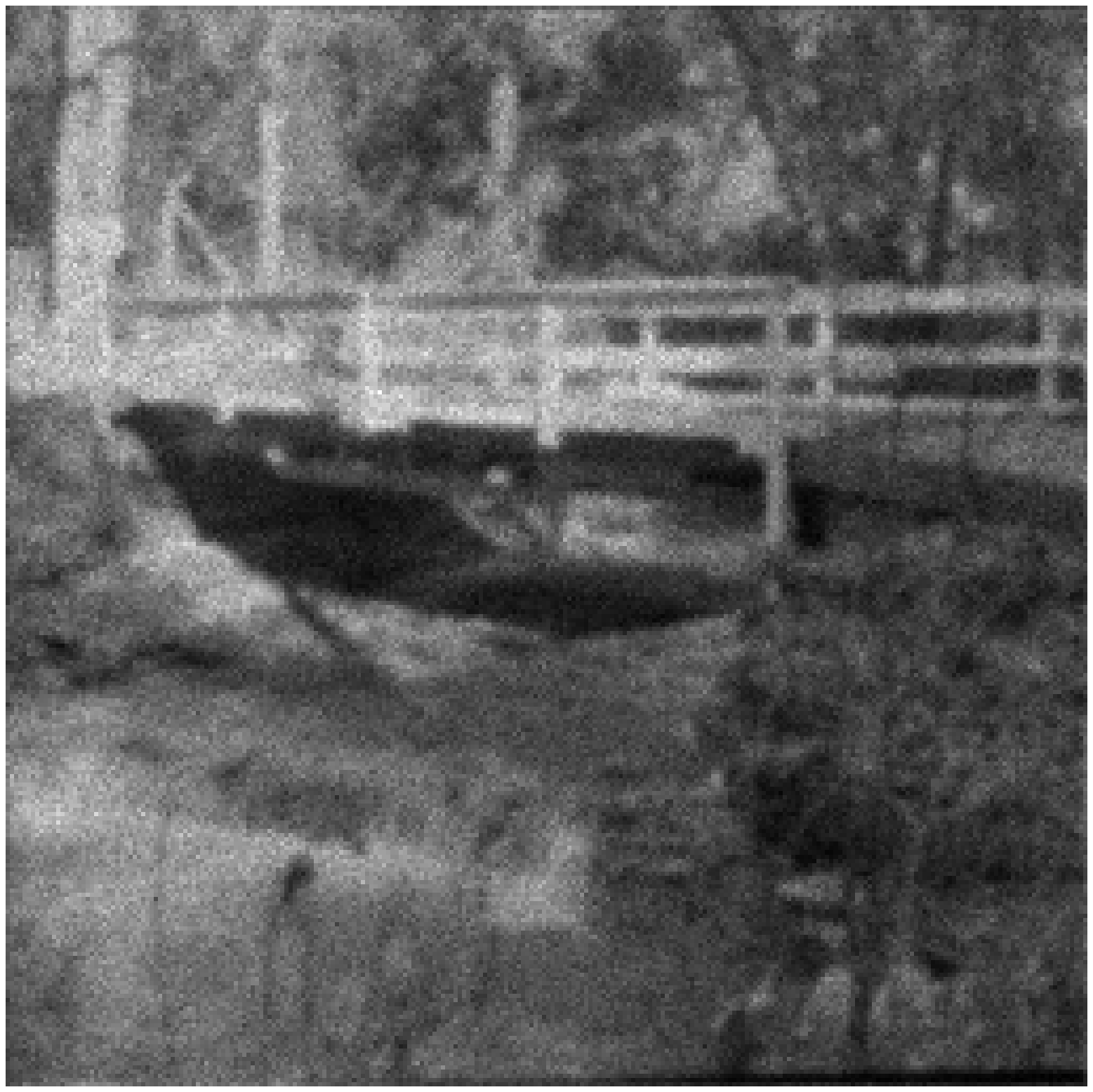}}
  \subfigure[]{
    \label{fig4.6:subfig:c}
    \includegraphics[width=1.5in,clip]{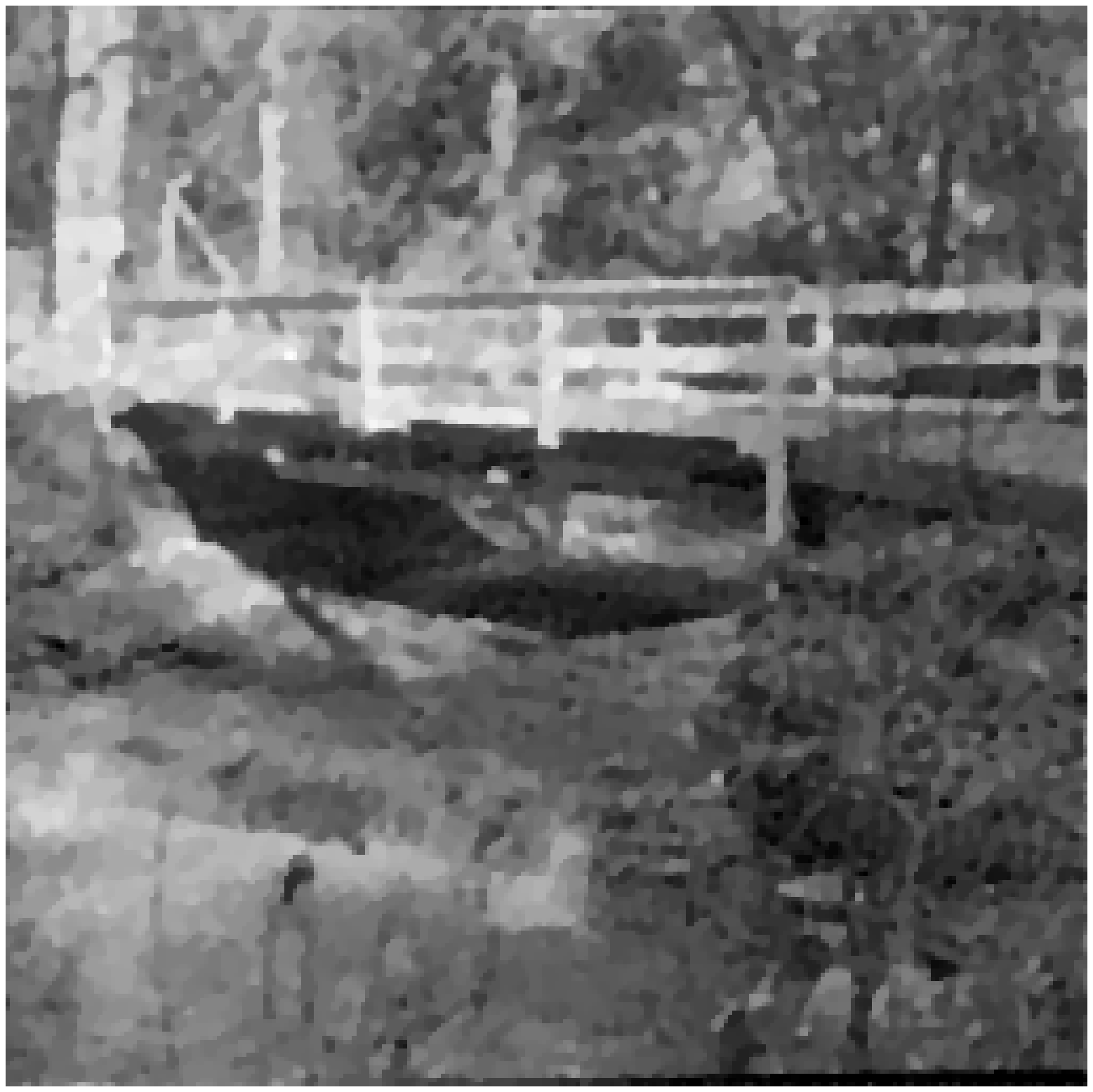}}
  \subfigure[]{
    \label{fig4.6:subfig:d}
    \includegraphics[width=1.5in,clip]{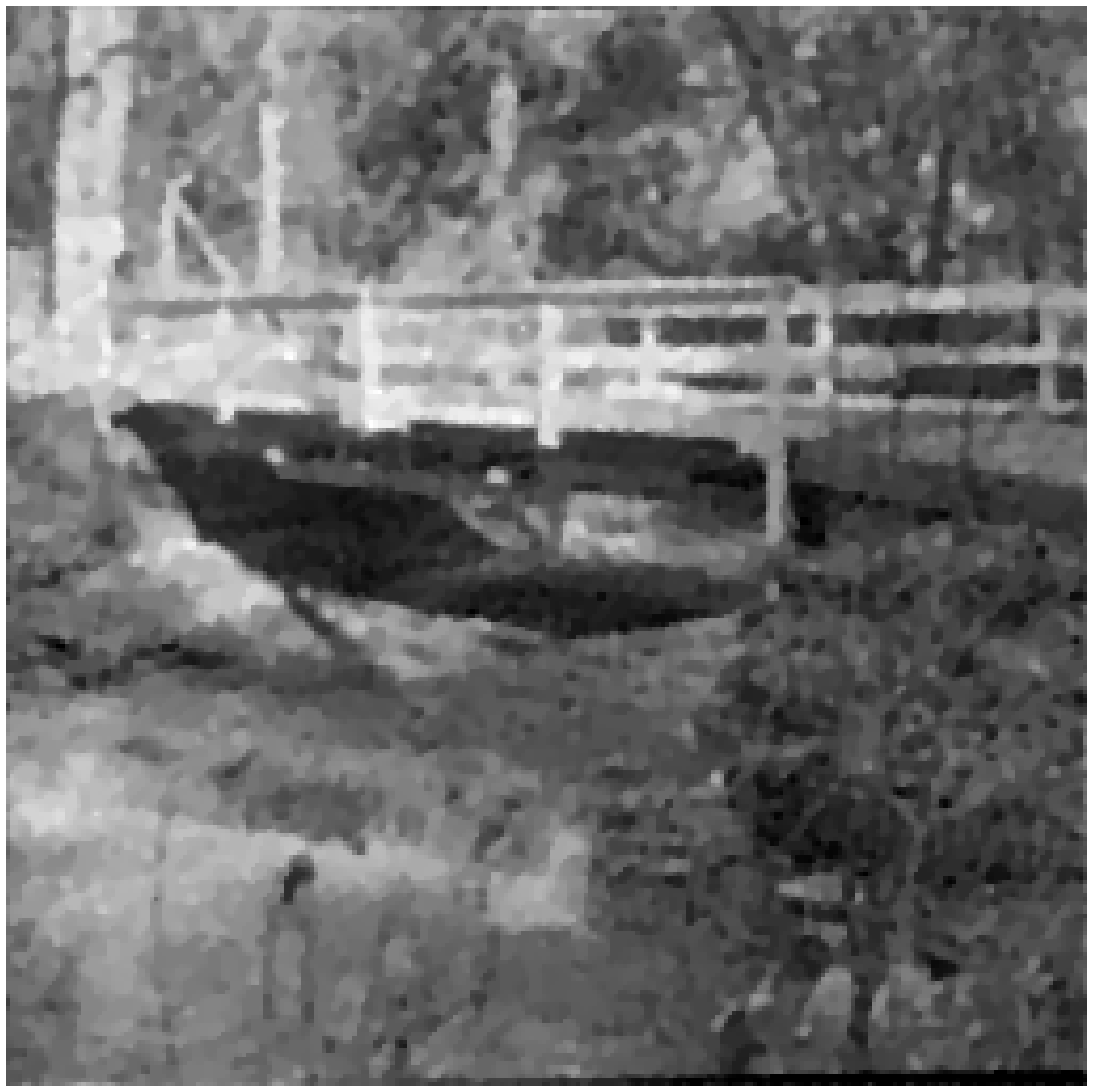}}
  \subfigure[]{
    \label{fig4.6:subfig:e}
    \includegraphics[width=1.5in,clip]{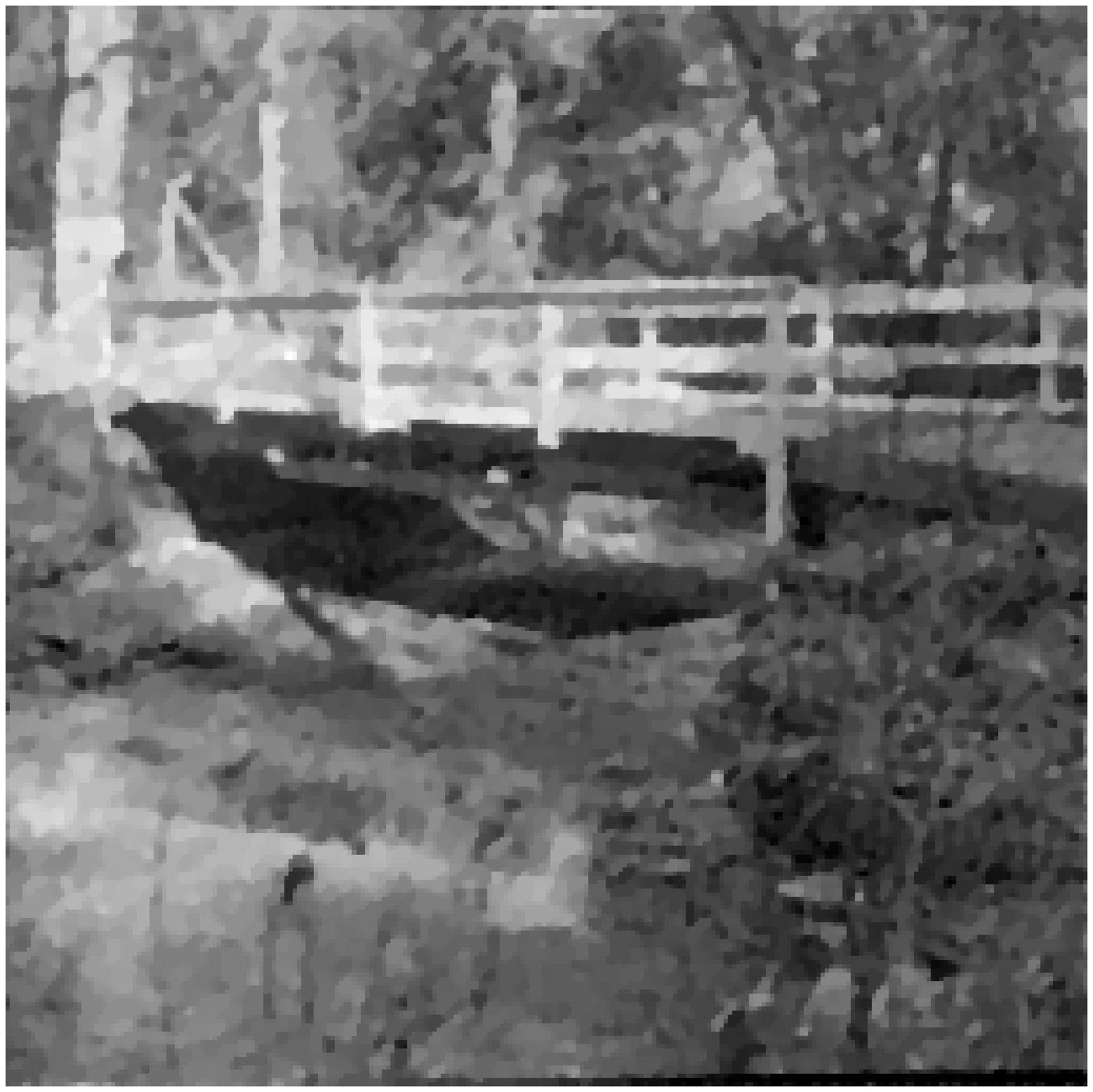}}
  \subfigure[]{
    \label{fig4.6:subfig:f}
    \includegraphics[width=1.5in,clip]{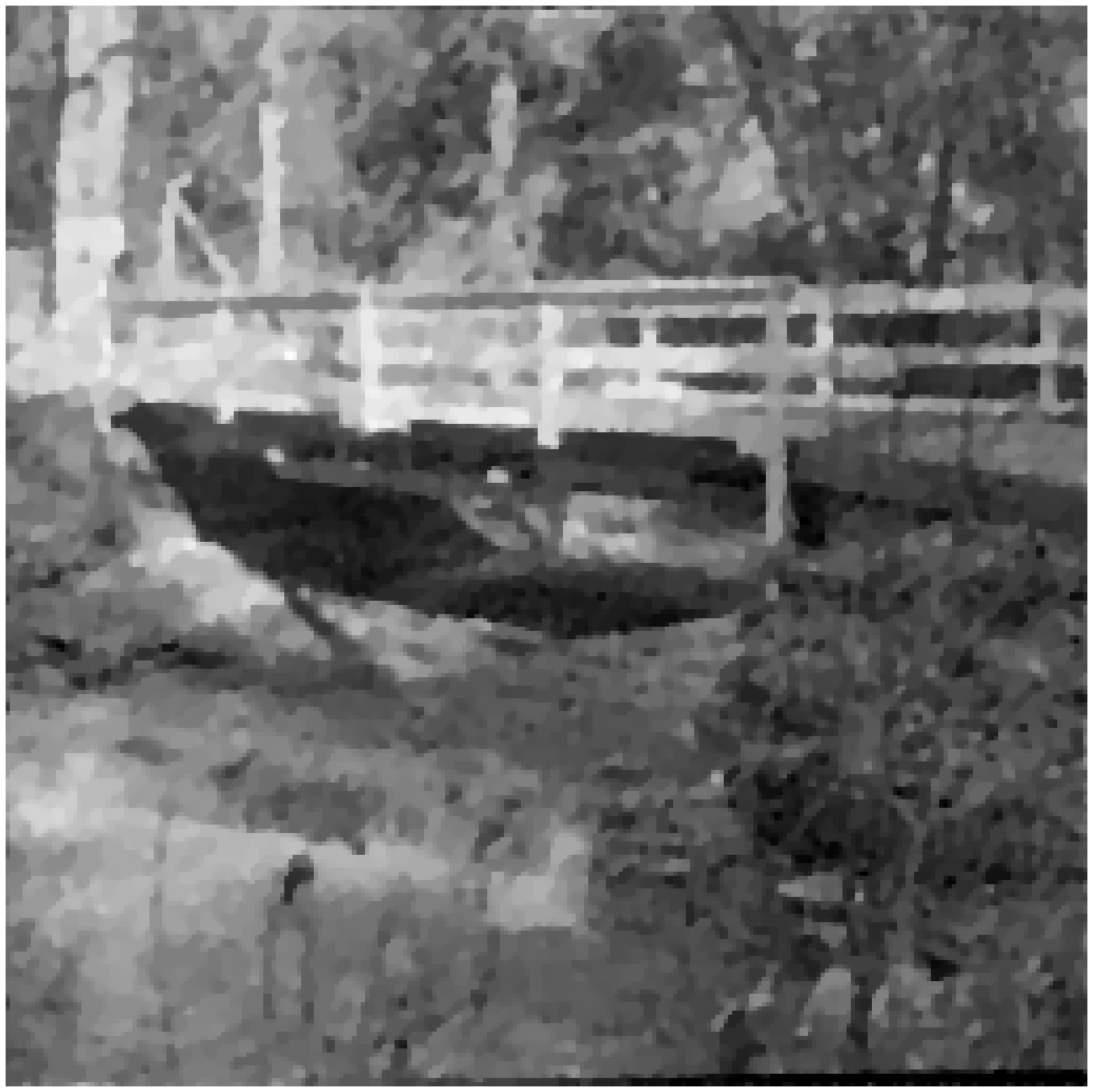}}
\caption{(a) The original Bridge image, (b) the blurry and noisy image: Gaussian blur with $I_{\max}=200$, (c) the image restored by PIDAL, (d) the image restored by PLAD, (e) the image restored by IADMND, (f) the image restored by IADMNDA.
}
\label{fig4.6}
\end{figure}

\begin{figure}
  \centering
  \subfigure[]{
    \label{fig4.9:subfig:a}
    \includegraphics[width=1.5in,clip]{boat.eps}}
  \subfigure[]{
    \label{fig4.9:subfig:b}
    \includegraphics[width=1.5in,clip]{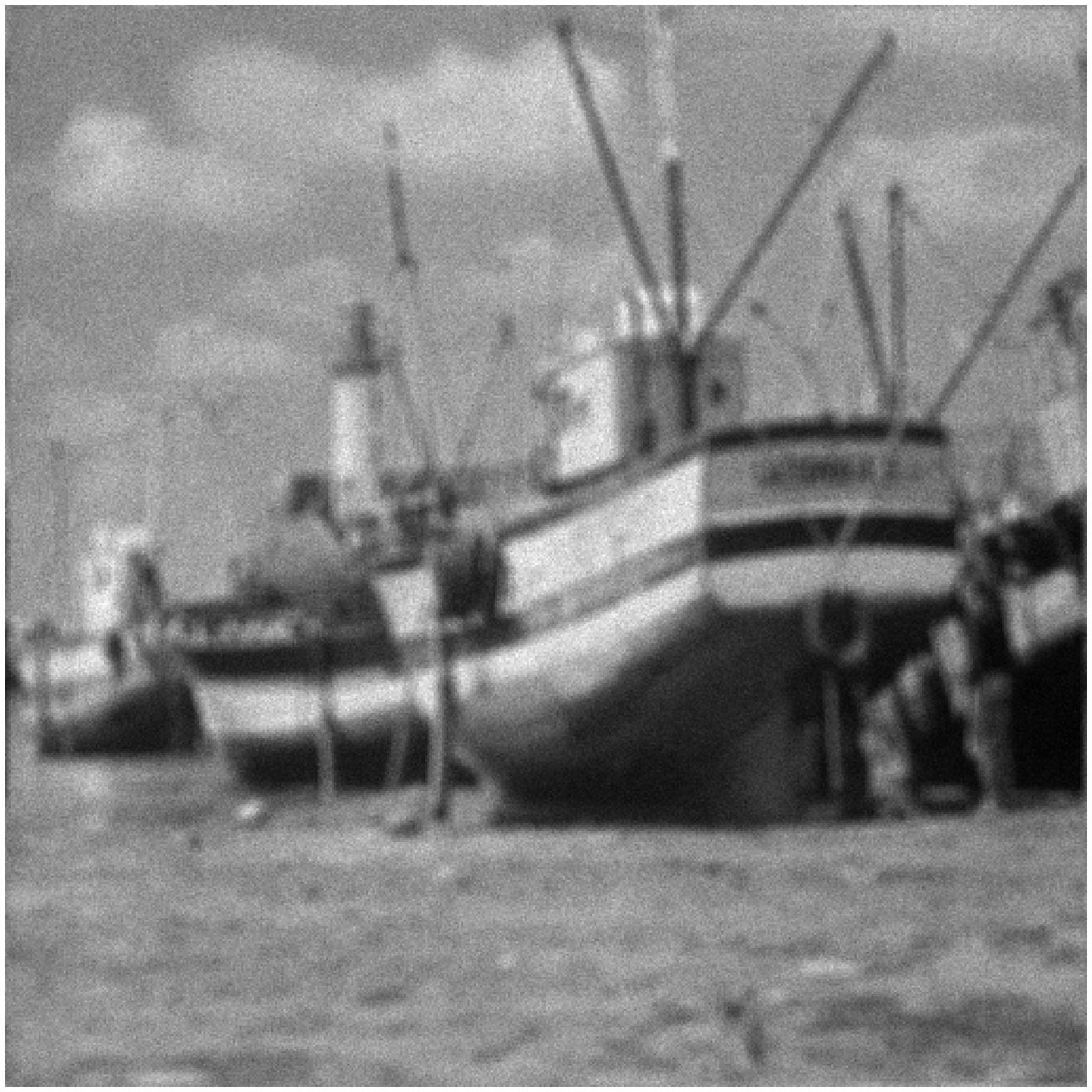}}
  \subfigure[]{
    \label{fig4.9:subfig:c}
    \includegraphics[width=1.5in,clip]{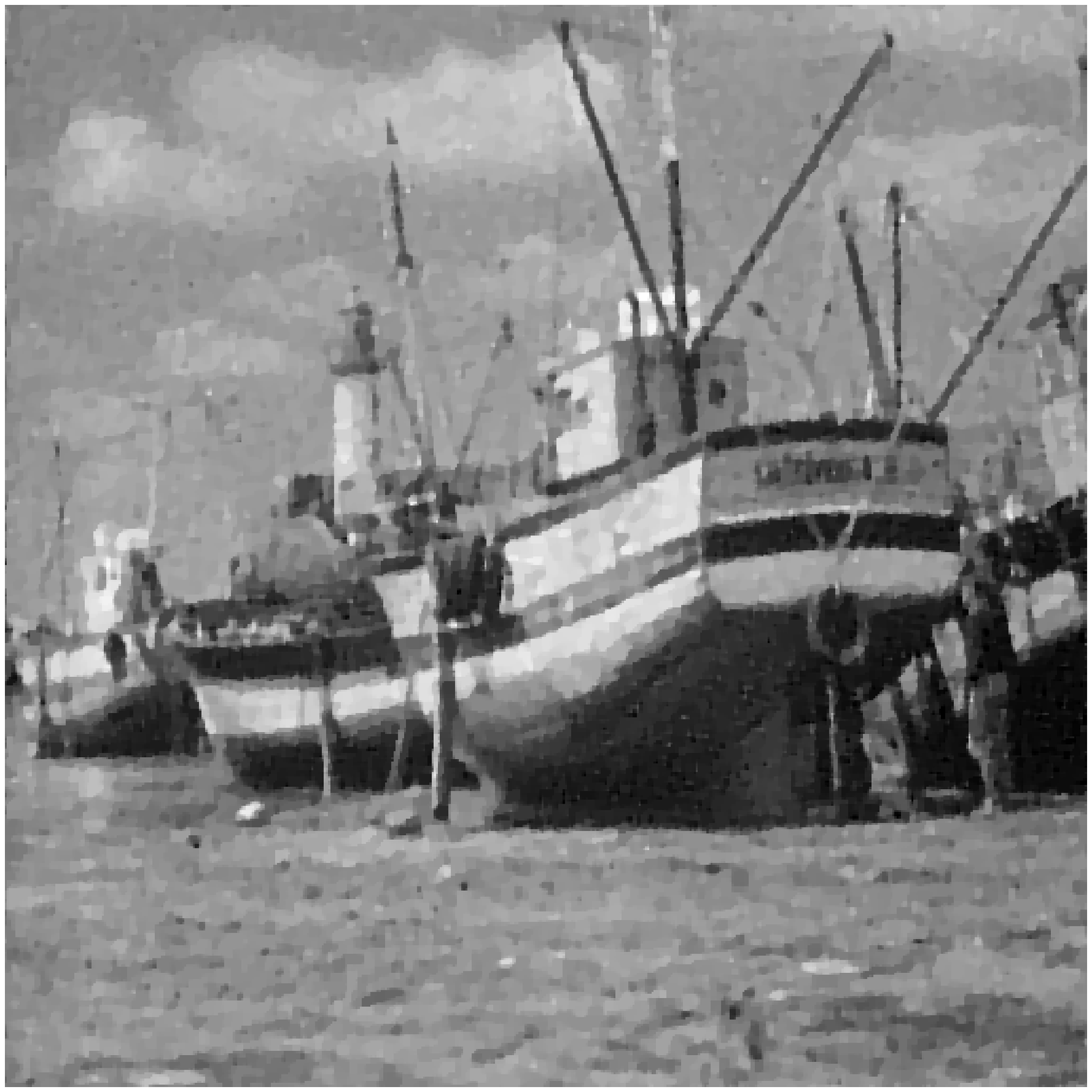}}
  \subfigure[]{
    \label{fig4.9:subfig:d}
    \includegraphics[width=1.5in,clip]{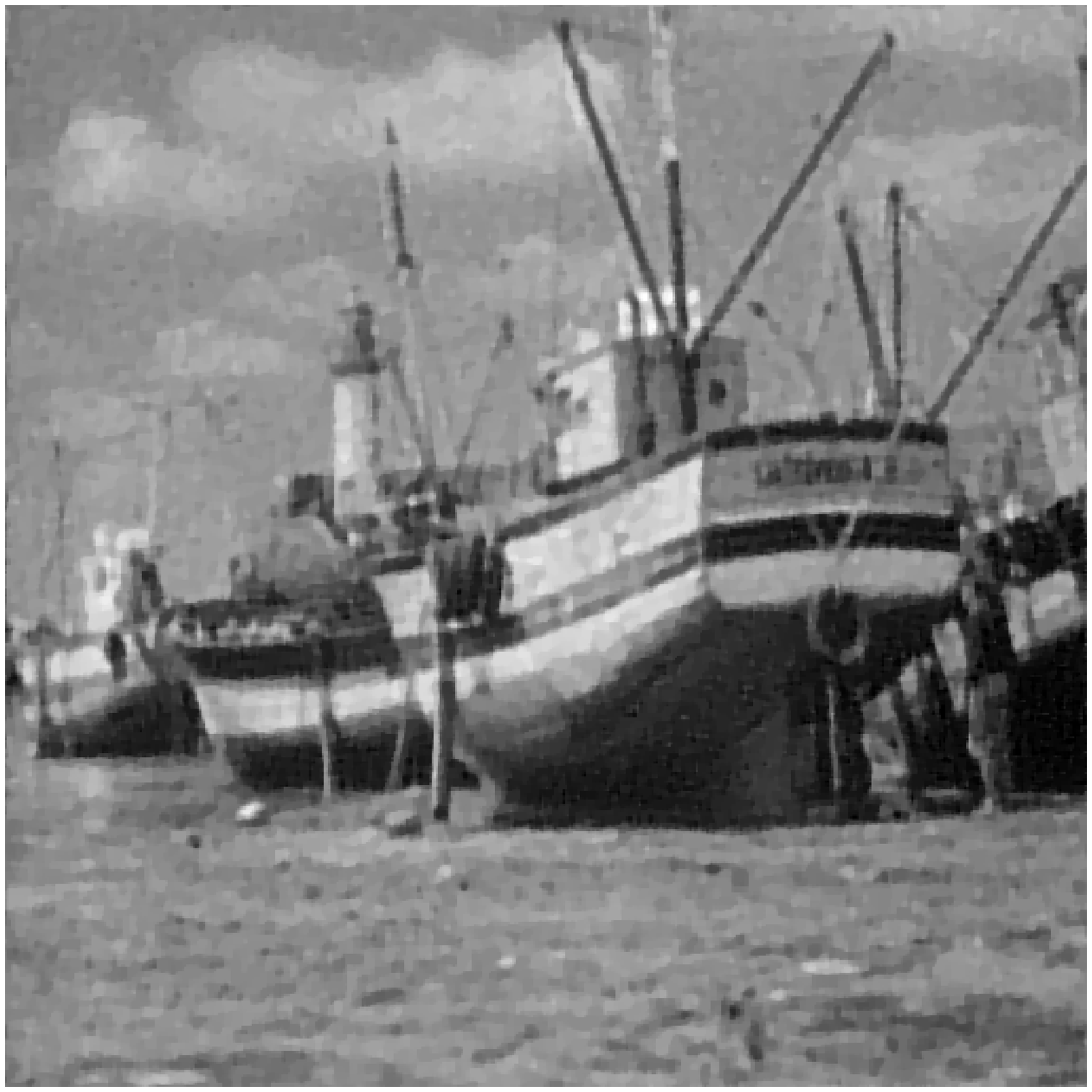}}
  \subfigure[]{
    \label{fig4.9:subfig:e}
    \includegraphics[width=1.5in,clip]{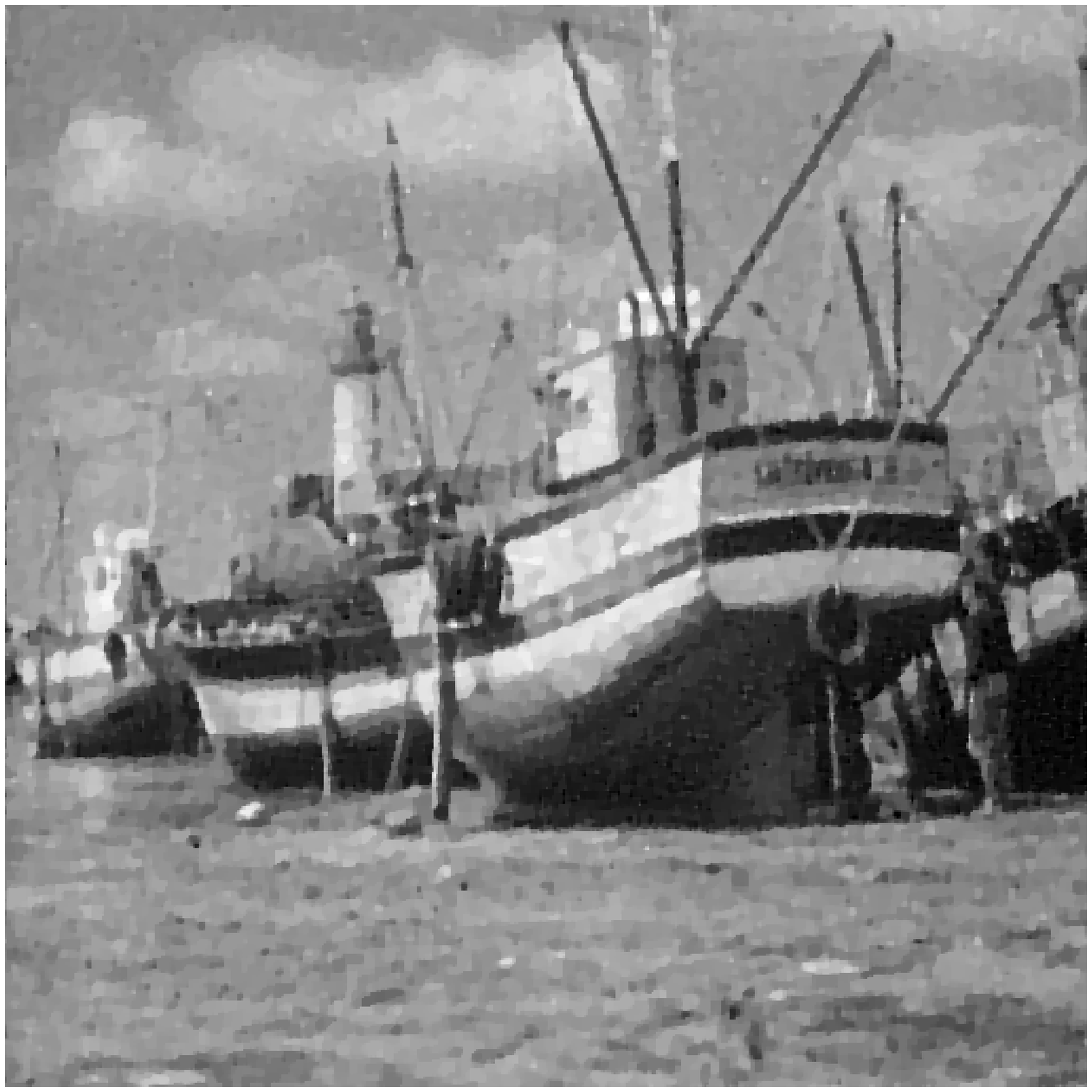}}
  \subfigure[]{
    \label{fig4.9:subfig:f}
    \includegraphics[width=1.5in,clip]{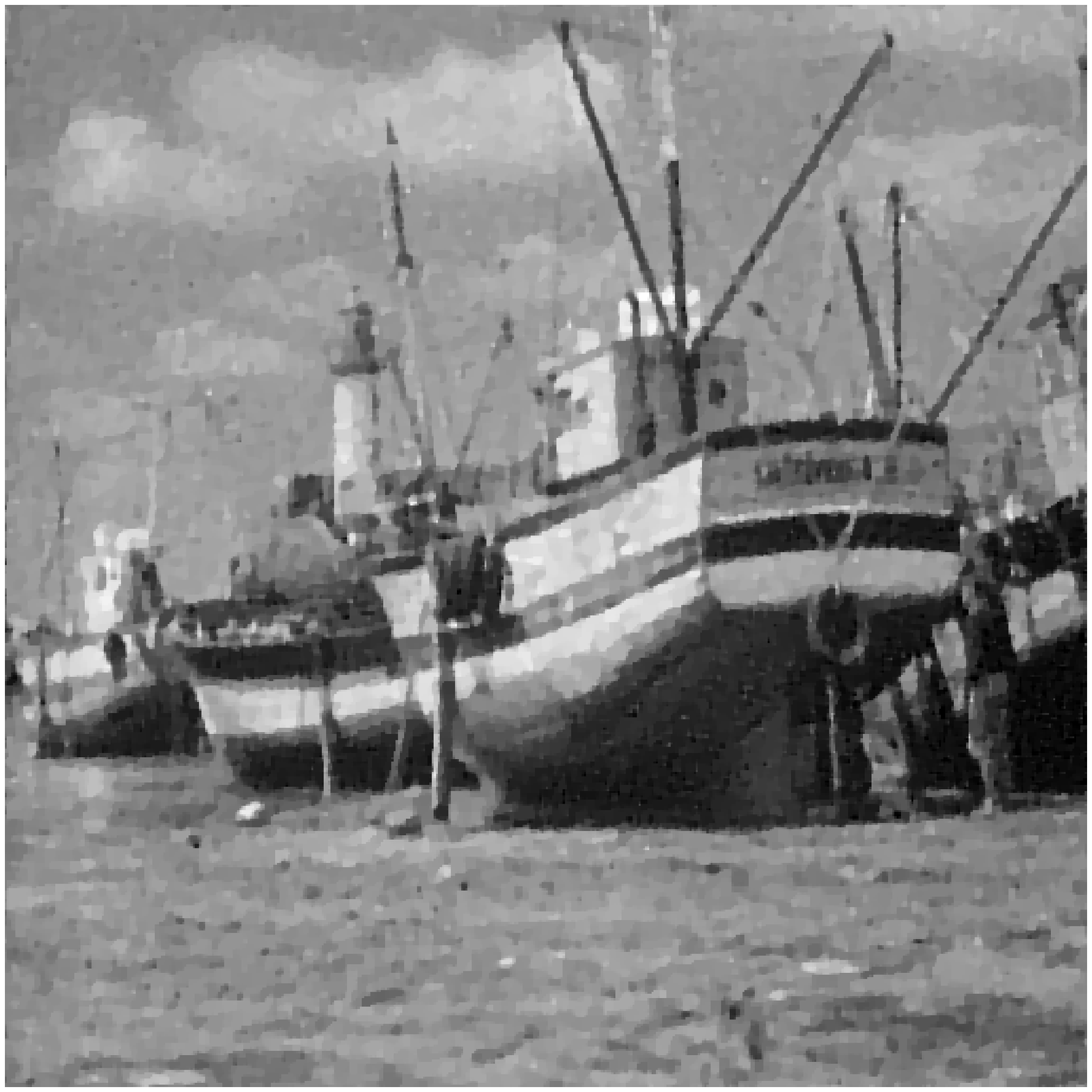}}
\caption{(a) The original Boat image, (b) the blurry and noisy image: uniform blur with $I_{\max}=500$, (c) the image restored by PIDAL, (d) the image restored by PLAD, (e) the image restored by IADMND, (f) the image restored by IADMNDA.
}
\label{fig4.9}
\end{figure}

Figures \ref{fig4.4}--\ref{fig4.9} show the recovery results of the PIDAL methods, the PLAD algorithm and the proposed IADMND and IADMNDA algorithms with respect to the Cameraman, Bridge and Boat images respectively. It is observed that the visual qualities of images generated by these algorithms are more or less the same.

\begin{figure}
  \centering
  \subfigure[]{
    \label{fig4.10:subfig:a}
    \includegraphics[width=1.5in,clip]{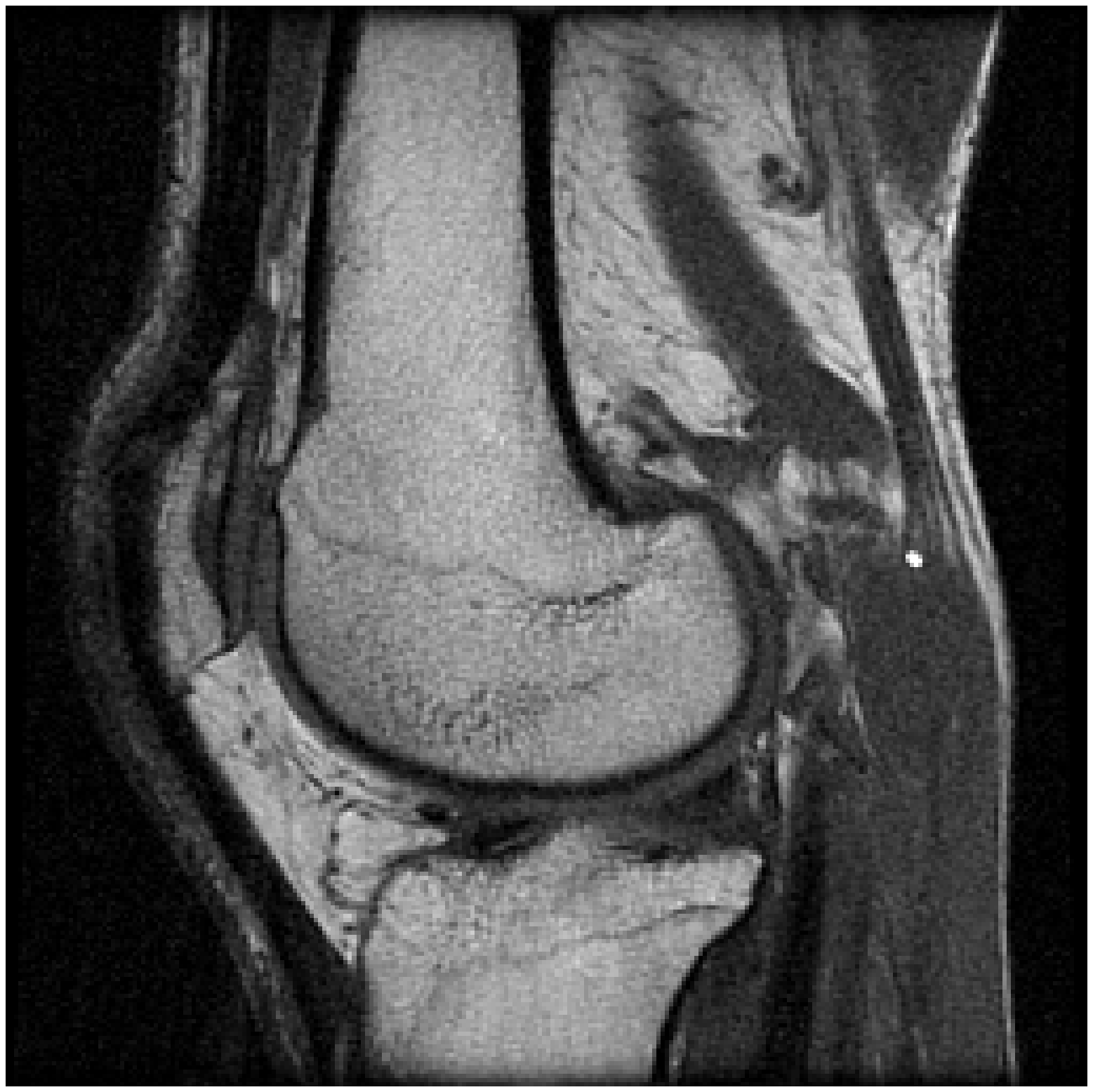}}
  \subfigure[]{
    \label{fig4.10:subfig:b}
    \includegraphics[width=1.5in,clip]{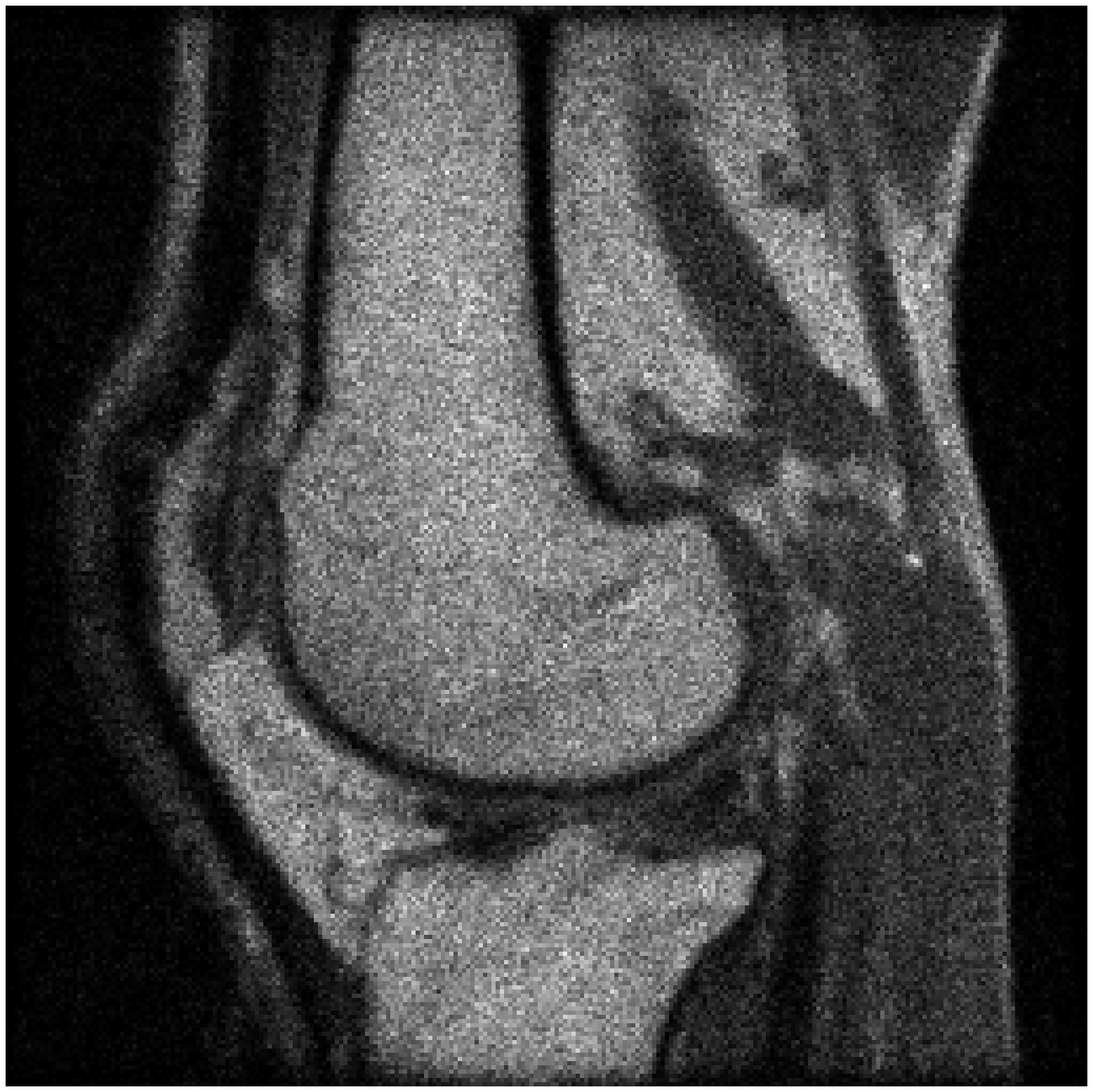}}
  \subfigure[]{
    \label{fig4.10:subfig:c}
    \includegraphics[width=1.5in,clip]{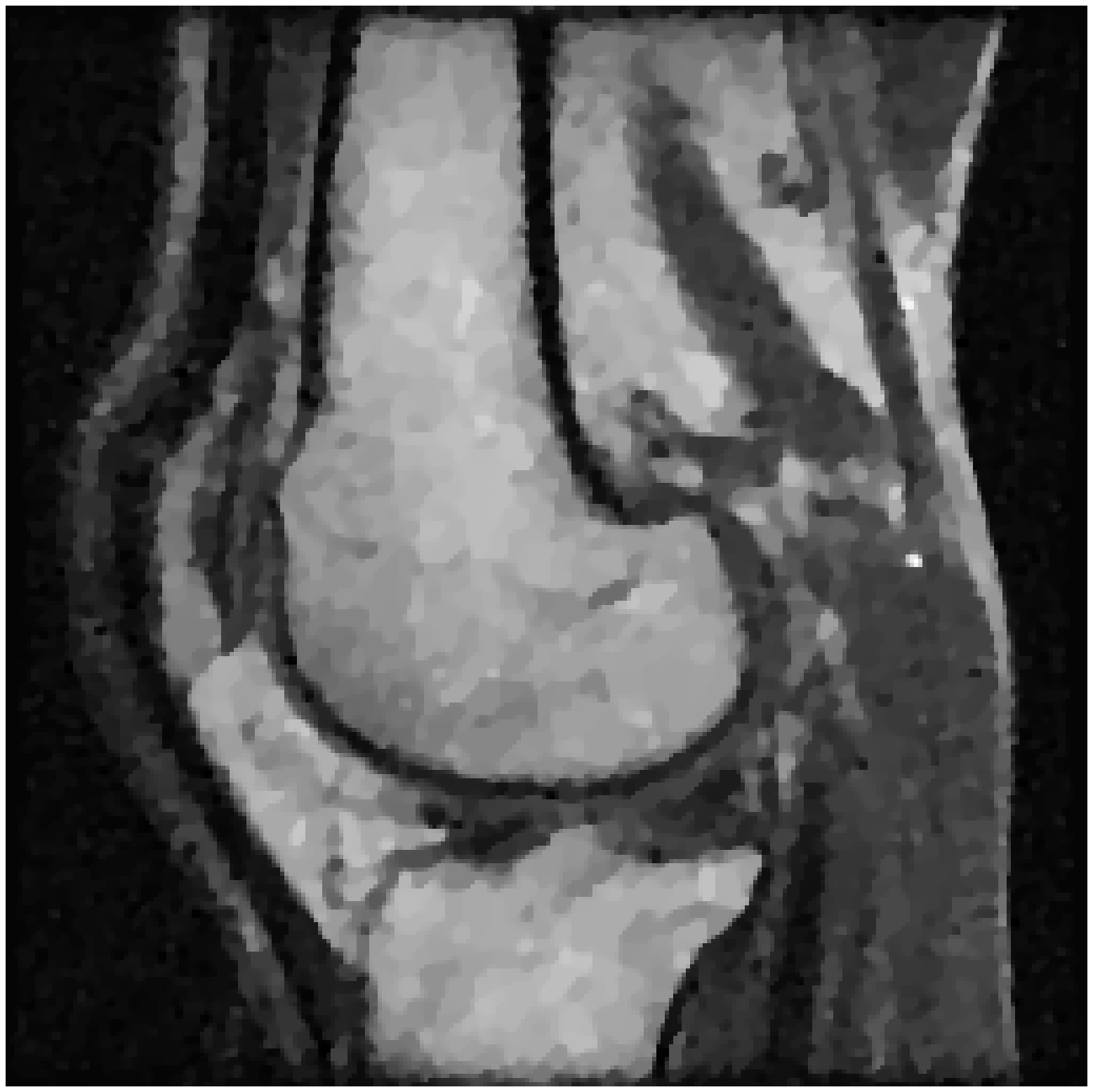}}
  \subfigure[]{
    \label{fig4.10:subfig:d}
    \includegraphics[width=1.5in,clip]{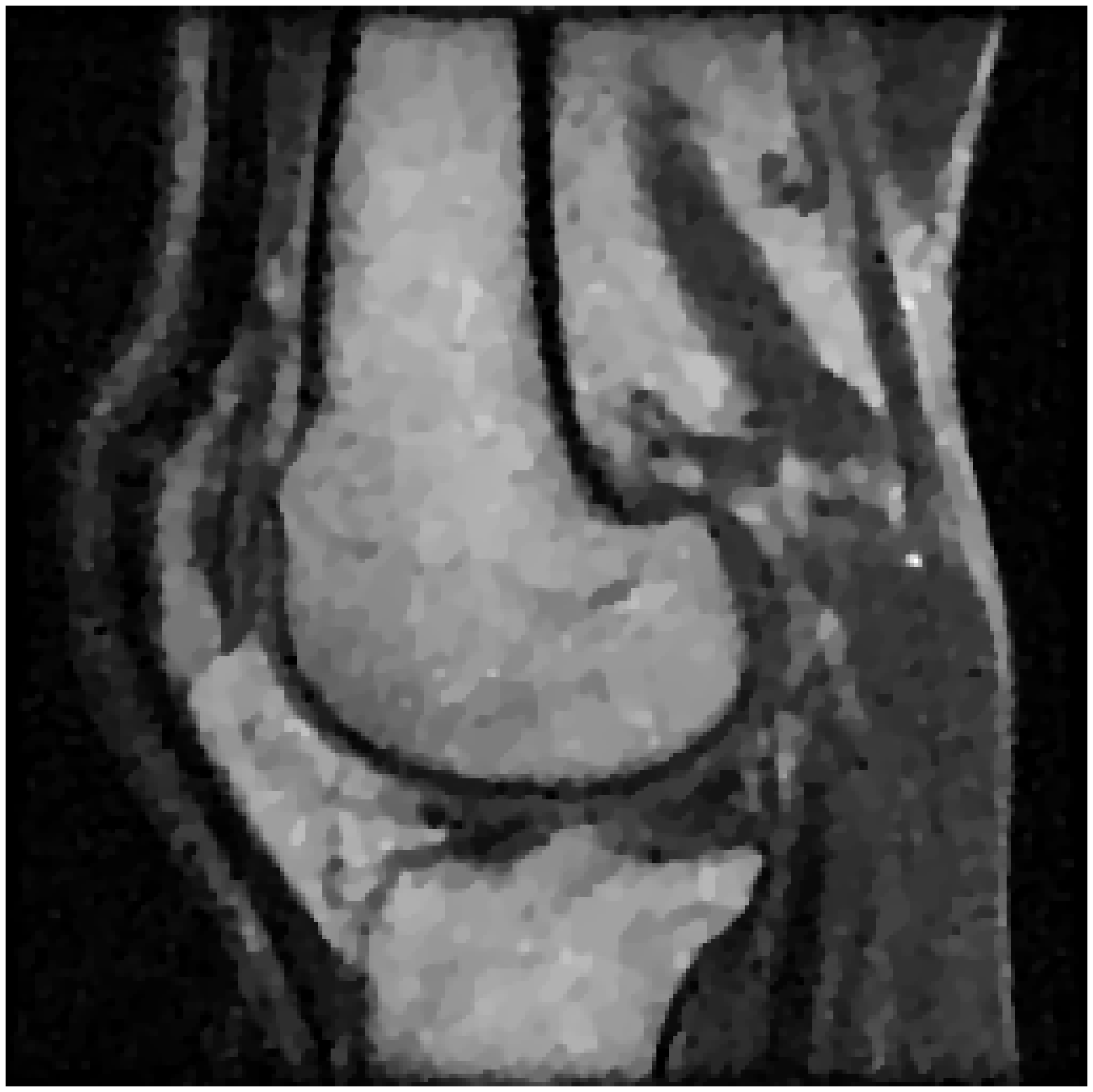}}
  \subfigure[]{
    \label{fig4.10:subfig:e}
    \includegraphics[width=1.5in,clip]{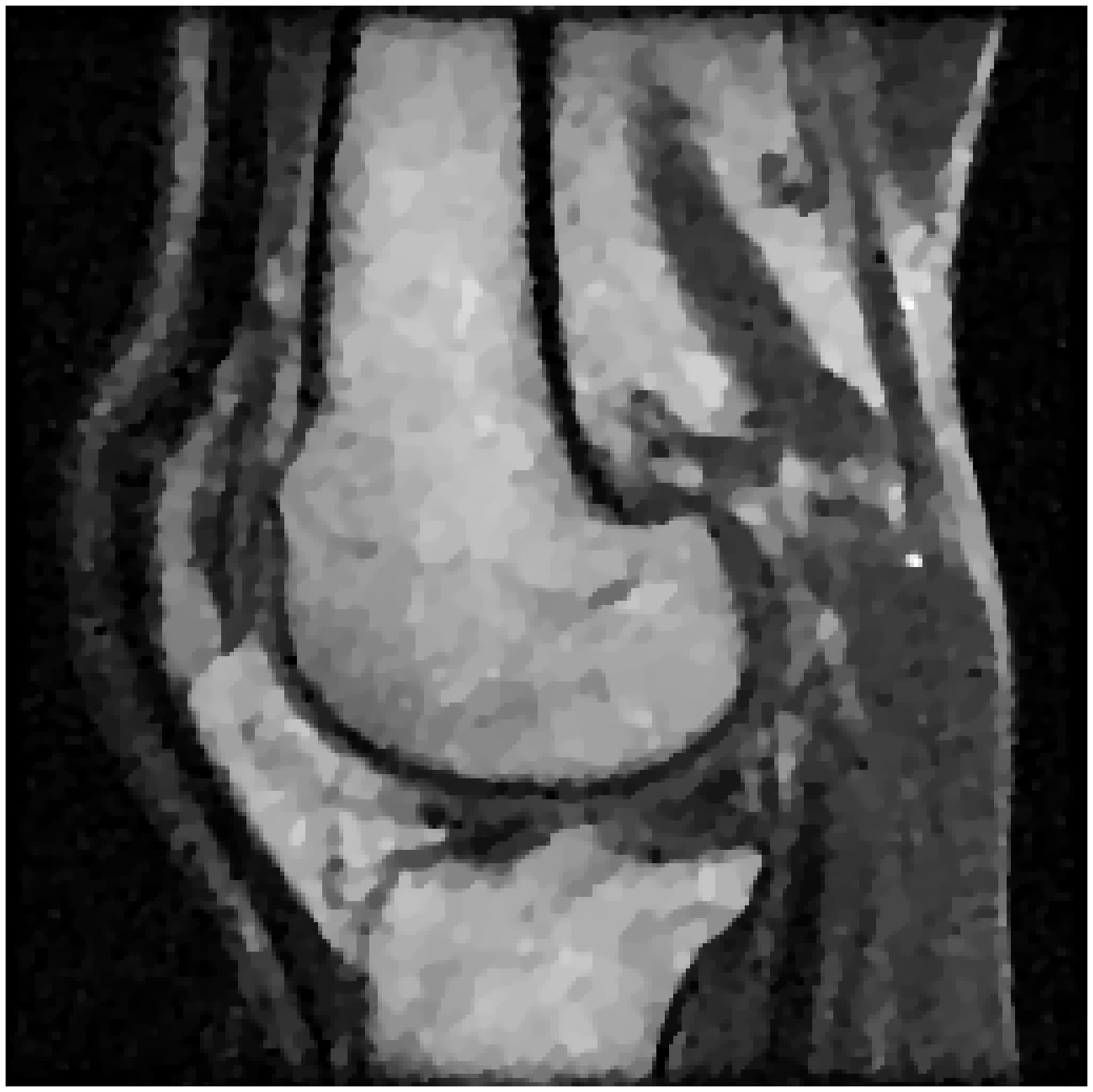}}
  \subfigure[]{
    \label{fig4.10:subfig:f}
    \includegraphics[width=1.5in,clip]{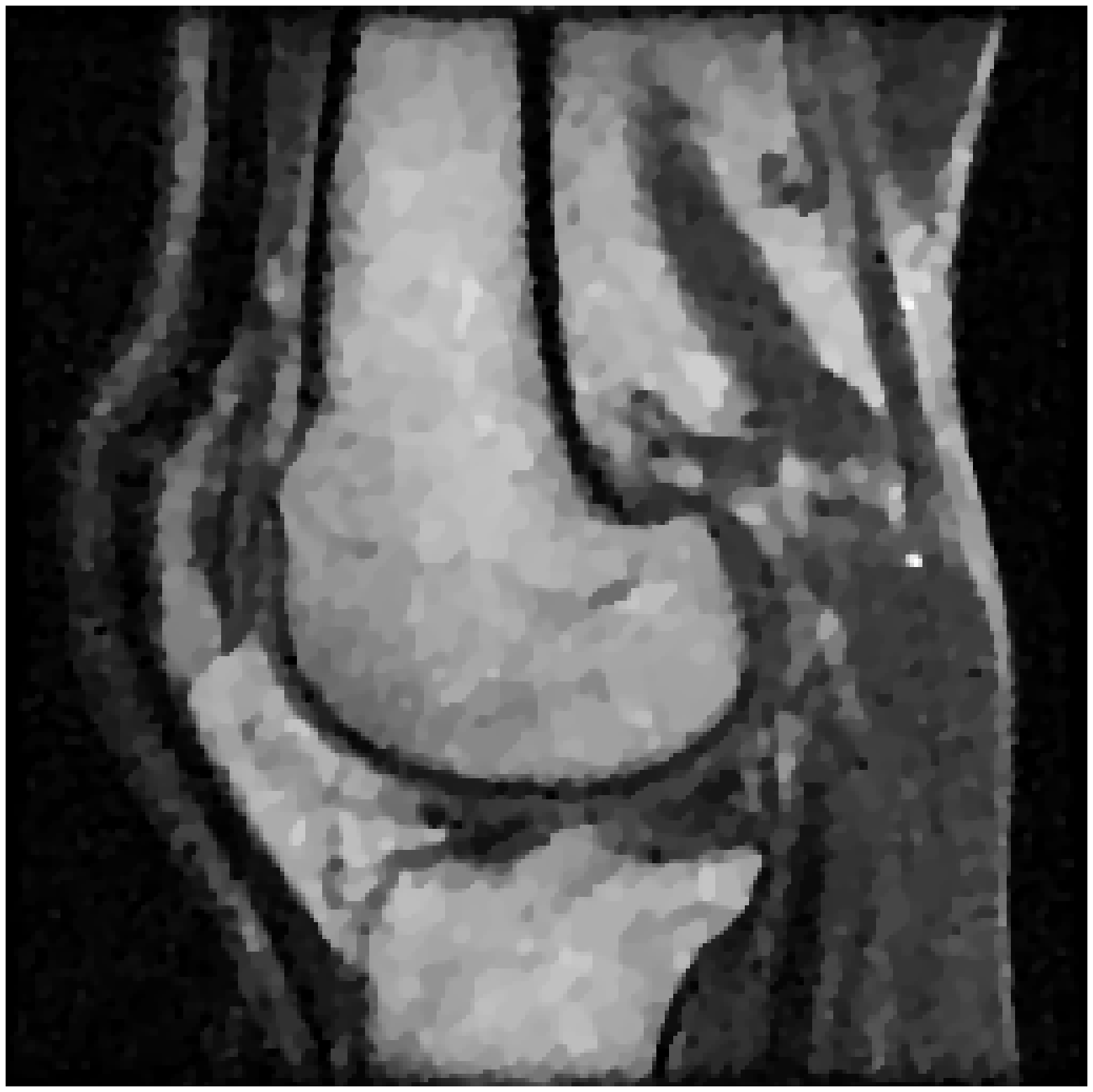}}
\caption{(a) The original rkknee image, (b) the blurry and noisy image: Gaussian blur with $I_{\max}=50$, (c) the image restored by PIDAL, (d) the image restored by PLAD, (e) the image restored by IADMND, (f) the image restored by IADMNDA.
}
\label{fig4.10}
\end{figure}

Finally, we consider two MRI images called ``rkknee" and ``chest". Table \ref{tab4.4} lists the SNR values, the number of iterations and CPU time of different algorithms for images with different blur kernels and noise levels. The regularization parameter $\lambda$ is set to $0.06$ and $0.04$ for images with $I_{\max}=50$ and convoluted by Gaussian and uniform blur kernels respectively. Similarly, we notice that the proposed algorithms are the most efficient in the computational time. Some recovery results are shown in Figures \ref{fig4.10}. We find that the quality of recovery images by these algorithms is very similar.

\begin{table} [htbp]
\centering \caption{The comparison of the performance of different algorithms under mild blur condition: the given numbers are SNR (dB)/Iteration number/CPU time(second) }
\scalebox{0.9}{
\begin{tabular}{ccccccc}
  \hline
  Image & blur kernel & $I_{\max}$ & PIDAL \cite{TIP:PIDAL} & PLAD \cite{SIAMJSC:Linearized} & IADMND & IADMNDA \\
  \hline
    \multirow{6}{*}{Cameraman} &  & 100 & 13.55/56/2.76 & \textbf{13.57}/91/2.57 & 13.53/56/\textbf{1.51} & 13.55/53/1.74   \\
  \cline{3-7}
     & Gaussian & 200 & \textbf{14.36}/50/2.26 & 14.23/132/3.57 & 14.35/46/\textbf{1.19} & \textbf{14.36}/47/1.47 \\
  \cline{3-7}
     &  & 500 & \textbf{15.44}/56/2.64 & 15.21/109/3.18 & 15.34/64/1.84 & 15.42/42/\textbf{1.36} \\
  \cline{2-7}
     &  & 100 & 11.31/83/3.88 & 11.17/139/3.65 & \textbf{11.32}/63/\textbf{1.75} & \textbf{11.32}/58/1.78 \\
  \cline{3-7}
     & Uniform & 200 & \textbf{11.83}/85/4.04 & 11.74/199/5.58 & \textbf{11.83}/54/\textbf{1.56} & 11.82/67/2.32 \\
  \cline{3-7}
     &  & 500 & \textbf{12.68}/99/5.02 & 12.31/199/5.55 & 12.62/76/2.11 & 12.66/56/\textbf{1.92} \\
  \hline\hline
    \multirow{6}{*}{Barbara} &  & 100 & 9.93/45/2.03 & \textbf{9.95}/80/2.26 & 9.92/49/\textbf{1.28} & 9.92/38/1.43   \\
  \cline{3-7}
     & Gaussian & 200 & 10.55/39/1.89 & \textbf{10.72}/120/2.94 & 10.54/39/\textbf{1.00} & 10.54/33/1.15 \\
  \cline{3-7}
     &  & 500 & \textbf{12.23}/39/1.78 & 12.10/107/2.48 & 12.15/56/1.44 & 12.22/31/\textbf{1.01} \\
  \cline{2-7}
     &  & 100 & 9.04/67/3.23 & \textbf{9.09}/116/3.10 & 9.02/54/\textbf{1.39} & 9.02/45/1.67 \\
  \cline{3-7}
     & Uniform & 200 & 9.19/58/2.92 & \textbf{9.29}/126/3.34 & 9.18/46/1.17 & 9.17/38/\textbf{1.12} \\
  \cline{3-7}
     &  & 500 & 9.56/59/2.76 & \textbf{9.63}/124/3.28 & 9.56/68/1.73 & 9.55/35/\textbf{1.04} \\
  \hline\hline
    \multirow{6}{*}{Bridge} &  & 100 & 10.51/54/2.70 & \textbf{10.53}/93/2.56 & 10.50/52/\textbf{1.42} & 10.51/45/1.56   \\
  \cline{3-7}
     & Gaussian & 200 & 11.26/48/2.32 & \textbf{11.27}/129/3.46 & 11.25/45/\textbf{1.32} & 11.26/41/1.48 \\
  \cline{3-7}
     &  & 500 & 12.08/43/2.12 & \textbf{12.15}/110/2.96 & 12.05/63/1.62 & 12.08/35/\textbf{1.06} \\
  \cline{2-7}
     &  & 100 & \textbf{8.43}/81/3.96 & \textbf{8.43}/137/3.62 & 8.42/61/1.70 & \textbf{8.43}/51/\textbf{1.64} \\
  \cline{3-7}
     & Uniform & 200 & 9.00/72/3.37 & \textbf{9.03}/161/4.37 & 8.99/54/\textbf{1.39} & 8.99/46/1.50 \\
  \cline{3-7}
     &  & 500 & \textbf{9.64}/69/3.32 & 9.59/171/4.62 & 9.63/81/2.11 & \textbf{9.64}/42/\textbf{1.39} \\
  \hline\hline

    \multirow{6}{*}{Pepper} &  & 100 & 11.60/51/2.65 & \textbf{11.65}/88/2.17 & 11.58/52/\textbf{1.25} & 11.58/47/1.65   \\
  \cline{3-7}
     & Gaussian & 200 & 12.30/45/2.15 & \textbf{12.32}/125/3.47 & 12.29/44/1.37 & 12.29/39/\textbf{1.32} \\
  \cline{3-7}
     &  & 500 & 13.11/45/2.75 & 13.09/100/2.66 & \textbf{13.14}/59/1.74 & \textbf{13.14}/38/\textbf{1.22} \\
  \cline{2-7}
     &  & 100 & 9.97/84/4.01 & \textbf{9.99}/141/3.81 & 9.95/62/\textbf{1.62} & 9.95/53/1.78 \\
  \cline{3-7}
     & Uniform & 200 & 10.45/75/3.90 & \textbf{10.48}/157/4.26 & 10.44/55/\textbf{1.53} & 10.44/50/1.81 \\
  \cline{3-7}
     &  & 500 & \textbf{11.67}/74/3.70 & 11.21/163/4.27 & 11.59/81/2.18 & 11.65/49/\textbf{1.64} \\
  \hline\hline

    \multirow{6}{*}{Lenna} &  & 100 & 13.41/56/2.96 & \textbf{13.43}/88/2.24 & 13.39/57/\textbf{1.73} & 13.41/48/1.79   \\
  \cline{3-7}
     & Gaussian & 200 & \textbf{14.29}/49/2.45 & 14.19/126/3.67 & 14.27/48/\textbf{1.22} & 14.28/43/1.62 \\
  \cline{3-7}
     &  & 500 & 15.16/44/1.83 & \textbf{15.21}/99/2.79 & 15.14/62/1.76 & 15.15/39/\textbf{1.34} \\
  \cline{2-7}
     &  & 100 & \textbf{11.04}/86/4.40 & 11.00/140/3.67 & 11.02/66/\textbf{1.68} & 11.02/56/2.02 \\
  \cline{3-7}
     & Uniform & 200 & 11.29/75/3.71 & \textbf{11.34}/151/3.93 & 11.28/58/1.58 & 11.28/51/\textbf{1.45} \\
  \cline{3-7}
     &  & 500 & 12.07/69/3.40 & 11.89/148/3.95 & \textbf{12.08}/80/2.15 & 12.07/45/\textbf{1.56} \\
  \hline\hline

    \multirow{6}{*}{Boat} &  & 100 & 12.77/46/8.46 & \textbf{12.82}/80/9.20 & 12.73/49/5.57 & 12.75/40/\textbf{5.19}   \\
  \cline{3-7}
     & Gaussian & 200 & 13.51/39/7.35 & \textbf{13.54}/121/13.35 & 13.49/39/\textbf{4.85} & 13.50/37/5.21 \\
  \cline{3-7}
     &  & 500 & 14.44/39/7.53 & \textbf{14.59}/94/9.81 & 14.44/53/6.32 & 14.44/33/\textbf{4.48} \\
  \cline{2-7}
     &  & 100 & 10.41/64/12.48 & 10.40/111/12.26 & 10.41/53/\textbf{5.83} & \textbf{10.42}/46/5.85 \\
  \cline{3-7}
     & Uniform & 200 & 10.91/65/13.04 & \textbf{10.96}/199/22.28 & 10.90/47/\textbf{5.15} & 10.90/43/5.53 \\
  \cline{3-7}
     &  & 500 & \textbf{11.74}/68/13.10 & 11.69/199/22.70 & 11.73/69/7.71 & \textbf{11.74}/38/\textbf{4.01} \\
  \hline
\end{tabular}}
\label{tab4.2}
\end{table}

\begin{table} [htbp]
\centering \caption{The comparison of the performance of different algorithms under mild blur condition: the given numbers are SNR (dB)/Iteration number/CPU time(second) }
\scalebox{0.9}{
\begin{tabular}{ccccccc}
  \hline
  Image & blur kernel & $I_{\max}$ & PIDAL \cite{TIP:PIDAL} & PLAD \cite{SIAMJSC:Linearized} & IADMND & IADMNDA \\
  \hline
    \multirow{6}{*}{rkknee} &  & 50 & 12.99/59/2.97 & \textbf{13.04}/142/3.72 & 12.97/58/\textbf{1.46} & 12.98/52/1.84   \\
  \cline{3-7}
     & Gaussian & 100 & 14.14/51/2.21 & \textbf{14.25}/190/5.05 & 14.12/62/\textbf{1.52} & 14.12/53/1.73 \\
  \cline{3-7}
     &  & 200 & \textbf{14.82}/53/2.53 & 14.72/198/5.24 & 14.79/71/2.05 & 14.80/56/\textbf{1.70} \\
  \cline{2-7}
     &  & 50 & 11.63/86/3.92 & \textbf{11.75}/158/4.41 & 11.60/61/\textbf{1.75} & 11.60/60/2.09 \\
  \cline{3-7}
     & Uniform & 100 & 12.39/83/4.07 & \textbf{12.57}/192/5.51 & 12.37/69/\textbf{1.92} & 12.37/63/2.27 \\
  \cline{3-7}
     &  & 200 & 12.66/108/5.38 & 12.53/199/5.37 & \textbf{12.67}/98/2.58 & 12.64/99/3.40 \\
  \hline\hline
    \multirow{6}{*}{chest} &  & 50 & 11.83/72/2.17 & \textbf{11.93}/156/2.93 & 11.83/63/\textbf{1.14} & 11.83/63/1.37  \\
  \cline{3-7}
     & Gaussian & 100 & 13.17/63/2.03 & \textbf{13.35}/155/2.79 & 13.13/67/\textbf{1.36} & 13.14/63/1.48 \\
  \cline{3-7}
     &  & 200 & 14.25/59/1.81 & \textbf{14.59}/187/3.35 & 14.15/88/1.64 & 14.22/64/\textbf{1.20}  \\
  \cline{2-7}
     &  & 50 & 9.03/94/2.70 & \textbf{9.05}/152/3.03 & 9.03/66/\textbf{1.10} & 9.03/70/1.49 \\
  \cline{3-7}
     & Uniform & 100 & \textbf{9.73}/91/2.45 & 9.70/198/3.74 & 9.72/75/\textbf{1.37} & 9.72/83/2.04 \\
  \cline{3-7}
     &  & 200 & 10.38/94/2.70 & \textbf{10.43}/199/3.65 & 10.34/111/\textbf{2.28} & 10.36/99/\textbf{2.28} \\
  \hline
\end{tabular}}
\label{tab4.4}
\end{table}

\section{Conclusion}\label{sec5}

In this article, through further analyze the drawback of the recently proposed linearization techniques for image restoration, we develop an inexact alternating direction method based on the proximal Hessian matrix. Compared with the existing algorithms, the main difference is that the second-order derivative of the
objective function is just approximated by a proximal Hessian matrix in the proposed algorithm, rather than a identity matrix multiplied by a constant. Besides, we also propose a strategy for updating the proximal Hessian matrix. The convergence of the proposed algorithms is further investigated under certain conditions, and numerical experiments demonstrate that the proposed algorithms outperform the widely used linearized augmented Lagrangian methods in the computational time.

\section*{acknowledgement}
The work was supported in part by the National Natural Science Foundation of China under Grant 61271014 and 61401473.

\end{document}